%% file: main_arxiv_v3_nov2021.tex
\documentclass{article}

\usepackage[final,nonatbib]{neurips_2021}

\usepackage{tikz}
\usepackage{amsthm}
\usepackage[T1]{fontenc}
\usepackage{lmodern}

\usepackage{tcolorbox,wrapfig}
\usepackage[labelfont={bf},font=small]{caption}
\usepackage{subcaption}
\usepackage[ruled,vlined]{algorithm2e}
\tcbuselibrary{skins}
\usepackage[utf8]{inputenc} 
\usepackage[T1]{fontenc}    
\usepackage{hyperref}       
\usepackage{url}            
\usepackage{amsfonts}       
\usepackage{nicefrac}       
\usepackage{microtype}      
\usepackage{xcolor}         

\usepackage{longtable}

\input{commands}

\newcommand{\blue}[1]{\textcolor{blue}{#1}}
\newcommand{\red}[1]{\textcolor{red}{#1}}

\usepackage{lipsum}

\title{Label-Imbalanced and Group-Sensitive Classification under Overparameterization}

\author{%
Ganesh Ramachandra Kini \\
  University of California, Santa Barbara\\
  \texttt{kini@ucsb.edu}\\
\And
Orestis Paraskevas \\
University of California, Santa Barbara\\
\texttt{orestis@ucsb.edu}
\And
Samet Oymak\\
University of California, Riverside \\
\texttt{oymak@ece.ucr.edu}
\And
Christos Thrampoulidis \\
University of British Columbia \\
\texttt{cthrampo@ece.ubc.ca}
}

\begin{document}

\maketitle
\begin{abstract}
	The goal in label-imbalanced and group-sensitive classification is to optimize relevant metrics such as {balanced error} and {equal opportunity}. Classical methods, such as weighted cross-entropy, fail when training deep nets to the {terminal phase of training} (TPT), that is training {beyond zero training error}. This observation has motivated recent flurry of activity in developing heuristic alternatives following the intuitive mechanism of promoting larger margin for minorities. In contrast to previous heuristics, we follow a principled analysis explaining how different loss adjustments affect margins. First, we prove that for all linear classifiers trained in TPT, it is necessary to introduce \emph{multiplicative}, rather than \emph{additive}, logit adjustments so that the interclass margins change appropriately. To show this, we discover a connection of the multiplicative CE modification to the cost-sensitive support-vector machines. {Perhaps counterintuitively, we also find that, at the start of training, the same multiplicative weights can actually harm the minority classes.} Thus, while additive adjustments are ineffective in the TPT, we show that they can speed up convergence {by countering the initial negative effect of the multiplicative weights}. Motivated by these findings, we formulate the \emph{vector-scaling (VS) loss}, that captures existing techniques as special cases.
Moreover, we introduce a natural extension of the VS-loss to group-sensitive classification, thus treating the two common types of imbalances (label/group) in a unifying way. Importantly,
our experiments on state-of-the-art datasets are fully consistent with our theoretical insights and confirm the superior performance of our algorithms.
 Finally, for imbalanced Gaussian-mixtures data, we perform a generalization analysis, revealing tradeoffs between balanced / standard error and equal opportunity.
\end{abstract}


\input{intro}

\input{related}
\input{setup}

\input{algos}
\input{imbalanced}

\input{generalization}

\input{exp}
\input{conc}
\section*{Acknowledgments}
This work is supported by the National Science Foundation under grant Numbers CCF-2009030, by 
HDR-193464, by a CRG8 award from KAUST and by an NSERC Discovery Grant. C. Thrampoulidis would also like to acknowledge his affiliation with University of California, Santa Barbara. S. Oymak is partially supported by the NSF award
CNS-1932254 and by the NSF CAREER award CCF-2046816.
\newpage
\bibliographystyle{alpha}
\bibliography{refs}

\newpage
\appendix
\input{Organization}


\input{imba_exp}

\input{group_exp_app}

\input{more_num}

\input{imbalanced_proofs}

\input{gradient_flow}

\input{optimal_tuning}

\input{asymptotic}

\input{GS-SVM_proofs}

\end{document}

%% file: commands.tex
\usepackage{hyperref,graphicx,amsmath,amsfonts,amssymb,bm,url,breakurl,epsfig,epsf,color,MnSymbol,mathbbol,fmtcount,semtrans,caption,multirow,comment, boldline}
\usepackage{wrapfig}
\usepackage{enumitem}
\usepackage{amssymb}

\usepackage[utf8]{inputenc} 
\usepackage[T1]{fontenc}    
\usepackage{url}            
\usepackage{booktabs}       
\usepackage{amsfonts}       
\usepackage{nicefrac}       
\usepackage{microtype}      
\usepackage{dsfont}

  \usepackage{mathtools}

\usepackage{titlesec}

\usepackage{tikz}
\usepackage{pgfplots}
\usetikzlibrary{pgfplots.groupplots}

\usepackage{movie15}

\usepackage{caption}
\usepackage[bottom,hang,flushmargin]{footmisc} 

\newcommand{\tsn}[1]{{\left\vert\kern-0.25ex\left\vert\kern-0.25ex\left\vert #1 
    \right\vert\kern-0.25ex\right\vert\kern-0.25ex\right\vert}}

\definecolor{darkred}{RGB}{150,0,0}
\definecolor{darkgreen}{RGB}{0,150,0}
\definecolor{darkblue}{RGB}{0,0,200}
\hypersetup{colorlinks=true, linkcolor=darkred, citecolor=darkgreen, urlcolor=darkblue}

\newtheorem{observation}{Observation}
\newtheorem{theorem}{Theorem}

\newtheorem{lemma}{Lemma}

\newtheorem{propo}{Proposition}

\newtheorem{remark}{Remark}

\newcommand{\widesim}[2][1.5]{
  \mathrel{\overset{#2}{\scalebox{#1}[1]{$\sim$}}}
}
\newcommand{\cut}[1]{\textcolor{red}{}}
\newcommand{\W}{\mathbf{W}}
\newcommand{\M}{\mathbf{M}}
\newcommand{\Z}{\mathbf{Z}}
\newcommand{\Ub}{\mathbf{U}}
\newcommand{\G}{\mathbf{G}}

\newcommand{\Sb}{\mathbf{S}}
\newcommand{\X}{\mathbf{X}}

\newcommand{\Y}{\mathbf{Y}}
\newcommand{\Vb}{\mathbf{V}}
\newcommand{\A}{\mathbf{A}}
\newcommand{\Db}{\mathbf{D}}

\newcommand{\rb}{\mathbf{r}}
\newcommand{\fb}{\mathbf{f}}

\newcommand{\x}{\mathbf{x}}
\newcommand{\ub}{\mathbf{u}}
\newcommand{\w}{\mathbf{w}}
\newcommand{\tb}{\mathbf{t}}
\newcommand{\g}{\mathbf{g}}
\newcommand{\vb}{\mathbf{v}}

\newcommand{\eb}{\mathbf{e}}
\newcommand{\y}{\mathbf{y}}

\newcommand{\z}{\mathbf{z}}

\newcommand{\h}{\mathbf{h}}
\newcommand{\f}{\mathbf{f}}

\newcommand{\Sc}{{\mathcal{S}}}
\newcommand{\Bc}{{\mathcal{B}}}
\newcommand{\Yc}{\mathcal{Y}}
\newcommand{\Dc}{\mathcal{D}}
\newcommand{\Xc}{\mathcal{X}}

\newcommand{\Rc}{\mathcal{R}}
\newcommand{\Nn}{\mathcal{N}}
\newcommand{\Lc}{\mathcal{L}}

\newcommand{\Gc}{{\mathcal{G}}}

\newcommand{\Ec}{\mathcal{E}}

\newcommand{\beq}{\begin{equation}}
\newcommand{\eeq}{\end{equation}}
\newcommand{\bea}{\begin{align}}
\newcommand{\eea}{\end{align}}

\newcommand{\vp}{\vspace{5pt}}

\newcommand{\R}{\mathbb{R}}
\newcommand{\E}{\mathbb{E}}
\newcommand{\Pro}{\mathbb{P}}
\newcommand{\nn}{\notag}

\newcommand{\tn}[1]{\|{#1}\|_{\ell_2}}

\newcommand{\rP}{\stackrel{{P}}{\longrightarrow}}

 \newcommand{\om}{\omega}

  \newcommand{\Sigmab}{\boldsymbol\Sigma}
  \newcommand{\mub}{\boldsymbol\mu}
    \newcommand{\rhob}{\boldsymbol\rho}
  \newcommand{\eps}{\epsilon}

 \newcommand{\deltab}{\boldsymbol\delta}

\newcommand{\Rcbal}{\Rc_{\text{bal}}}
\newcommand{\Rco}{\overline\Rc}
\newcommand{\Rcbalinf}{\overline{\Rc}_{\text{bal}}}
\newcommand{\Rtrain}{\Rc_{\text{train}}}
\newcommand{\Rdeo}{\Rc_{\text{deo}}}
\newcommand{\wh}{\hat{\w}}
\newcommand{\bh}{\hat{b}}
\newcommand{\ind}[1]{\mathds{1}[#1]}

\newcommand{\bt}{\tilde b}
\newcommand{\qt}{\tilde q}

\newcommand{\rhot}{\tilde \rho}

\newcommand{\vbh}{\hat\vb}

\DeclarePairedDelimiterX{\inp}[2]{\langle}{\rangle}{#1, #2}

\newcommand{\Del}{\Delta}
\newcommand{\wt}{\widetilde\w}

\newcommand{\wtd}{\widetilde}

\newcommand{\ones}{\mathbf{1}}

\providecommand{\norm}[1]{\lVert#1\rVert}
\providecommand{\abs}[1]{\lvert#1\rvert}

\newcommand{\wte}{\widetilde}

%% file: intro.tex
\vspace{-0.2in}
\section{Introduction}\label{sec:introduction}
\vspace{-0.1in}
\subsection{Motivation and contributions}
\vspace{-0.1in}
Equitable learning in the presence of data imbalances is a classical  machine learning (ML) problem, but one with increasing importance as ML decisions are adapted in increasingly more complex applications directly involving people \cite{barocas2016big}. 
Two common types of imbalances are those appearing in \emph{label-imbalanced} and \emph{group-sensitive} classification.  In the first type, examples from a target class are heavily outnumbered by examples from the rest of the classes. The standard metric of average misclassification error is insensitive to such imbalances and among several classical alternatives
the \emph{balanced error} is a widely used metric. In the second type, the broad goal is to ensure fairness with respect to a protected underrepresented group (e.g. gender, race). While acknowledging that there is no universal fairness metric \cite{kleinberg2016inherent,friedler2016possibility}, several suggestions have been made in the literature including 
 \emph{Equal Opportunity} favoring same true positive rates across groups \cite{hardt2016equality}.

Methods for imbalanced data are broadly categorized into data- and algorithm- level ones. In the latter category, belong \emph{cost-sensitive methods} and, specifically, those that modify the {training loss} to account for varying class/group penalties. Corresponding state-of-the-art (SOTA) research is motivated by observations that classical methods, such as weighted cross-entropy (wCE) fail when training overparameterized deep nets without regularization and with train-loss minimization continuing well beyond zero train-error, in the so-called \emph{terminal phase of training (TPT)} (\cite{NC} and references therein). Intuitively, failure of wCE when trained in TPT is attributed to the failure to appropriately adjust the relative margins between different classes/groups in a way that favors minorities. To overcome this challenge, recent works have proposed a so-called {logit-adjusted (LA) loss} that modifies the cross-entropy (CE) loss by including extra \emph{additive} hyper-parameters acting on the logits \cite{cosen,TengyuMa,Menon}. Even more recently, \cite{CDT} suggested yet another modification that introduces \emph{multiplicative} hyper-parameters on the logits leading to a {class-dependent temperature (CDT) loss.} Empirically, both adjustments show performance improvements over wCE. However, it remains unclear: \emph{Do both additive and multiplicative hyper-parameters lead to  margin-adjustments favoring minority classes? If so, what are the individual mechanisms that lead to this behavior? How effective are different adjustments at each stage of training?} 

\begin{wrapfigure}{h}{0.41\textwidth}
	\centering
	\vspace{-0.05in}
	\includegraphics[width=0.38\textwidth,height=0.23\textwidth]{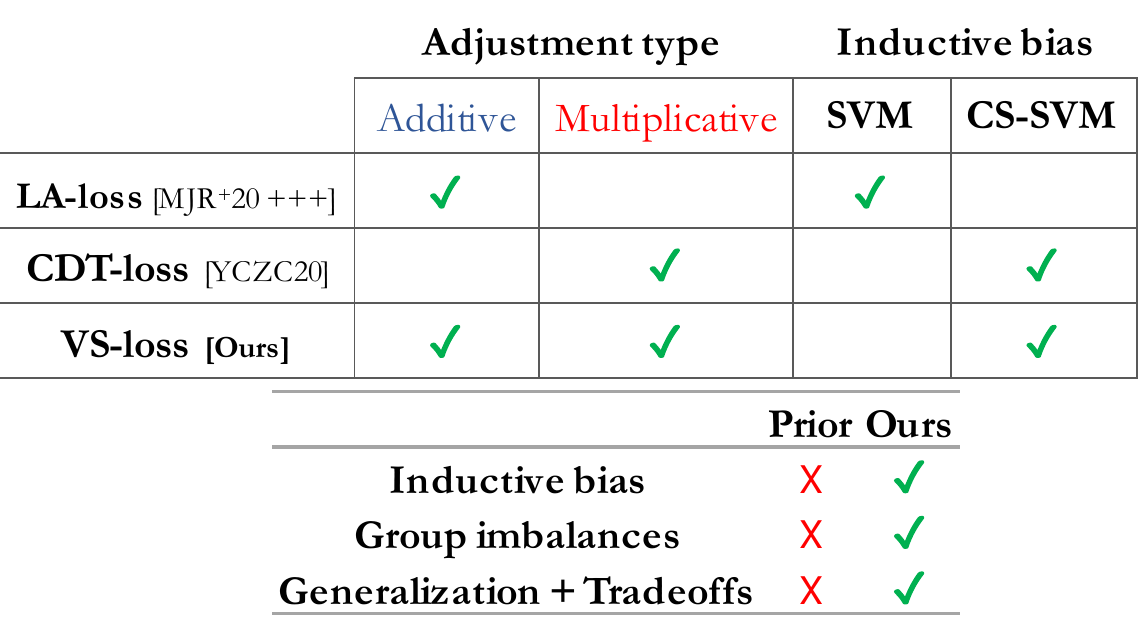} 
	\caption{Summary of contributions.}
	\label{fig:intro_table}
\end{wrapfigure}

This paper answers the above questions. Specifically, we argue that multiplicative hyper-parameters are most effective for margin adjustments in TPT, while additive parameters can be useful in the initial phase of training. Importantly, this intuition justifies our algorithmic contribution: we introduce the \emph{vector-scaling (VS) loss} that combines both types of adjustments and attains improved performance on SOTA imbalanced datasets. Finally, using the same set of tools, we extend the VS-loss to instances of group-sensitive classification. We make multiple  contributions as summarized below; see also Figure \ref{fig:intro_table}.

\noindent $\bullet$~\textbf{Explaining the distinct roles of additive/multiplicative adjustments.}~We show that when optimizing in TPT \emph{multiplicative} logit adjustments are critical. Specifically, we prove for linear models that multiplicative adjustments find classifiers that are solutions to cost-sensitive support-vector-machines (CS-SVM), which by design create larger margins for minority classes. While effective in TPT, we also find that, at the start of  training, the same adjustments can actually harm minorities. Instead, \emph{additive} adjustments can speed up convergence by countering the initial negative effect of the multiplicative ones. The analytical findings are consistent with our experiments.

\noindent $\bullet$~\textbf{An improved algorithm: VS-loss.}~Motivated by the unique roles of the two different types of adjustments, we propose the vector-scaling (VS) loss that combines the best of both worlds  and outperforms existing techniques on benchmark datasets. 

\noindent $\bullet$~\textbf{Introducing logit-adjustments for group-imbalanced data.}~We introduce a version of VS-loss tailored to {group-imbalanced datasets}, thus treating, for the first time, loss-adjustments for label and group imbalances  in a unifying way. For the latter, we propose a new algorithm combining our VS-loss with the previously proposed DRO-method to achieve state-of-the-art performance 
 in terms of both Equal Opportunity and worst-subgroup error. 

\noindent $\bullet$~\textbf{Generalization analysis / fairness tradeoffs.}~We present a sharp generalization analysis of the VS-loss on binary overparameterized Gaussian mixtures.
Our formulae are explicit in terms of data geometry, priors, parameterization ratio and hyperparameters; thus, leading to tradeoffs between standard error and fairness measures. We find that VS-loss can improve both balanced and standard error over CE. Interestingly, the optimal hyperparameters that minimize balanced error also optimize Equal Opportunity.

%% file: related.tex
\vspace{-0.1in}
\subsection{Connections to related literature}\label{sec:lit}
\vspace{-0.1in}
\noindent\textbf{CE adjustments.}~The use of wCE for imbalanced data  is rather old \cite{xie1989logit}, but it becomes ineffective under overparameterization, e.g. \cite{byrd2019effect}. This deficiency has led to the idea of \emph{additive} label-based parameters $\iota_{y}$ on the logits  \cite{cosen,TengyuMa,tan2020equalization,Menon,Wang_2018}. Specifically, \cite{Menon} proved that setting $\iota_y=\log(\pi_y)$ ($\pi_y$ denotes the prior of class $y$) leads to a \emph{Fisher consistent} loss, termed LA-loss, which outperformed other heuristics (e.g., focal loss \cite{lin2018focal}) on SOTA datasets. However, Fisher consistency is only relevant in the large sample size limit. Instead, we focus on  overparameterized models. In a recent work, \cite{CDT} proposed the CDT-loss, which instead uses \emph{multiplicative} label-based parameters $\Delta_y$ on the logits. The authors arrive at the CDT-loss as a heuristic means of compensating for the empirically observed phenomenon of that the last-layer minority features deviate between training and test instances \cite{KimKim}.  {Instead, we arrive at the CDT-loss via a different viewpoint: we show that the multiplicative weights are necessary to move decision boundaries towards majorities when training overparameterized linear models in TPT. Moreover, we argue that while additive weights are not so effective in the TPT, they can help in the initial phase of training. Our analysis sheds light on the individual roles of the two different modifications proposed in the literature and naturally motivates the VS-loss in \eqref{eq:loss_ours}. Compared to the above works we also demonstrate the successful use of VS-loss in group-imbalanced setting and show its competitive performance over alternatives in \cite{sagawa2019distributionally,hu_does,oren2019distributionally}.} Beyond CE adjustments there is active research on alternative methods to improve fairness metrics, e.g. \cite{kang2020decoupling,zhou2020bbn,liu2019largescale,ouyang2016factors}. These are orthogonal to CE adjustments and can potentially be used in conjunction. 
\\
\noindent\textbf{Relation to vector-scaling calibration.} Our naming of the VS-loss 
 is inspired by the vector scaling (VS) calibration  \cite{guo2017calibration}, a \emph{post-hoc procedure} that modifies the logits $\vb$ \emph{after training} via $\vb\rightarrow \boldsymbol{\Delta}\odot\vb+\boldsymbol{\iota}$, where $\odot$ is the Hadamard product. \cite{zhao2020role} shows that VS can improve calibration for imbalanced classes, but, in contrast to VS calibration, the multiplicative/additive scalings in our VS-loss are part of the loss and directly affect training. 
\\
\noindent\textbf{Blessings/curses of overparameterization.} 
Overparameterization acts as a catalyst for deep neural networks   \cite{nakkiran2019deep}. In terms of optimization, \cite{soudry2018implicit,oymak2019overparameterized,ji2018risk,azizan2018stochastic} show that gradient-based algorithms are \emph{implicitly biased}
towards favorable min-norm solutions. Such solutions, are then analyzed in terms of generalization showing that they can in fact lead to benign overfitting e.g. \cite{bartlett2020benign,hastie2019surprises}. While implicit bias is key to benign overfitting it may come with certain downsides. As a matter of fact, we show here that certain hyper-parameters (e.g. additive ones) can be ineffective in the interpolating regime in promoting fairness. Our argument essentially builds on characterizing the implicit bias of wCE/LA/CDT-losses.  Related to this, \cite{sagawa2020investigation}  demonstrated the ineffectiveness of $\omega_y$ in learning with groups. 
\\

%% file: setup.tex
\vspace{-0.32in}
\section{Problem setup}\label{sec:problem_setup}
\vspace{-0.15in}


\textbf{Data.}~Let training set $\{(\x_i,g_i,y_i)\}_{i=1}^n$ consisting of $n$ i.i.d. samples   from a  distribution $\Dc$ over $\Xc\times\Gc\times\Yc$;  $\Xc\subseteq\R^d$ is the input space, $\Yc=[C]:=\{1,\ldots,C\}$ the set of $C$ labels, and, $\Gc=[K]$ refers to group membership among $K\geq 1$ groups. Group-assignments are known for training data, but unknown at test time. For concreteness, we  focus here on the binary setting, i.e. $C=2$ and $\Yc=\{-1,+1\}$; we present multiclass extensions in the Experiments and in the Supplementary Material (SM).
We assume throughout that $y=+1$ is  minority class. 


\textbf{Fairness metrics.}~Given a training set we learn $f_\w:\mathcal{X}\mapsto\Yc$ parameterized by $\w\in\R^p$. For instance, linear models take the form $f_\w=\inp{\w}{h(\x)}$ for some feature representation $h:\mathcal{X}\mapsto\R^p$.
  Given a new sample $\x$, we decide class membership $\hat y = {\rm sign}(f_\w(\x)).$
The (standard) \emph{risk} or \emph{misclassification error} is 
$
\Rc:=\Pro\left\{\hat y \neq y\right\}.
$
Let $s=(y,g)$ define a subgroup for given values of $y$ and $g$. We also define the \emph{class-conditional risks}
$
\Rc_{\pm}=\Pro\left\{\hat y \neq y\,| y=\pm1 \right\},
$
and,  
the \emph{sub-group-conditional risks}
$
\Rc_{\pm,j}=\Pro\left\{\hat y \neq y\,| y=\pm1, g=j \right\},~~j\in[K].
$ 
%
The \emph{balanced error} averages the conditional risks of the two classes:  $
\Rcbal := \left({\Rc_+ + \Rc_-\,}\right)\big/{2}.
$ 
%
Assuming $K=2$ groups, Equal Opportunity requires $\Rc_{+,1}=\Rc_{+,2}$ \cite{hardt2016equality}. More generally, we consider the ({signed}) \emph{difference of equal opportunity (DEO)} 
$
\Rdeo:=\Rc_{+,1}-\Rc_{+,2}.
$ 
In our experiments, we also measure the worst-case subgroup error $\max_{(y\in{\pm1},g\in[K])} \Rc_{y,g}.$

\textbf{Terminal phase of training (TPT).}~Motivated by modern training practice, 
we assume overparameterized $f_\w$ so that $\Rtrain=\frac{1}{n}\sum_{i\in[n]}\ind{{\rm{sign}}(f_\w(\x_i))\neq y_i}$ can be driven to zero. 
Typically, training such large models continues well-beyond zero training error as the training loss is being pushed toward zero. 
As in \cite{NC}, we call this the \emph{terminal phase of training}.

%% file: algos.tex
\vspace{-0.1in}
\subsection{Algorithms}
\vspace{-0.1in}
\noindent\textbf{Cross-entropy adjustments.}~We introduce the \textbf{vector-scaling (VS) loss}, which combines \emph{both} additive and {{multiplicative}} logit adjustments, previously suggested in the literature in isolation. The following is the \textbf{binary VS-loss} for labels $y\in\{\pm1\}$, weight parameters $\omega_\pm>0$, \blue{additive} logit parameters $\blue{\iota_\pm}\in\R$, and \red{multiplicative} logit parameters $\red{\Delta_\pm}>0$:
\begin{align}\label{eq:loss_ours_bin}
\ell_{\rm VS}(y,f_\w(\x)) = \omega_y\cdot\log\left(1+e^{\blue{\iota_y}}\cdot e^{-{\red{\Delta_{y}}} yf_\w(\x)}\right).
\end{align}
For imbalanced datasets with $C>2$ classes, the \textbf{VS-loss} takes the following form:
\begin{align}\label{eq:loss_ours}
\ell_{\rm VS}(y,\fb_\w(\x)) &= -\omega_y \,\log\Big({e^{{\color{red}\Delta_y}\fb_y(\x)+\blue{\iota_y}}}\big/{\sum_{c\in[C]}e^{{\color{red}\Delta_{c}}\fb_{c}(\x)+\blue{\iota_{c}}}}\Big).
\end{align}
Here $\fb_\w:\R^d\rightarrow\R^C$ and $\fb_\w(\x)=[\fb_1(\x),\ldots,\fb_C(\x)]$ is the vector of logits. The VS-loss (Eqns. \eqref{eq:loss_ours_bin},\eqref{eq:loss_ours}) captures existing techniques as special cases by tuning accordingly the \blue{additive}/\red{multiplicative} hyperparameters. 
Specifically, we recover: (i) \textbf{weighted CE (wCE) loss} by ${\Delta_y}=1,{\iota_y}=0,\omega_y=\pi_y^{-1}$; (ii) \textbf{LA-loss} by ${\Delta_y}=1$; (iii) \textbf{CDT-loss} by ${\iota_y}=0$.

With the goal of (additionally) ensuring fairness with respect to \emph{sensitive groups}, we extend the VS-loss by introducing parameters $(\red{\Delta_{y,g}},\blue{\iota_{y,g}},\omega_{y,g})$  that depend \emph{both} on  class and group membership (specified by $y$ and $g$, respectively). Our proposed \textbf{group-sensitive VS-loss} is as follows (multiclass version can be defined accordingly):
\begin{align}\label{eq:loss_ours_bin_group}
	\ell_{\rm Group-VS}(y,g,f_\w(\x)) = \omega_{y,g}\cdot\log\big(1+e^{\blue{\iota_{y,g}}}\cdot e^{-\red{\Delta_{y,g}} yf_\w(\x)}\big).
\end{align}
 \noindent\textbf{CS-SVM.} 
For linear classifiers $f_\w(\x) = \inp{\w}{h(\x)}$ with $h:\Xc\rightarrow\R^p$, CS-SVM \cite{masnadi2010risk} solves
\begin{align}\label{eq:CS-SVM}
\hspace{-0.1in}\min_{\w}~\|\w\|_2~\text{sub. to} \begin{cases} \inp{\w}{h(\x_i)}\geq\delta &\hspace{-0.05in}, y_i=+1 \\  \inp{\w}{h(\x_i)}\leq -1 &\hspace{-0.05in}, y_i=-1 \end{cases}, i\in[n],
\end{align}
for hyper-parameter $\delta\in\R_+$ representing the ratio of margins between classes. $\delta=1$ corresponds to (standard) SVM, while tuning $\delta > 1$ (resp. $\delta<1$) favors a larger margin $\delta/\|\wh_\delta\|_2$ for the minority vs $1/\|\wh_\delta\|_2$ for the majority classes.
Thus,  $\delta\rightarrow+\infty$ (resp. $\delta\rightarrow 0$) corresponds to  the decision boundary starting right at the boundary of class $y=-1$ (resp. $y=+1$). 

\noindent\textbf{Group-sensitive SVM.}~The \emph{group-sensitive} version of CS-SVM (GS-SVM), for $K=2$ protected groups adjusts the constraints in \eqref{eq:CS-SVM} so that $y_i\inp{\w}{h(\x_i)} \geq\delta $ (or $\geq1$),  {if} $g_i=1$ (or $g_i=2.$)
$\delta>1$, GS-SVM favors larger margin for the sensitive group $g=1$. Refined versions when classes are also imbalanced modify  the constraints to $y_i\inp{\w}{\h(\x_i)}\geq \delta_{y_i,g_i}$. 
{Both CS-SVM and GS-SVM are feasible iff data are linearly separable (see SM). However, we caution that the GS-SVM  hyper-parameters are in general harder to interpret as ``margin-ratios". 

%% file: imbalanced.tex
\vspace{-0.1in}
\section{Insights on the VS-loss}\label{sec:insights_gen}
\vspace{-0.1in}
Here, we shed light on the distinct roles of the VS-loss hyper-parameters $\omega_y, \iota_y$ and $\Delta_y$.

\vspace{-0.1in}
\subsection{CDT-loss vs LA-loss: Why multiplicative weights?}\label{sec:insights}
\begin{figure}[t]
\begin{center}
	\includegraphics[width=0.85\textwidth,height=0.3\textwidth]{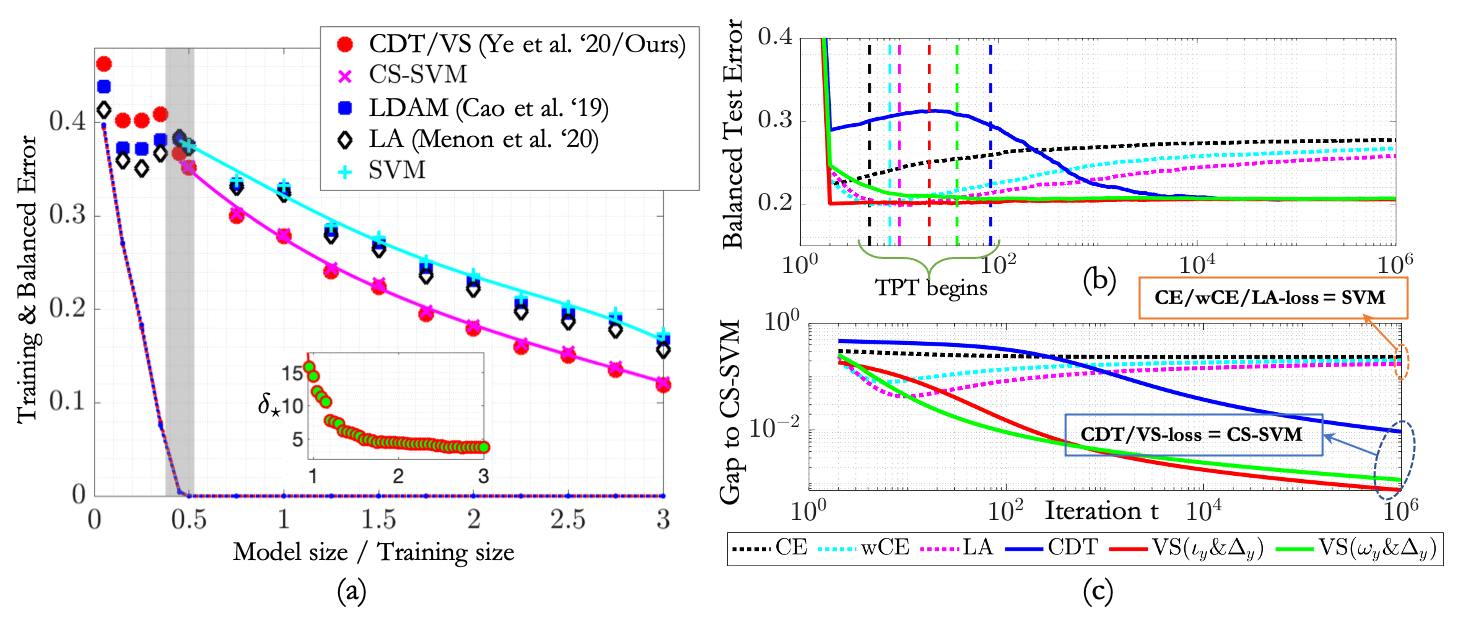} 
\end{center}
\caption{Insights on various cost-sensitive modifications of the CE-loss. \textbf{(a)} CDT has superior balanced-error performance over LA in the separable regime. 
Also, its performance matches that of CS-SVM, unlike LA matching SVM; Sec. \ref{sec:insights} for more details. Solid lines follow theory of Sec. \ref{sec:generalization}. 
\textbf{(b)} Although critical in TPT, multiplicative weights (aka CDT) can harm minority classes in initial phase of training by guiding the classifier in the wrong direction. Properly tuned additive weights (aka LA) can mitigate this effect and speed up convergence. This explains why VS can be superior compared to CDT (see Observation \ref{obs:CDT_bad}). 
Dashed lines show where TPT starts for each loss. \textbf{(c)} CDT and VS  converge to CS-SVM, unlike LA and wCE. We prove this in Theorem \ref{propo:gd}. 
}
\label{fig:mismatch_intro}
\end{figure}

\vspace{-0.1in}
We first demonstrate the unique role played by the multiplicative weights $\Delta_y$ through a motivating experiment on synthetic data in Fig. \ref{fig:mismatch_intro}. 
%
We  generated a binary Gaussian-mixture dataset of $n=100$ examples in $\R^{300}$ with data means sampled independently from the Gaussian distribution and normalized such that $\|\mub_{+1}\|_2=2\|\mub_{-1}\|_2=4$. We set prior $\pi_+=0.1$ for the minority class $+1$.  For varying model size values $p\in[5:5:50\,,\,75:25:300]$ we trained linear classifier $f_\w(x)=\inp{\w}{h(\x)}$ using only the first $p$ features, i.e. $h(\x)=\x(1:p)\in\R^p$. This allows us to investigate performance versus the parameterization ratio $\gamma=p/n.$ \footnote{Such simple models have been used in e.g. \cite{hastie2019surprises,deng2019model,chang2020provable,dhifallah2020precise,Sur14516} for analytic studies of double descent \cite{belkin2018understand,nakkiran2019deep} in terms ofclassification error. Fig. \ref{fig:mismatch_intro}(a) reveals a double descent for the balanced error.}  We train the model $\w$ using the following special cases of the VS-loss (Eqn.~\eqref{eq:loss_ours_bin}): (i) \emph{CDT-loss} with  $\Delta_+=\delta_\star^{-1}, \Delta_-=1$ ($\delta_\star>0$ is set to the value shown in the inset plot; {see SM for details}). (ii) \emph{LDAM-loss:} $\iota_+=\pi^{-1/4}, \iota_-=(1-\pi)^{-1/4}$ (special case of \emph{LA-loss} \cite{TengyuMa}). (iii) \emph{LA loss:} $\iota_+=\log\big(\frac{1-\pi}{\pi}\big), \iota_-=\log\big(\frac{\pi}{1-\pi}\big)$ (Fisher-consistent values \cite{Menon}). We ran gradient descent and averaged over $25$ independent experiments. 
The \emph{balanced error} was computed on a test set of size $10^4$ 
 and reported values are shown in red/blue/black markers. 
We also plot the training errors, which are zero for $\gamma\gtrsim0.45$. The shaded region highlights the transition to the overparameterized / separable regime.
In this regime, we continued training in the TPT. 
The plots reveal the following clear message:
\emph{The CDT-loss has better balanced-error performance compared to the LA-loss when both trained in TPT.} Moreover, they offer an intuitive explanation by uncovering a connection to max-margin classifiers: \emph{In the TPT,  
(a) LA-loss performs the same as SVM, and, (b) CDT-loss performs the same as CS-SVM.}

We formalize those empirical observations in the theorem below, which holds for arbitrary linearly separable datasets (beyond  Gaussian mixtures of the experiment). Specifically, for a sequence of norm-constrained minimizations of the VS-loss, we show that: As the norm constraint $R$ increases (thus, the problem approaches the original unconstrained loss), the direction of the constrained minimizer $\w_R$ converges to that of the CS-SVM solution $\wh_{\Delta_-/\Delta_+}$. 

\begin{theorem}[VS-loss=CS-SVM]\label{propo:gd} Fix a binary training set $\{\x_i,y_i\}_{i=1}^n$ with at least one example from each of the two classes.  Assume feature map $h(\cdot)$ such that the data are linearly separable, that is $\exists \w: y_i\w^Th(\x_i)\geq1, \forall i\in[n].$ Consider training a linear model $f_\w(\x)=\inp{\w}{h(\x)}$ by minimizing the VS-loss $\Lc_n(\w):=\sum_{i\in[n]}\ell_{\rm VS}(y_i,f_\w(\x_i))$ with $\ell_{\rm VS}$ defined in \eqref{eq:loss_ours_bin} for positive parameters $\Delta_\pm, \omega_\pm \geq 0$ and arbitrary $\iota_{\pm}$. Define the norm-constrained optimal classifier
$
\w_R=\arg\min_{\|{\w}\|_2\leq R}\Lc_n(\w).
$
Let $\wh_\delta$ be the CS-SVM solution of \eqref{eq:CS-SVM} with $\delta=\Delta_-/\Delta_+$. Then,
$
\lim_{R\rightarrow\infty}{\w_R}\big/{\|\w_R\|_2} = {\hat\w_\delta}\big/{\|\hat\w_\delta\|_2}.
$
\end{theorem}

\vspace{-0.1in}
On the one hand, the theorem makes clear that  $\omega_\pm$ and $\iota_\pm$ become ineffective in the TPT as they all result in the same SVM solutions.  
On the other hand, the multiplicative parameters $\Delta_\pm$ lead to the same classifier as that of CS-SVM, thus favoring solutions that move the classifier towards the majority class provided that $\Delta_->\Delta_+\Leftrightarrow \delta>1.$
%
%
%
The proof is given in the SM together with extensions for multiclass datasets. 
In the SM, we also strengthen Theorem \ref{propo:gd} by characterizing the \emph{implicit bias} of gradient-flow on VS-loss. Finally, we show that group-sensitive VS-loss with $\Delta_{y,g}=\Delta_g$ converges to the corresponding GS-SVM. 

\begin{remark}
Thm \ref{propo:gd} is reminiscent of Thm. 2.1 in \cite{rosset2003margin} who 
 showed for a regularized ERM with CE-loss that when the regularization parameter vanishes, the normalized solution converges to the SVM classifier. Our result connects nicely to \cite{rosset2003margin} extending their theory to VS-loss / CS-SVM, as well as, to the group-case.
In a similar way, our result on the implicit bias of gradient-flow on the VS-loss connects to more recent works \cite{soudry2018implicit,ji2018risk} that pioneered corresponding results for CE-loss. Although related, our results on the properties of the VS-loss are not obtained as special cases of these existing works. As a final remark, in Fig. \ref{fig:mismatch_intro}(b,c) we kept constant learning rate $0.1$. Significantly faster convergence is observed with normalized GD schemes 
\cite{nacson2019convergence,ji2021characterizing}; see the SM for a detailed numerical study. We also note that Thm. \ref{propo:gd}   gives a modern interpretation to the CS-SVM via the lens of implicit bias theory.

\end{remark}


\vspace{-0.1in}
\subsection{VS-loss: Best of two worlds}\label{sec:best}
\vspace{-0.1in}
We have shown that multiplicative weights $\Delta_\pm$ are responsible for good balanced accuracy in the TPT. Here, we show that, at the initial phase of training, the same multiplicative weights can actually harm the minority classes. The following observation supports this claim. 
\begin{observation}\label{obs:CDT_bad}
Assume $f_\w(x)=0$ at initialization. Then, the gradients of CDT-loss with multiplicative logit factors $\Delta_y$ are \underline{identical} to the gradients of wCE-loss with weights $\omega_y=\Delta_y.$ Thus, we conclude the following where say $y=+1$ is minority.  On the one hand, wCE, which typically sets $\omega_{\rm +}>\omega_{\rm -}$ (e.g., $\omega_y=1/\pi_y$), {helps minority examples by weighing down the loss over majority}. On the other hand, the CDT-loss requires {the reverse direction} ${\Delta_{+}<\Delta_{-}}$ as per Theorem \ref{propo:gd}, thus initially it {guides} the classifier in the wrong direction {to penalize} minorities.
%
\end{observation}

To see why the above is true note that for $f_\w(x)=\inp{\w}{h(\x)}$ the gradient of VS-loss is $\nabla_\w\ell_{\rm VS}(y,f_\w(\x))=-\omega_y \Delta_y\,\sigma\big(-\Delta_y y f_\w(x)+\iota_y\big)\cdot {y h(\x)}$ where $\sigma(t)=(1+\exp(-t))^{-1}$ is the sigmoid function. It is then clear that at $f_\w(\x)=0$, the logit factor $\Delta_y$ plays the same role as the weight $\omega_y$. From Theorem \ref{propo:gd}, we know that pushing the margin towards majorities (which favors balancing the conditional errors) requires $\Delta_+<\Delta_-$. Thus, gradient of minorities becomes smaller, initially pushing the optimization in the wrong direction. Now, we turn our focus at the impact of $\iota_y$'s at the start of training. Noting that $\sigma(\cdot)$ is increasing function, we see that setting $\iota_+>\iota_-$ increases the gradient norm for minorities. This leads us to a second observation:
\emph{By properly tuning the additive logit adjustments $\iota_y$ we can counter the initial negative effect of the multiplicative adjustment, thus speeding up training.}
The observations above naturally motivated us to formulate the VS-loss in Eqn. \eqref{eq:loss_ours} bringing together the best of two worlds: the $\Delta_y$'s that play a critical role in the TPT and the $\iota_y$'s that compensate for the harmful effect of the $\Delta_y$'s in the beginning of training.

Figure \ref{fig:mismatch_intro}(b,c) illustrate the discussion above. In the  binary linear classification setting of Fig. \ref{fig:mismatch_intro}(a), we investigate the effect of the additive adjustments on the training dynamics. Specifically, we trained using gradient descent: (i) \emph{CE}; (ii) \emph{wCE} with $\omega_y = 1/\pi_y$; (iii) \emph{LA-loss} with $\iota_y=\log(1/\pi_{y})$;  (iv) \emph{CDT-loss} with $\Delta_+=\delta_\star^{-1}, \Delta_-=1$; (v) \emph{VS-loss} with $\Delta_+=\delta_\star^{-1}, \Delta_-=1$, $\iota_y=\log(1/\pi_{y})$ and $\omega_y=1$; (vi) \emph{VS-loss} with same $\Delta$'s, $\iota_y=0$ and $\omega_y=1/\pi_y.$
Figures \ref{fig:mismatch_intro}(b) and (c) plot balanced test error $\Rcbal$ and angle-gap to CS-SVM solution as a function of iteration number for each algorithm. The vertical dashed lines mark the iteration after which training error stays zero and we enter the TPT. Observe in Fig. \ref{fig:mismatch_intro}(c) that CDT/VS-losses, both converge to the CS-SVM solution as TPT progresses verifying Theorem \ref{propo:gd}. This also results in lowest test error in the TPT in Fig. \ref{fig:mismatch_intro}(b). However,  compared to CDT-loss, the VS-loss enters faster in the TPT  and converges orders of magnitude faster to small values of $\Rcbal$. Note in Fig. \ref{fig:mismatch_intro}(c) that this behavior is correlated with the speed at which the two losses converge to CS-SVM. Following the discussion above, we attribute this favorable behavior during the initial phase of training to the inclusion of the $\iota_y$'s. This is also supported by Fig. \ref{fig:mismatch_intro}(c) as we see that LA-loss (but also wCE) achieves significantly better values of $\Rcbal$ at the first stage of training compared to CDT-loss. In Sec. \ref{sec:exp_label} we provide deep-net experiments on an imbalanced CIFAR-10 dataset that further support these findings.

%% file: generalization.tex
\vspace{-0.1in}
\section{Generalization analysis and fairness tradeoffs}\label{sec:generalization}
\vspace{-0.1in}
Our results in the previous section regarding VS-loss/CS-SVM hold for arbitrary linearly-separable training datasets. Here, under additional distributional assumptions, we establish a sharp asymptotic theory for VS-loss/CS-SVM and their group-sensitive counterparts. 

\noindent\textbf{Data model.}~We study binary Gaussian-mixture generative models (GMM) for the data distribution $\Dc$. For the label $y\in\{\pm1\}$, let $\pi:=\Pro\{y=+1\}.$ Group membership is decided conditionally on the label such that $\forall j\in[K]:$ $\Pro\{g=j|y=\pm1\}=p_{\pm,j}$, with $\sum_{j\in[K]}p_{+,j}=\sum_{j\in[K]}p_{-,j}=1$. Finally, the feature conditional given label $y$ and group $g$ is a multivariate Gaussian of mean $\mub_{y,g}\in\R^d$ and covariance $\Sigmab$, i.e. 
$\x\big|(y,g)\,\widesim{} \Nn(\mub_{y,g},\Sigmab).
$
Specifically for  \emph{label-imbalances}, we let $K=1$ and $
\x\big|y\,\widesim{} \Nn(\mub_{y},\mathbf{I}_d)$ (see SM for $\Sigmab\neq \mathbf{I}_d$). For \emph{group-imbalances}, we focus on two groups with $p_{+,1}=p_{-,1}=p<1-p=p_{+,2}=p_{-,2},\,j=1,2$ and $\x\,|\,(y,g) \sim \Nn(y\mub_{g},\mathbf{I}_d)$. In both cases,  $\M$ denotes the matrix of means, i.e. $\M=\begin{bmatrix} \mub_{+} & \mub_{-}\end{bmatrix}$ and $\M=\begin{bmatrix} \mub_{1} & \mub_{2}\end{bmatrix}$, respectively. Also, consider the eigen-decomposition:
$
\M^T\M = \Vb\Sb^2\Vb^T,~~ \Sb\succ \mathbf{0}_{r\times r},\Vb\in\R^{2\times r}, r\in\{1,2\},
$
with $\Sb$ an $r\times r$ diagonal positive-definite matrix and $\Vb$ an orthonormal matrix obeying  $\Vb^T\Vb=\mathbf{I}_r$. We study linear classifiers with $h(\x)=\x$. 

\noindent\textbf{Learning regime.}~
We focus on the separable regime. For the models above, linear separability undergoes a sharp phase-transition as $d,n\rightarrow\infty$ at a proportional rate $\gamma=\frac{d}{n}$. That is, there exists threshold $\gamma_\star:=\gamma_\star(\Vb,\Sb,\pi)\leq1/2$ 
for  the label-case, such
that data are linearly separable with probability approaching one provided that $\gamma>\gamma_\star$ (accordingly for the group-case) \cite{candes2020phase,montanari2019generalization,deng2019model,kammoun2020precise,gkct_icassp}. See SM for formal statements and explicit definitions. 

\noindent\textbf{Analysis of CS/GS-SVM.}~We use $\rP$ to denote convergence in probability and $Q(\cdot)$ the standard normal tail.  We let $(x)_-:=\min\{x,0\}$; $\ind{\Ec}$ the indicator function of  event $\Ec$; $\Bc_2^r$ the unit ball in $\R^r$; and, $\eb_1=[1,0]^T, \eb_2= [0,1]^T$  standard basis vectors in $\R^2$.
We further need the following definitions. Let random variables as follows:
$
G\sim \Nn(0,1)$, $Y$ symmetric Bernoulli with $\Pro\{Y=+1\}=\pi$, 
$E_Y=\eb_1\ind{Y=1} - \eb_2\ind{Y=-1}$ and 
$\Delta_Y=\delta\cdot\ind{Y=+1} +\ind{Y=-1}$, for $\delta>0$.
With these define key function $\eta_\delta:\R_{\geq0}\times\Bc_2^{r}\times\R\rightarrow\R$ as 
$\eta_\delta(q,\rhob,b):=\E\Big[\big(G+E_Y^T\Vb\Sb\rhob + \frac{b Y-\Delta_Y}{q}\big)_{-}^2\big] \nn- (1-\|\rhob\|_2^2)\gamma.
$ Finally, define $(q_\delta,\rhob_\delta,b_\delta)$ as the \emph{unique} triplet (see SM for proof) satisfying $\eta_\delta(q_\delta,\rhob_\delta,b_\delta)= 0$ and 
$(\rhob_\delta,b_\delta):=\arg\min_{\substack{\|\rhob\|_2\leq 1, b\in\R}}\eta_\delta(q_\delta,\rho,b).$ Note that these triplets can be easily computed numerically for given values of $\gamma, \delta, \pi, p$ and means' Gramian $\M^T\M = \Vb\Sb^2\Vb^T.$

\begin{theorem}[Balanced error of CS-SVM]\label{thm:main_imbalance}
Let GMM data with label imbalances and learning regime as described above. 
Consider the CS-SVM classifier in \eqref{eq:CS-SVM} with $h(\x)=\x$, intercept $b$ (i.e. constraints $\inp{\x}{\w}+b\geq\{\delta \text{ or } 1\}$ in \eqref{eq:CS-SVM}) and fixed margin-ratio $\delta>0.$ 
Define 
$\Rco_+:=Q\left(\eb_1^T\Vb\Sb\rhob_\delta+b_\delta/q_\delta\right)$ and $\Rco_-:=Q\left(-\eb_2^T\Vb\Sb\rhob_\delta-b_\delta/q_\delta\right).$
Then, as $n,d\rightarrow\infty$ with $d/n=\gamma>\gamma_\star$, it holds that 
$\Rc_+\rP\Rco_+$ and $\Rc_-\rP\Rco_-$.
In particular, 
$\Rcbal\rP\Rcbalinf:=(\Rco_++\Rco_-)\big/2.$
\end{theorem}
The theorem further shows  
$(\|\hat\w_\delta\|_2, \frac{\wh_\delta^T\mub_+}{\|\wh_\delta\|_2}, \frac{\wh_\delta^T\mub_-}{\|\wh_\delta\|_2} , \hat b_\delta) \rP (q_\delta,\eb_1^T\Vb\Sb\rhob_\delta,\eb_2^T\Vb\Sb\rhob_\delta, b_\delta).
$
Thus, $b_\delta$ is the asymptotic the intercept, $q_\delta^{-1}$ is the asymptotic classifier's margin ${1}/{\|\wh_\delta\|_2}$ to the majority, and $\rhob_\delta$ determines the asymptotic alignment of the classifier with the class mean. The proof uses the convex Gaussian min-max theorem (CGMT) framework \cite{StojLAS,thrampoulidis2015regularized}; see SM for background, the proof, as well as, (a) simpler expressions when the means are antipodal ($\pm \mub$) and (b) extensions to general covariance model ($\Sigmab\neq\mathbf{I}$). The experiment (solid lines) in Figure \ref{fig:mismatch_intro}(a) validates the theorem's predictions. 
Also, in the SM, we characterize the DEO of GS-SVM for GMM data.
Although similar in nature, that characterization differs to Thm. \ref{thm:main_imbalance} since each class is now itself a Gaussian mixture as described in the model above.

\begin{figure}[t]
\begin{subfigure}[b]{0.55\textwidth}
			\centering
\includegraphics[width=0.8\textwidth,height=0.42\textwidth]{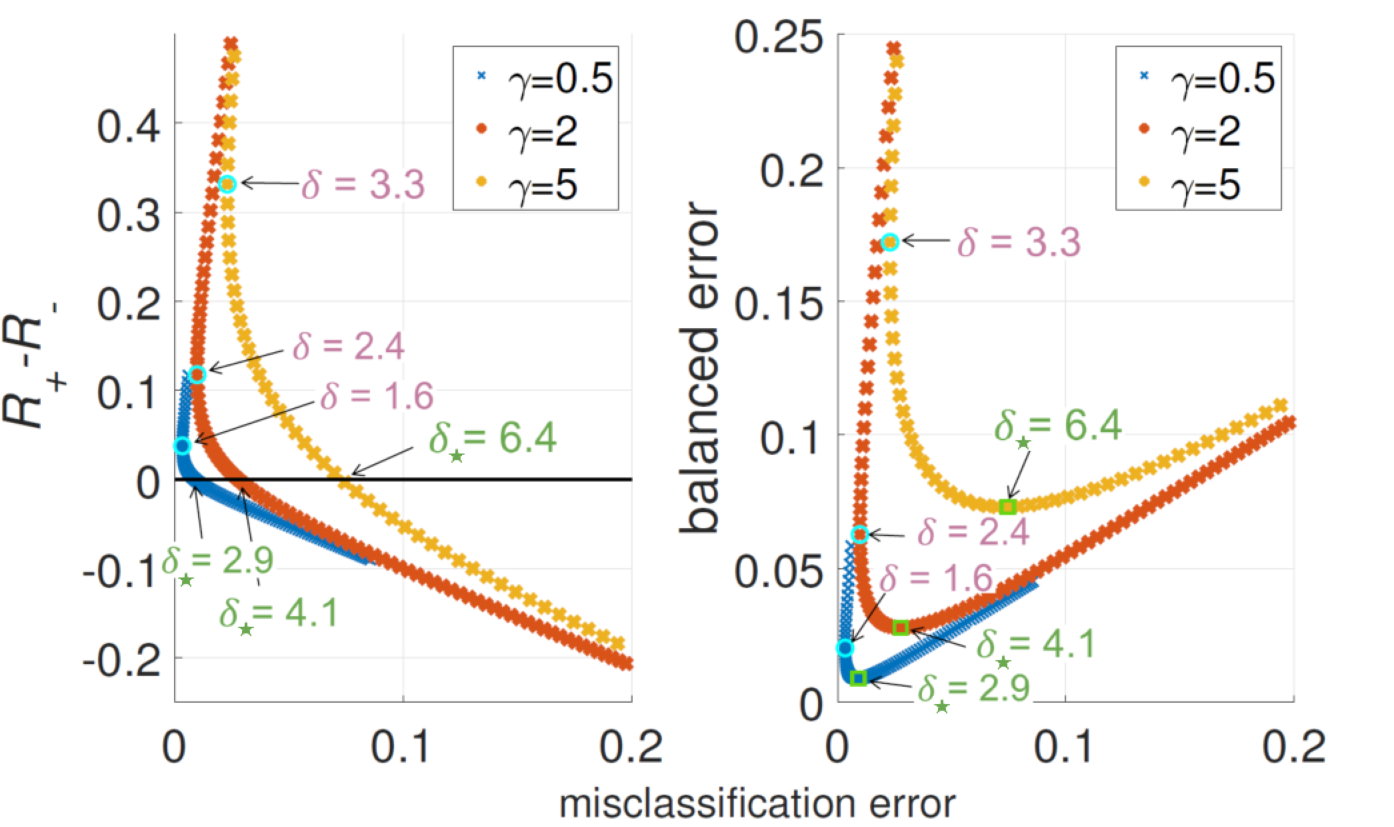}
			\caption{}
		\end{subfigure}
\begin{subfigure}[b]{0.45\textwidth}
			\centering
\includegraphics[width=0.8\textwidth,height=0.48\textwidth]{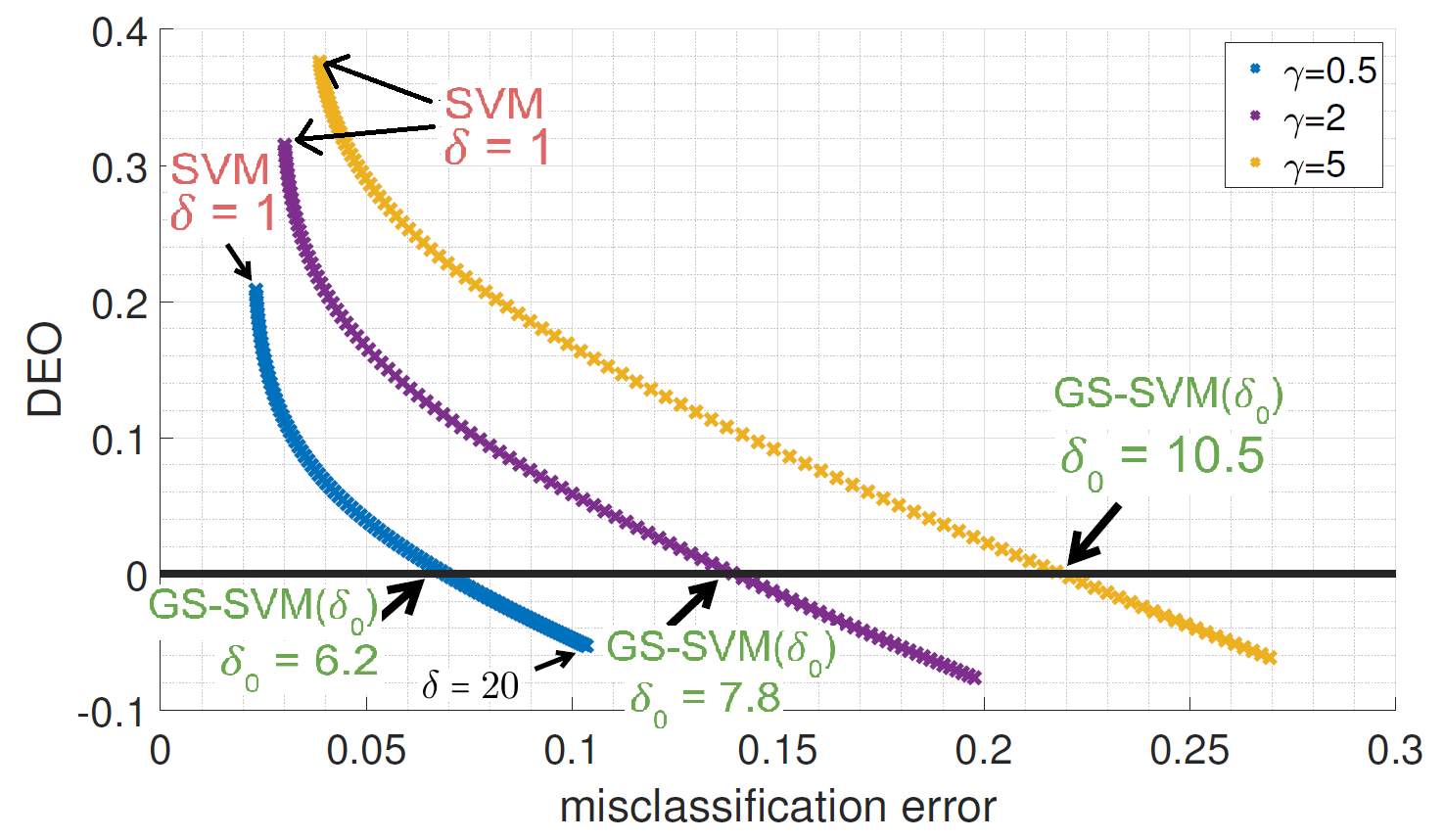}
			\caption{}
		\end{subfigure}
	\caption{Fairness tradeoffs between classification error and error-imbalance/balanced-error/DEO on GMM data achieved by \textbf{(a)} CS-SVM for class prior  $\pi=0.05$ and \textbf{(b)} GS-SVM  for group prior $p=0.05$, as a function of the margin-ratio hyperparameter $\delta\geq 1$ and for various values of overparameterization $\gamma$.  Plots in (a) are  generated using our sharp predictions in Theorem \ref{thm:main_imbalance}. Plots in (b) use corresponding result for GS-SVM given in the SM.
	 See text for interpretations.
	} 
	\label{fig:tradeoffs}	
\end{figure}

\noindent\textbf{Fairness tradeoffs.}~The theory above allow us to study tradeoffs between misclassification / balanced error / DEO in Fig. \ref{fig:tradeoffs}. Fig. \ref{fig:tradeoffs}(a) focuses on label imbalances. We make the following  observations. (1) The optimal value $\delta_\star$ minimizing $\Rcbal$ also achieves perfect balancing between the conditional errors of the two classes, that is $\Rc_+=\Rc_-=Q(\frac{\ell_-+\ell_+}{2}).$ We prove this interesting property in the SM 
by deriving an explicit formula for $\delta_\star$ that only requires computing the triplet $(q_1,\rhob_1,b_1)$ for $\delta=1$ corresponding to the standard SVM. Such {closed-form} formula is rather unexpected given the seemingly involved nonlinear dependency of $\Rcbal$ on $\delta$ in Thm. \ref{thm:main_imbalance}. In the SM, we also use this formula to formulate a theory-inspired heuristic for hyperparameter tuning, which shows good empirical performance on simple datasets such as imbalanced MNIST. (2) The value of $\delta$ minimizing standard error $\Rc$ (shown in magenta) is not equal to $1$, hence CS-SVM also improves $\Rc$ (not only $\Rcbal$). In Fig. \ref{fig:tradeoffs}(b), we investigate the effect of $\delta$ and the improvement of  GS-SVM over SVM. The largest DEO and smallest misclassification error are achieved by the SVM ($\delta=1$). But, with increasing $\delta$, misclassification error is traded-off for reduction in absolute value of DEO. Interestingly, for some $\delta_0=\delta_0(\gamma)$ (with value increasing with $\gamma$) GS-SVM guarantees Equal Opportunity (EO) $\Rc_{\rm deo}=0$ (without explicitly imposing such constraints as in \cite{olfat2018spectral,donini2018empirical}).

%% file: exp.tex
\vspace{-0.1in}
\section{Experiments}\label{sec:experiments}
\vspace{-0.1in}
We show experimental results further justifying theoretical findings. {(Code available in \cite{code}).} 
\vspace{-0.1in}
\subsection{Label-imbalanced data}\label{sec:exp_label}
\begin{figure}[h]
	\begin{center}
		\begin{subfigure}[b]{0.49\textwidth}
			\centering
			\includegraphics[width=0.5\textwidth]{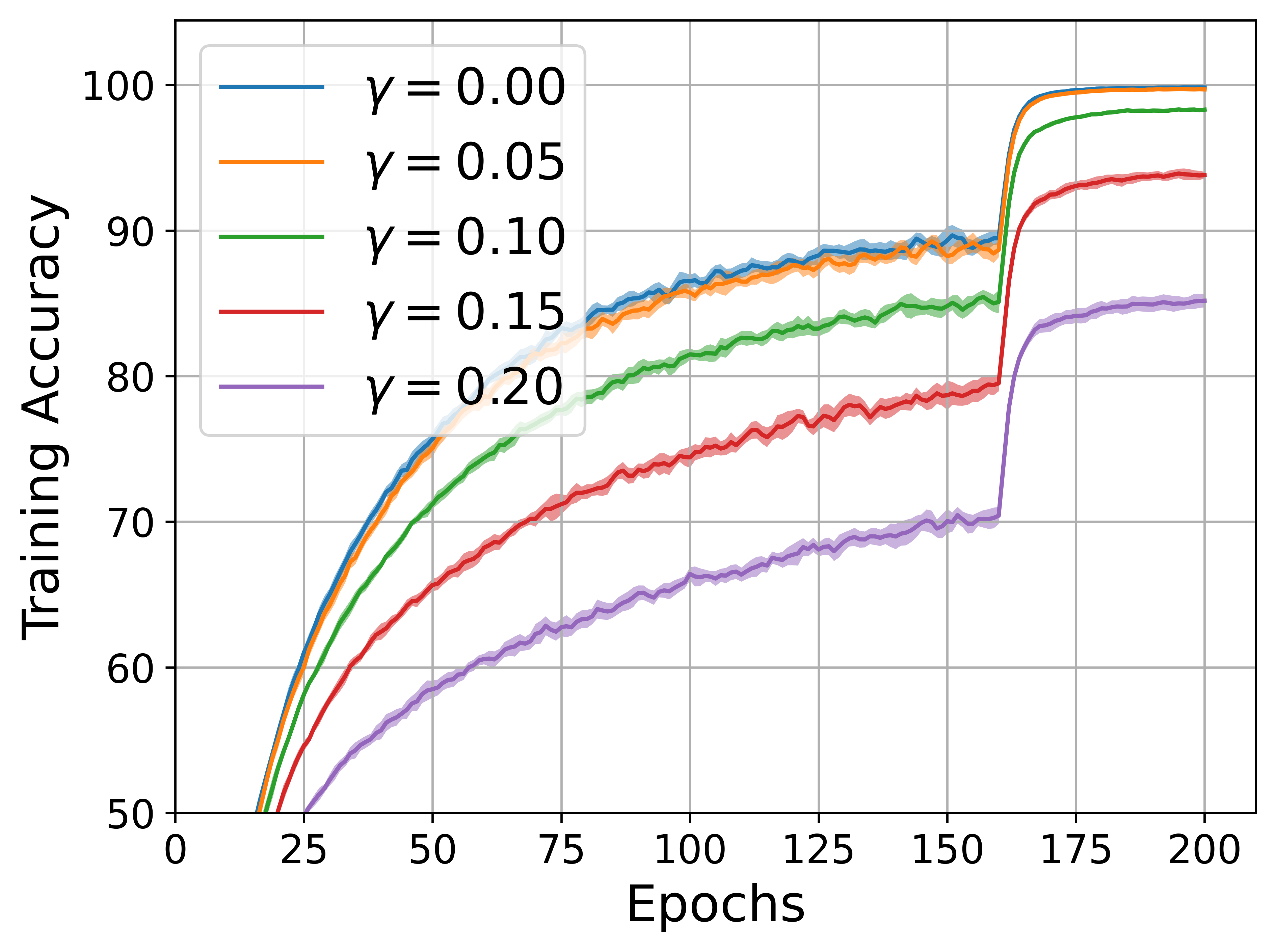} 
			\caption{$\Delta_y$'s (parameterized by $\gamma$) can hurt training.}
		\end{subfigure}
		\begin{subfigure}[b]{0.49\textwidth}
			\centering
			\includegraphics[width=0.5\textwidth]{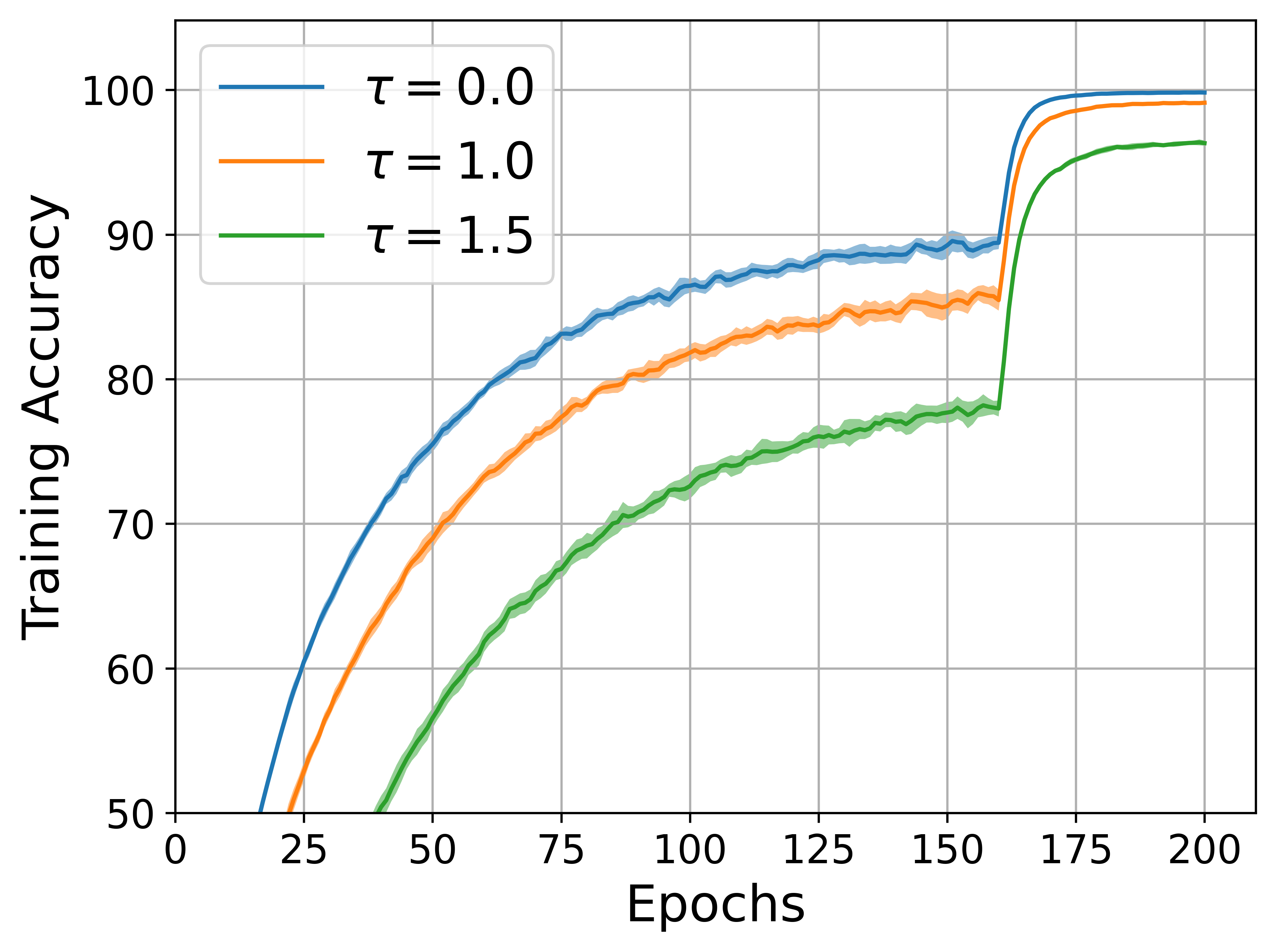} 
			\caption{LA trains easier than CDT.}
		\end{subfigure}
		\\
		\begin{subfigure}[b]{\textwidth}
				\includegraphics[width=\textwidth]{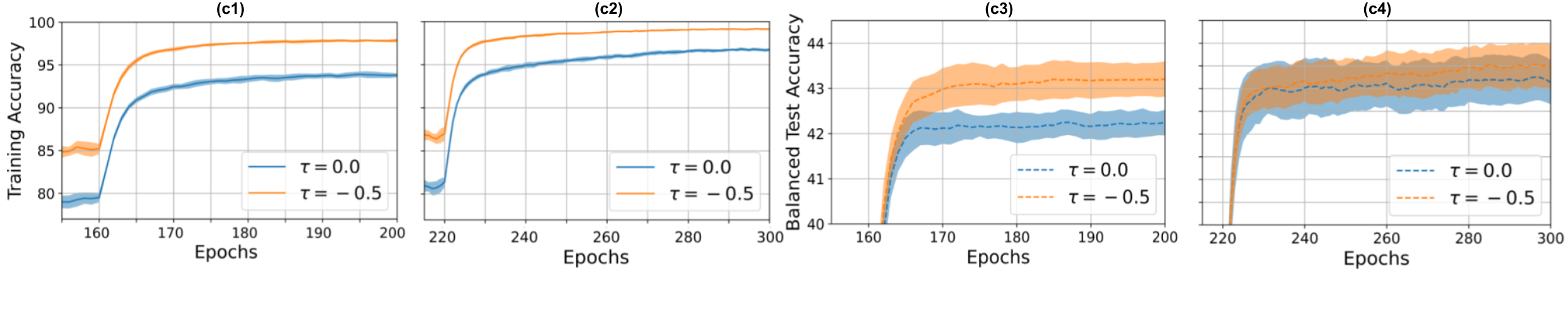} 
				\caption{$\iota_y$'s mitigate the effect of $\Delta_y$'s (c1,c2), but $\Delta_y's$ dominate TPT performance (c3,c4).}
		\end{subfigure}
	\end{center}
	\caption{ Experiments on CIFAR10 with Long-tailed LT-100 imbalance demonstrating the effects of additive/multiplicative parameters at different phases of training. All results are averaged over 5 runs and shaded regions indicate the $95\%$ confidence intervals. See text for details and interpretations.}
	\label{fig:tau_gamma_effect_on_training_main}
\end{figure}

\vspace{-0.1in}
Our first experiment (Table \ref{table:CIFAR})  shows that \emph{non-trivial combinations} of additive/multiplicative adjustments 
can improve balanced accuracy over \emph{individual} ones. Our second experiment (Fig. \ref{fig:tau_gamma_effect_on_training_main}) validates the theory of Sec. \ref{sec:insights_gen} by examining how these adjustments affect training.




\noindent\textbf{Datasets.} 
Table \ref{table:CIFAR} evaluates LA/CDT/VS-losses on imbalanced instances of CIFAR-10/100.
Following \cite{TengyuMa}, we consider: (1) \textit{STEP} imbalance,  
reducing the sample size of half of the classes to a fixed number. (2) Long-tailed (\textit{LT}) imbalance, 
which exponentially decreases the number of training images across different classes. We set an imbalance ratio $N_{\max}/ N_{\min}=100$, where $N_{\max}=\max_y{N_y}, N_{\min}=\min_y{N_y}$ and $N_y$ are  sample sizes of class $y$. For consistency with \cite{he2016deep, TengyuMa, Menon, CDT} we keep a balanced test set and in addition to evaluating our models on it, we treat it as our validation set and use it to tune our hyperparameters. More sophisticated tuning strategies (perhaps using bi-level optimization) are deferred to future work. 
We use data-augmentation exactly as in \cite{he2016deep, TengyuMa, Menon, CDT}.
See SM for more implementation details.

\noindent\textbf{Model and Baselines.} 
We compare the following:  
\textit{(1) CE-loss}. \textit{(2) Re-Sampling} that includes each data point  in the batch with probability ${\pi_y}^{-1}$. \textit{(3) wCE} with weights $\omega_y = {\pi_y}^{-1}$. \textit{(4) LDAM-loss} \cite{TengyuMa}, special case of LA-loss where $\iota_y = \frac{1}{2} ({N_{\min}}/{N_y})^{1/4}$ is \emph{subtracted} from the logits.
\begin{wraptable}{r}{0.7\textwidth}
	\caption{Top-1 accuracy results on balanced validation set ($\%$). 
	}
	\label{table:CIFAR}
	\begin{center}\resizebox{0.7\textwidth}{!}{
			\begin{tabular}{lllllllll} \hline
				\textbf{Dataset}         & \multicolumn{2}{c}{\textbf{CIFAR 10}} & \multicolumn{2}{c}{\textbf{CIFAR 100}} \\ \hline
				\textbf{Imbalance Profile}  & \multicolumn{1}{c}{\textbf{LT-100}} & \multicolumn{1}{c}{\textbf{STEP-100}} & \multicolumn{1}{c}{\textbf{LT-100}} & \multicolumn{1}{c}{\textbf{STEP-100}} \\ \hline \hline
				CE                       						& $71.94 \pm 0.38$ & $62.69 \pm 0.50$ & $38.82 \pm 0.69$  & $39.49 \pm 0.16$  \\ 
				Re-Sampling              					& 71.2  & 65.0  & 34.7  & 38.4 \\ 
				wCE             							& 72.6  & 67.3  & 40.5  & 40.1 \\ \hline
				LDAM  \cite{TengyuMa}.                 			& 73.35 & 66.58 & 39.60  & 39.58 \\ 
				LDAM-DRW  \cite{TengyuMa}               		& 77.03 & 76.92  & 42.04 & 45.36 \\ 
				LA ($\tau=\tau^{*}$) \cite{Menon}                       	& $80.81 \pm 0.30$ & $78.23 \pm 0.52$ & $42.87 \pm 0.32$ & $45.69 \pm 0.27$ \\
				CDT ($\gamma=\gamma^{*}$)  \cite{CDT}         	& $79.55 \pm 0.35$ & $73.26 \pm 0.29$ & $42.57 \pm 0.32$ & $44.12 \pm 0.17$  \\ \hline \hline
				VS ($\tau=\tau^{*}, \gamma=\gamma^{*}$)         & $\textbf{80.82} \pm 0.37 $ & $\textbf{79.10} \pm 0.66$ & $\textbf{43.52} \pm 0.46$ & $\textbf{46.53} \pm 0.17$  \\ 
		\end{tabular}}
	\end{center}
\end{wraptable}
\textit{(5) LDAM-DRW} \cite{TengyuMa}, combining LDAM with deferred re-weighting. \textit{(6) LA-loss} \cite{Menon}, with the Fisher-consistent parametric choice $\iota_y = \tau \log(\pi_y)$. \textit{(7) CDT-loss} \cite{CDT}, with $\Delta_y = ({N_{y}}/{N_{max}})^{\gamma}$.  \textit{(8) VS-loss}, with combined hyperparameters $\iota_y = \tau \log(\pi_y)$ and $\Delta_y = ({N_{y}}/{N_{\max}})^{\gamma}$, parameterized by $\tau, \gamma>0$ respectively \footnote{{Here, the hyperparameter $\gamma$ is used with some abuse of notation and is important to not be confused with the parameterization ratio in the linear models in Sec. \ref{sec:insights_gen} and \ref{sec:generalization}. We have opted to use the same notation as in \cite{CDT} to ease direct comparisons of experimental findings.}}. 
The works introducing (5)-(7) above, all trained for a different number of epochs,  with dissimilar regularization and learning rate schedules. For consistency, we follow the training setting in \cite{TengyuMa}. Thus, for LDAM we adapt results reported by \cite{TengyuMa}, but for LA and CDT, we reproduce our own in that setting. Finally, for a fair comparison we ran LA-loss for optimized $\tau=\tau^*$ (rather than $\tau=1$ in \cite{Menon}).

\noindent\textbf{VS-loss balanced accuracy.} Table \ref{table:CIFAR} shows Top-1 accuracy on balanced validation set (averaged over 5 runs). 
We use a grid to pick the best $\tau$ / $\gamma$ / ($\tau,\gamma$)-pair for the LA / CDT / VS losses 
on the validation set. 
Since VS includes LA and CDT as special cases (corresponding to $\gamma=0$ and $\tau=0$ respectively), we expect that it is at least as good as the latter over our hyper-parameter grid search. 
We find that the optimal $(\tau^*,\gamma^*)$-pairs correspond to non-trivial combinations of each individual parameter. 
Thus, VS-loss has better balanced accucy as shown in the table.
See SM for optimal hyperparameters choices.
\vspace*{-0.5cm}

\noindent\textbf{How hyperparameters affect training?}  We perform three experiments. \textbf{(a)} Figure \ref{fig:tau_gamma_effect_on_training_main}(a) shows that larger values of hyperparameter $\gamma$ (corresponding to more dispersed $\Delta_y$'s between classes) hurt training performance and delay entering to TPT. Complementary Figures \ref{fig:tau_gamma_effect_on_training_main}(c1,c2) show that eventually, if we train longer, then, train accuracy approaches 100\%. These findings are in line with Observation \ref{obs:CDT_bad} in Sec.  \ref{sec:best}. \textbf{(b)} Figure \ref{fig:tau_gamma_effect_on_training_main}(b) shows  training accuracy of LA-loss for changing hyperparameter $\tau$ controlling additive adjustments. On the one hand, increasing values of $\tau$ delay training accuracy to reach 100\%. On the other hand, when compared to the effect of $\Delta_y$'s in  Fig. \ref{fig:tau_gamma_effect_on_training_main}(a), we observe that the impact of additive adjustments on training is significantly milder than that of multiplicative adjustments. Thus, LA trains easier than CDT. \textbf{(c)} Figure \ref{fig:tau_gamma_effect_on_training_main}(c) shows train and balanced accuracies for (i) CDT-loss in blue: $\tau=0$, $\gamma=0.15$, (ii) VS-loss in orange: $\tau=-0.5$, $\gamma=0.15$. In Fig. \ref{fig:tau_gamma_effect_on_training_main}(c1,c3) we trained for $200$ epochs, while in Fig. \ref{fig:tau_gamma_effect_on_training_main}(c2,c4) we trained for $300$ epochs. For $\gamma=0.15$, CDT-loss does \emph{not} reach good training accuracy within 200 epochs ($\sim93$\% at epoch 200 in Fig. \ref{fig:tau_gamma_effect_on_training_main}(c1)), but the addition of $\iota_y$'s with $\tau=-0.5$ mitigates this effect achieving improved $\sim97$\% accuracy at 200 epochs. This also translates to balanced test accuracy: VS-loss has better accuracy at the end of training in Fig. \ref{fig:tau_gamma_effect_on_training_main}(c3). Yet, CDT-loss has not yet entered the interpolating regime in this case. So, we ask: What changes if we train longer so that both CDT and VS loss get (closer) to interpolation. In Fig. \ref{fig:tau_gamma_effect_on_training_main}(c2) train accuracy of both algorithms increases when training continues to 300 epochs. Again, thanks to the $\iota_y$'s VS-loss trains faster. However, note in Figure  \ref{fig:tau_gamma_effect_on_training_main}(c4) that the balanced accuracies of the two methods are now very close to each other. Thus, in the interpolating regime what dominates the performance are the multiplicative adjustments which are same for both losses. This is in agreement with the finding of Theorem \ref{propo:gd} and the synthetic experiment in Fig. \ref{fig:mismatch_intro}(b,c). 

\input{group_exp}

%% file: group_exp.tex
\vspace{-0.2cm}
\subsection{Group-sensitive data} \label{sec:GS_exp_main}
\vspace{-0.2cm}
The message of our experiments on group-imbalanced datasets is three-fold. (1) We demonstrate the practical relevance of logit-adjusted CE modifications to settings with imbalances at the level of (sub)groups. (2) We show that such methods are competitive to alternative state-of-the-art; specifically, distributionally robust optimization (DRO) algorithms. (3) We propose combining logit-adjustments with DRO methods for even superior performance. 

\noindent\textbf{Dataset.}
We study a setting with spurious correlations ---strong associations between label and background in image classification--- which can be cast as a subgroup-sensitive classification problem 
\cite{sagawa2019distributionally}. 
We consider the Waterbirds dataset 
\cite{sagawa2019distributionally}.
The goal is to classify images as either `waterbirds' or `landbirds', while their background ---either `water' or `land'--- can be spuriously correlated with the type of birds. 
Formally, each example has label $y\in\Yc=\{\pm1\}\equiv\{\text{waterbird},\text{landbird}\}$ and  belongs to a group 
$g\in\Gc=\{\pm1\}\equiv\{\text{water},\text{land}\}$. Let then $s=(y,g) \in \{\pm1\}\times\{\pm1\}$ be the four sub-groups with $(+1,-1)$, $(-1,+1)$ being minorities (specifically, $\hat p_{+1,+1}=0.22,\hat p_{+1,-1}=0.012,\hat p_{-1,+1}=0.038$ and $\hat p_{-1,-1}=0.73.$). Denote $N_{s}$ the number of training examples belonging to sub-group $s$ and  $N_{\text{max}}:=\max_s N_s$. 
For notational consistency with Sec. \ref{sec:problem_setup}, we note that the imbalance here is in subgroups; thus, Group-VS-loss in \eqref{eq:loss_ours_bin_group} consists of logit adjustments that depend on the subgroup $s=(y,g)$. 

\noindent\textbf{Model and Baselines.}
As in \cite{sagawa2019distributionally}, 
we train a ResNet50 starting with pretrained weights on Imagenet. Let $\beta_{s=(y,g)}=(N_{(y,g)}/N_{\text{max}})$. We propose training with the group-sensitive VS-loss in \eqref{eq:loss_ours_bin_group} with $\Delta_{y,g}=\Delta_{s}=\beta_{s}^\gamma$ and $\iota_{s}=-\beta_{s}^{-\gamma}$ with $\gamma=0.3$. We compare against CE and the DRO method of \cite{sagawa2019distributionally}. We also implement a new training scheme that combines Group-VS$+$DRO. We show additional results for Group-LA/CDT (not previously used in group contexts). 
For fair comparison, we reran the baseline experiments with CE and report our reproduced numbers. Since class $+1$ has no special meaning here, we use 
 $\text{Symm-DEO}=(|\Rc_{(+1,+1)}-\Rc_{(+1,-1)}|+|\Rc_{(-1,+1)}-\Rc_{(-1,-1)}|)/2$ and also report balanced and worst sub-group accuracies. 
We did not fine-tune $\gamma$ as the heuristic choice already shows the benefit of Group-VS-loss. We expect further improvements tuning over validation set. 

\noindent\textbf{Results.}
Table \ref{table:waterbirds}  reports test values obtained at last epoch ($300$ in total). 
\begin{wraptable}{r}{0.55\textwidth}
\caption{Symmetric DEO, balanced and worst-case subgroup accuracies on Waterbirds dataset; averages over 10 runs, along with standard deviations.} 
\label{table:waterbirds}
	\begin{center}\resizebox{0.55\textwidth}{!}{
			\begin{tabular}{lccc}
				\toprule
				\textbf{Loss} & \textbf{Symm. DEO} & \textbf{Bal. acc.} & \textbf{Worst acc.}\\
				\toprule
				CE & 25.3$\pm$0.66 & 84.9$\pm$0.29 & 68.1$\pm$2.2\\
				Group LA & 24.0$\pm$2.4 & 84.2$\pm$3.0 & 70.1$\pm$2.6 \\
				Group CDT & 18.5$\pm$0.46 & 87.2$\pm$1.2  & 75.4$\pm$2.2 \\
				Group VS & \textbf{18.1$\pm$0.65} & \textbf{88.1$\pm$0.38} & \textbf{76.7$\pm$2.3}\\ \midrule
				CE + DRO & 16.3$\pm$0.37 & 88.7$\pm$0.31 & 75.2$\pm$2.1\\
				Group LA + DRO & 16.3$\pm$0.82 & 88.7$\pm$0.40 & 74.3$\pm$2.5\\ 
				Group CDT + DRO & \textbf{11.7$\pm$0.15} & \textbf{90.3$\pm$0.2} & \textbf{79.9$\pm$1.5} \\ 
				Group VS + DRO & {11.8$\pm$0.70} & {90.2$\pm$0.22} & {78.9$\pm$1.0} \\
				\bottomrule
		\end{tabular}}
	\end{center}
\end{wraptable}
Our Group-VS loss significantly improves performance (measured with all three fairness metrics) 
over CE, providing a cure for the poor CE performance under overparameterization reported in \cite{sagawa2020investigation}.
Group-CDT/VS have comparable performances, with or without DRO. Also, both outperform Group-LA that only uses additive adjustments. While these conclusions hold for the specific heuristic tuning of $\iota_y$'s, $\Delta_y$'s described above, they are in alignment with our Theorem \ref{propo:gd}.
Interestingly, Group-VS improves by a small margin the worst accuracy over CE+DRO, despite the latter being specifically designed to minimize that objective.
Our proposed Group-VS + DRO outperforms the CE+DRO algorithm used in \cite{sagawa2019distributionally} when training continues in TPT. 
Finally, Symm. DEO appears correlated with balanced accuracy, in alignment with our discussion in Sec. \ref{sec:generalization}  (see Fig. \ref{fig:tradeoffs}(a)).

%% file: conc.tex
\vspace{-0.1in}
\section{Concluding remarks}\label{sec:future_work}
\vspace{-0.1in}
We presented a theoretically-grounded study of recently introduced cost-sensitive CE modifications for imbalanced data. To optimize key fairness metrics, we formulated a new such modification subsuming previous techniques as special cases and provided theoretical justification, as well as, empirical evidence on its superior performance against existing methods.  
 We suspect the VS-loss and our better understanding on the individual roles of different hyperparameters can benefit NLP and computer vision applications; we expect future work to undertake this opportunity with additional experiments. 
 When it comes to group-sensitive learning, it is of interest to extend our theory to other fairness metrics of interest. Ideally, our precise asymptotic theory could help contrast different fairness definitions and assess their pros/cons. Our results are the first to theoretically justify the benefits/pitfalls of specific logit adjustments used in \cite{cosen,TengyuMa,Menon,CDT}. The current theory is limited to settings with fixed features. While this assumption is prevailing in most related theoretical works  \cite{ji2018risk,nacson2019stochastic,hastie2019surprises,bartlett2020benign,muthukumar2020classification},   
 it is still far from deep-net practice where (last-layer) features are learnt jointly with the classifier. We expect recent theoretical developments on that front \cite{NC,mixon2020neural,lu2020neural} to be relevant in our setting when combined with our ideas.

%% file: organization.tex
\section*{Organization of the supplementary material}

The supplementary material (SM) is organized as follows.

\begin{enumerate}

\item In Section \ref{sec:exp_label_app} we provide additional technical information on the  label-imbalanced experiments of Sec. \ref{sec:exp_label}. We also show  experiments of imbalanced MNIST dataset.

\item In Section \ref{sec:exp_group_app} we provide missing details and additional results  on the group-imbalanced experiments of Sec. \ref{sec:GS_exp_main}.

\item In Section \ref{sec:more_num} we present synthetic experiments on both label-imbalanced and group-sensitive datasets further supporting our theoretical findings in Sections \ref{sec:insights_gen} and \ref{sec:generalization}.

\item In Section \ref{sec:propo_proof} we present and prove a more general version of Theorem \ref{propo:gd} (specifically, see Theorem \ref{thm:implicit_loss_gen}) on the connection of overparameterized VS-loss and to CS-SVM. We also discuss multiclass extensions (see Theorem \ref{thm:implicit_loss_multi}) and implicit bias of gradient flow (see Theorem \ref{thm:gradient_flow}).

\item In Section \ref{sec:opt_tune_app}, we present theoretical results on optimal tuning of CS-SVM. First, we state and prove Lemma \ref{lem:1_to_delta} which establishes a structural connection between the solution of CS-SVM to the solution of the standard SVM, allowing to view the former as a post-hoc adjustment to the latter. Then, we use this property together with the sharp characterizations of Theorem \ref{thm:main_imbalance} to derive an explicit formula for the optimal margin ratio under Gaussian mixture data.
%

\item In Section \ref{sec:cs-svm_proof} we prove Theorem \ref{thm:main_imbalance} on generalization of CS-SVM. We also discuss related works on sharp high-dimensional asymptotics and provide necessary background on the convex Gaussian min-max theorem. Finally, we include formulas for the phase-transition threshold of CS-SVM.

\item Finally, in Section \ref{sec:gs-svm_proof} we {state and prove Theorem \ref{thm:main_group_fairness} characterizing the DEO of GS-SVM} as mentioned in Section \ref{sec:generalization}. 

\end{enumerate}


%% file: imba_exp.tex
\section{Additional Experiments on Label-Imbalanced Datasets}\label{sec:exp_label_app}

In this section, we provide omitted information on the results of Section \ref{sec:exp_label}, as well, as additional experiments. 

\subsection{Deep-net experiments}

Here we provide additional implementation details and a more extensive discussion on the results presented in Table \ref{table:CIFAR} in Section \ref{sec:exp_label} of the main text.

\noindent\textbf{Technical details:} Following \cite{TengyuMa}, we train a ResNet-32 \cite{he2016deep}, using batch size $128$ and SGD with momentum $0.9$ and weight decay $2\times10^{-4}$. For the first $5$ epochs we use a linear warm up schedule until baseline learning rate of $0.1$. We train for a total of 200 epochs, while decaying our learning rate by $0.1$ at epochs $160$ and $180.$ For STEP-100 imbalance we trained for $300$ rather than $200$ epochs and adjusted the learning rate accordingly as we found this type of imbalance more difficult to learn.  We remark that the values for LDAM (adapted from \cite{TengyuMa}) used learning rate decay 0.01 and last-layer feature/classifier normalization. We have found convergence difficult otherwise. For other losses, we do \emph{not} use the above normalization of weights to isolate the impact of loss modifications. 

\noindent\textbf{Implementation details.}~A seed is used for each of the $5$ runs and the weights of the network are initialized with the same values for all the losses that we train. We only show $95\%$ confidence intervals for CE, LA, CDT and VS losses which we implemented. For the remaining algorithms (e.g., LDAM), we report averages over $5$ realizations as given in \cite{TengyuMa}. For LA, CDT and VS losses, we have tuned the hyper-parameters $(\tau,\gamma)$ over the validation set as described in Section \ref{sec:exp_label} (see Remark \ref{rem:taugamma} and Table \ref{table:hyperparameters}). More sophisticated tuning strategies over the validation set (e.g., based on bilevel optimization or Hyperband \cite{lorraine2020optimizing,li2017hyperband}) and the corresponding performance assessment on test set are left to future work. Same as in \cite{he2016deep, TengyuMa, Menon, CDT} before training we augment the data by padding the images to size $40 \times 40$, flipping them horizontally at random and then random cropping them to their original size. We use PyTorch \cite{paszke2017automatic} building on codes provided by \cite{TengyuMa,CDT}. Training is performed on 2 NVIDIA RTX-3080 GPUs.

 \begin{remark}[On the $(\tau,\gamma)$ parameterization of $\iota_y$'s \& $\Delta_y$'s]\label{rem:taugamma} As mentioned in Section \ref{sec:exp_label}, our deep-net experiments with VS-loss for label-imbalances, use the  following parameterization for the additive and multiplicative logit factors in terms of two hyperparameters $\tau$ and $\gamma$:
\begin{align}\label{eq:tau_gamma}
\iota_y = \tau\log(N_y/N_{\rm tot})\qquad\text{and}\qquad\Delta_y=(N_y/N_{\max})^\gamma,
\end{align}
where $N_y$ is the train-sample size of class $y$, $N_{\max}=\max_y N_y$ and $N_{\rm tot}=\sum_y N_y$. This parameterizations follow \cite{Menon} and \cite{CDT}, respectively. A convenient feature is that setting $\tau=0$ recovers the CDT-loss, and setting $\gamma=0$ recovers the LA-loss. 
\end{remark}

\noindent\textbf{Results and discussion.}~Table \ref{table:CIFAR} shows that our VS-loss performs favorably over the other methods across all experiments. The margins of improvement depend on the dataset / imbalance-type. Also, observe that in most cases LA-loss performs better than CDT-loss. This is likely because the CDT loss enters the TPT slower for the shown amount of training. Interestingly, VS-loss, even though it resembles the CDT-loss in the fact that it also adjusts the logits multiplicatively, does \emph{not} seem to suffer from the same problem. In Section \ref{sec:insights}, we presented experiments showing that: (i) If given enough time to train, CDT-loss can achieve similar or better results than LA-loss. (ii) The addition of the $\iota_y$'s in the VS-loss can mitigate the effect of $\Delta_y$ on the speed of convergence. In that sense, VS-loss fulfills the theoretical intuition in Section \ref{sec:insights}, as the method that combines additive and multiplicative adjustments for high accuracy and fast convergence.

\noindent\textbf{Tuning results.}~To promote reproducibility of our results and to give some insight on the range of $\tau$ and $\gamma$, in Table \ref{table:hyperparameters} we present the values of the hyperparameters that we determined through tuning and used to generate Table \ref{table:CIFAR}. As we discussed in Sec. \ref{sec:insights}, large values of $\tau$ and $\gamma$ can hinder training. Thus, when training with the VS loss, which adjusts the logits both in an additive and in a multiplicative way, it seems beneficial to use smaller values of these parameters, than when training with the LA or the CDT losses. 
Additionaly, note that if searching over a grid, it is possible that the best values found for the VS-loss, will be the same as those of the LA or CDT losses (but never worse than them). Searching over a fine enough grid though should yield parameter values for which VS-loss outperforms both of them. Finally, note that the $(\tau,\gamma)$-parameterization of the $\iota_y, \Delta_y$'s is itself restrictive and other alternatives might yield further improvements when combining both types of adjustments as observed in the other cases. 

\begin{table}[t]
   \caption{Hyperparameter tuning results for each dataset, imbalance profile and loss function.}
   \label{table:hyperparameters}
    \begin{center}{
        \begin{tabular}{lllllllll} \hline
        \textbf{Dataset}         & \multicolumn{2}{c}{\textbf{CIFAR 10}} & \multicolumn{2}{c}{\textbf{CIFAR 100}} \\ \hline
        \textbf{Imbalance Profile}  & \multicolumn{1}{c}{\textbf{LT-100}} & \multicolumn{1}{c}{\textbf{STEP-100}} & \multicolumn{1}{c}{\textbf{LT-100}} & \multicolumn{1}{c}{\textbf{STEP-100}} \\ \hline \hline
        LA ($\tau=\tau^{*}$) \cite{Menon}                       	& $2.25$ & $2.25$ & $1.375$ & $0.875$ \\
        CDT ($\gamma=\gamma^{*}$)  \cite{CDT}         	& $0.4$ & $0.3$ & $0.1$ & $0.1$  \\ \hline \hline
        VS ($\tau=\tau^{*}, \gamma=\gamma^{*}$)         & $(1.25, 0.15)$ & $(1.5, 0.2)$ & $(0.75, 0.05)$ & $(0.5, 0.05)$  \\ 
        \end{tabular}}
    \end{center}
\end{table}

\input{mnist_Orestis}

%% file: mnist_Orestis.tex
\subsection{Experiments on the MNIST dataset}\label{sec:MNIST}
Here, we present additional results on imbalanced MNIST data trained with linear and random-feature models. These results complement the synthetic experiment of Figure \ref{fig:mismatch_intro}(a). 

\begin{figure}[t]
\begin{center}
\begin{subfigure}[b]{0.45\textwidth}
         \centering
         \includegraphics[width=0.8\textwidth]{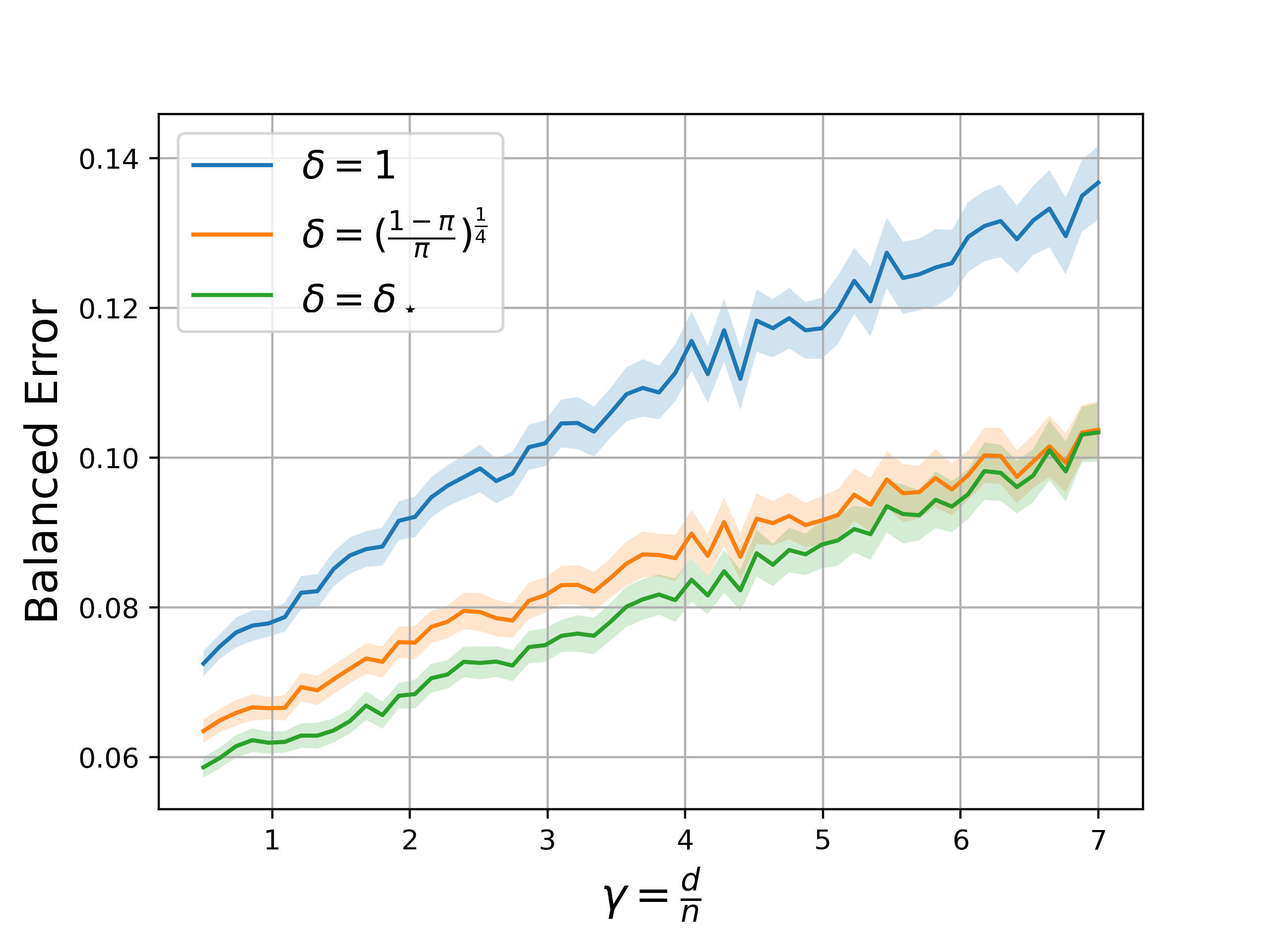}
                  \caption{Linear classifier}
\end{subfigure}
\begin{subfigure}[b]{0.45\textwidth}
         \centering
         \includegraphics[width=0.8\textwidth]{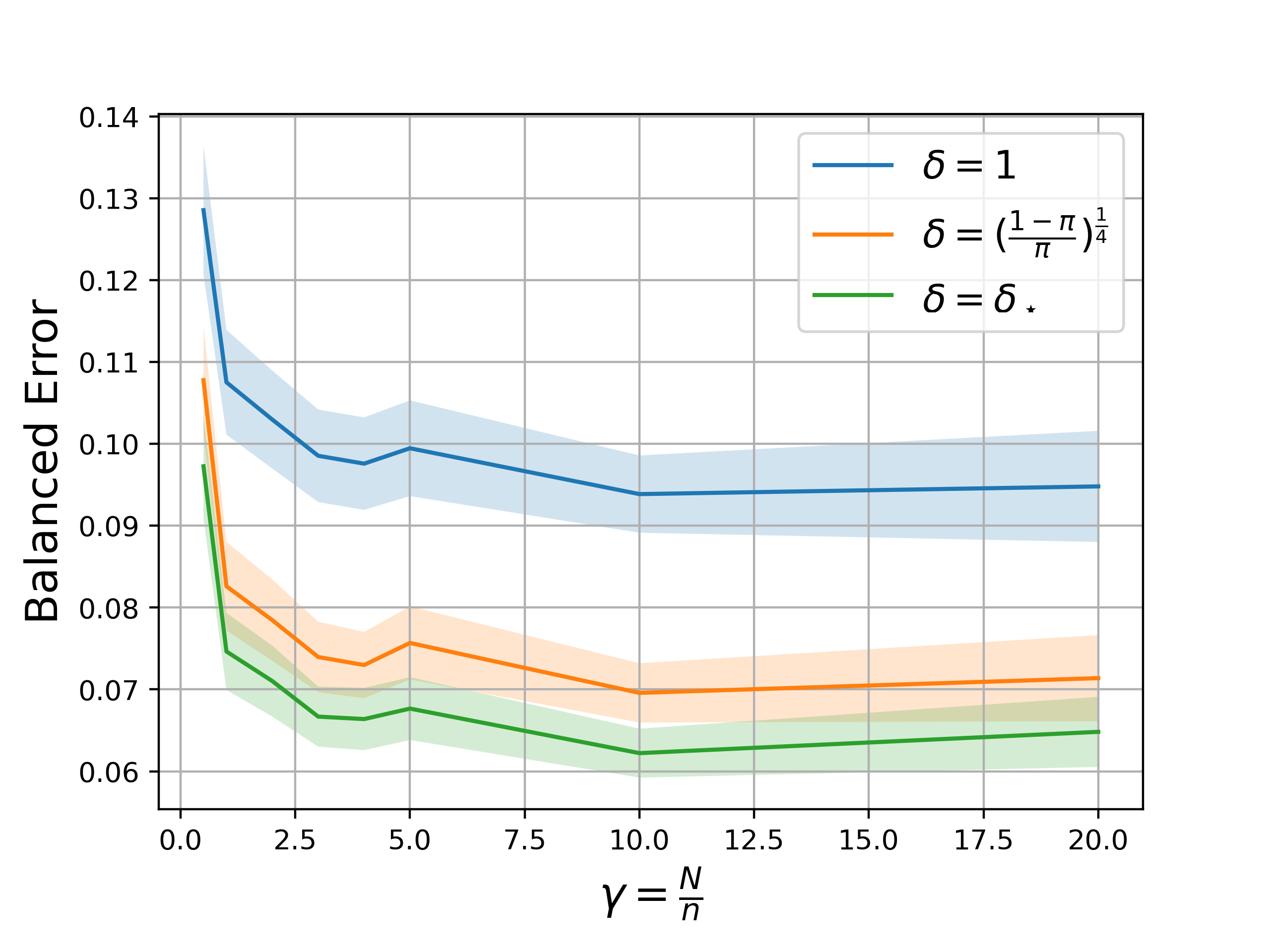}
         \caption{Random-features classifier}
\end{subfigure}
\end{center}
\caption{A comparison of CS-SVM balanced error against the overparameterization ratio $\gamma$, for the standard hard margin SVM ($\delta=1$), for a heuristic  $\delta = (\frac{1-\pi}{\pi})^\frac{1}{4}$ and for our \emph{approximation} of the optimal $\delta$ ($\delta=\tilde\delta_\star$) obtained by the data-dependent heuristic in Section \ref{sec:delta_star_estimate}. The experiment is performed on the MNIST dataset in a one-vs-rest classification task where the goal is to separate the minority class containing images of the digit $7$ from the majority class containing images of all other digits. See text for details.}
\label{fig:mnist_delta_tuning}
\end{figure}

Specifically, we designed an experiment where we perform binary one-vs-rest classification on the MNIST dataset to classify digit $7$ from the rest. Specifically, we split the dataset in two classes, the minority class containing images of the digit $7$ and the majority class containing images of all other digits. To be consistent with our notation we assign the label $+1$ to the minority class and the label $-1$ to the majority class. Here, $d=784$ and $\pi=0.1$ is the prior for the minority class. All test-error evaluations were performed on a test set of $1000$ samples. The results of the experiments were averaged over 200 realizations and the $90\%$ confidence intervals for the mean are shown in Figure \ref{fig:mnist_delta_tuning} as shaded regions.

\begin{figure}[t]
         \centering
         \includegraphics[width=0.5\textwidth]{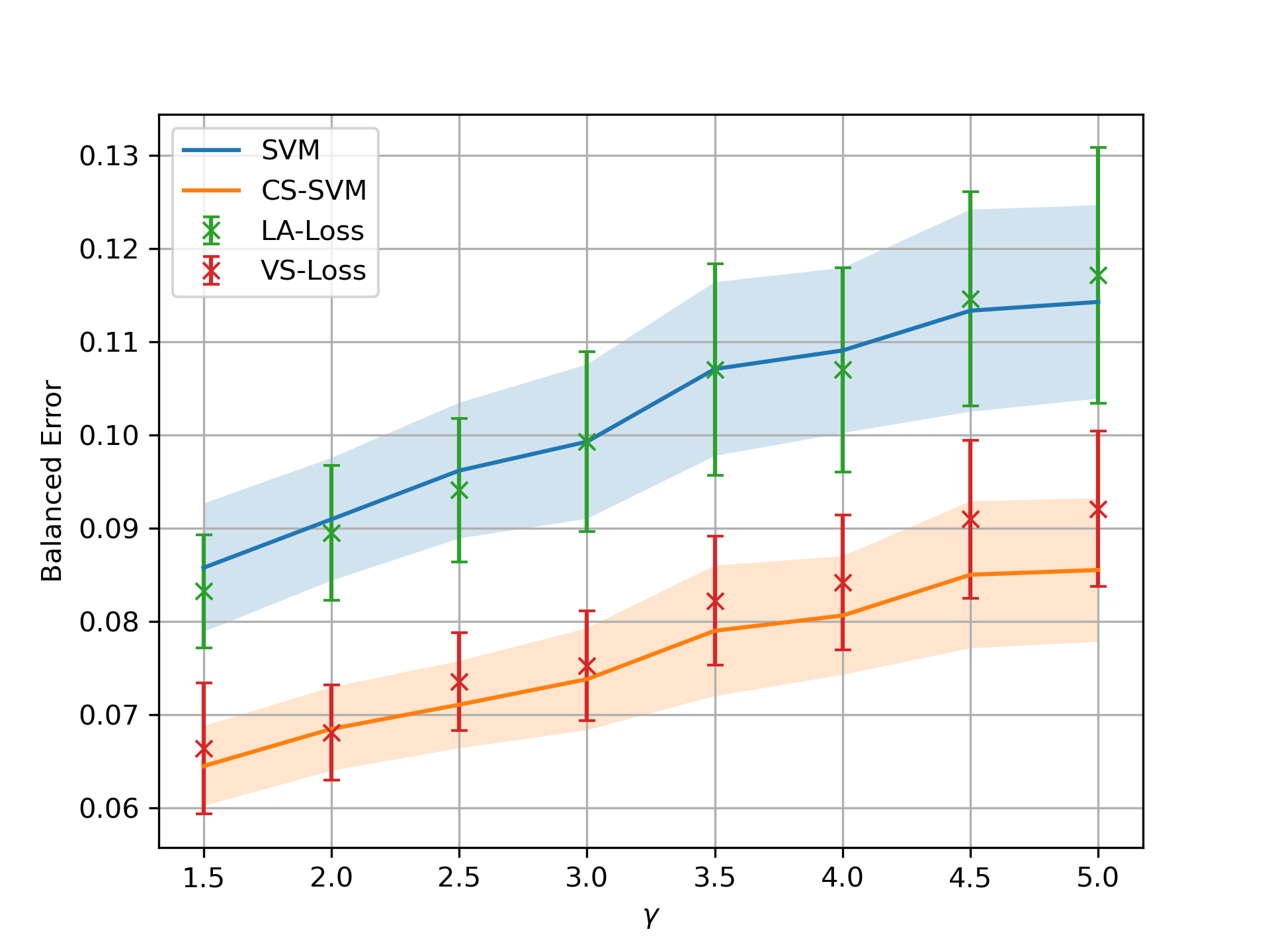}
\caption{In the overparameterized regime, our VS loss converges to the CS-SVM classifier, while the LA-loss converges to the inferior ---in terms of balanced-error performance--- SVM. The experiment was performed on the MNIST dataset in a one-vs-rest classification task where the goal is to separate the minority class containing images of the digit $7$ from the majority class containing images of all other digits. See text for details.}
\label{fig:la_vs_mnist_f}
\end{figure}

We ran two experiments. In the first one depicted in Figure \ref{fig:mnist_delta_tuning}(a), we trained linear classifiers using the standard SVM (blue), the CS-SVM with a heuristic value $\delta=(\frac{1-\pi}{\pi})^\frac{1}{4}$ (orange), and the CS-SVM with our heuristic data-dependent estimate of the optimal $\tilde\delta_\star$ (green). We compute such an estimate based on a recipe inspired by our exact expression in \eqref{eq:delta_star} for the GMM; see Section \ref{sec:delta_star_estimate} for details. We compute the three classifiers on training sets of varying sizes $n=d/\gamma$ for a range of values of $\gamma$ and report their balanced error. We observe that CS-SVM always outperforms SVM (aka $\delta=1$) and the heuristic optimal tuning of CS-SVM consistently outperforms the choice $\delta=(\frac{1-\pi}{\pi})^\frac{1}{4}$.

 

Next, in Figure \ref{fig:mnist_delta_tuning}(b) for the same dataset we trained a Random-features classifier. Specifically, for each one of the $n=300$ training samples $\x_i\in\R^{d=784}$ we generate random features $\wtd\x_i={\rm ReLU}(\A\x_i)$ for a matrix $\A\in\R^{N\times d}$ which we sample once such that it has entries IID standard normal and is then standardized such that each column becomes unit norm. 
In this case we control $\gamma$ by varying the number $N=\gamma n$ of rows of that matrix $\A$. Observe here that the balanced error decreases as $\gamma$ increases (an instance of benign overfitting, e.g. \cite{hastie2019surprises,bartlett2020benign,mei2019generalization} and that again the estimated optimal $\delta_\star$ results in tuning of CS-SVM that outperforms the other depicted choices.

In Figure \ref{fig:la_vs_mnist_f} we repeat the experiment of Figure \ref{fig:mnist_delta_tuning}(a) only this time additionally to training CS-SVM for $\delta=1$ and for $\delta=\tilde\delta_\star$ we also train using the LA-loss and our VS-loss. 
 For the VS loss we use \eqref{eq:loss_ours_bin} with the following choice of parameters: $\omega_\pm=1$, $\iota_\pm=0$ and $\Delta_y=\tilde\delta_\star^{-1}\,\ind{y=+1}+\ind{y=-1}$ (see Section \ref{sec:delta_star_estimate} for $\tilde\delta_\star$). In a similar manner, LA-loss is defined using the same formula \eqref{eq:loss_ours_bin}, but with parameters $\Delta_\pm=1$, $\omega_\pm=1$ and $\iota_+=\pi^{-1/4}, \iota_-=(1-\pi)^{-1/4}$ (as suggested in \cite{TengyuMa}). 
 
 The figure confirms our theoretical expectations: training with gradient descent on the LA and VS losses asymptotically (in the number of iterations) converge to the SVM and CS-SVM solutions respectively.

The training is performed over 200 epochs and for computing the gradient we iterate through the dataset in batches of size 64. The results are averaged over 200 realizations and the $90\%$ confidence intervals are plotted as shaded regions for the CS-SVM model and as errorbars for the VS loss.

%% file: group_exp_app.tex
\section{Further details and additional experiments on group-imbalances}\label{sec:exp_group_app}
\subsection{Deep-net experiments}
In this section, we elaborate on our proposed method of combining our group logit-adjusted losses with the DRO method. In all experiments, we chose $\Delta_{s}=(N_{s}/N_{\max})^\gamma$, $\iota_{s}=-(N_{s}/N_{\max})^{-\gamma}$ with $\gamma=0.3$. For example, Group-LA has $\iota_s=-(N_{s}/N_{\max})^{-0.3}$ and $\Delta_s=0$.

 
\noindent\textbf{Group-VS+DRO algorithm.}~For completeness, we elaborate on our proposed method of combining DRO with our Group VS-loss (see bottom half of Table \ref{table:waterbirds}). We recall from \cite{sagawa2019distributionally} that their proposed CE+DRO algorithm seeks a model that minimizes the worst subgroup empirical risk by instead minimizing the worst subgroup CE-loss: 
$
\max_{s\in\mathcal{S}}\mathbb{E}_{(\x,y)\sim\hat P_s}[\ell_{\rm CE}(y,f_\w(\x))], 
$
where $\hat P_s$ is the empirical distribution on  training samples from subgroup  $s$. Instead, our Group-VS+DRO method attempts to solve the following distributionally robust optimization problem:
\begin{align*}
	\min_\w\max_{s\in\mathcal{S}}\,\mathbb{E}_{(\x,y)\sim\hat P_s}[\ell_{\rm Group-VS}(y,s,f_\w(\x))],
\end{align*}
with $\ell_{\rm Group-VS}(y,s,f_\w(\x))=\omega_{s}\cdot\log\big(1+e^{{\iota_{s}}}\cdot e^{-{\Delta_{s}} yf_\w(\x)}\big)$ (see Equation \eqref{eq:loss_ours_bin_group}). To solve the above non-convex non-differentiable minimization, we employ the same online optimization algorithm  given in \cite[Algorithm 1]{sagawa2019distributionally}, but changing the CE loss to the Group-VS. 

\subsection{GS-SVM experiments}

 \begin{figure}[t]
	\begin{center}
		\begin{subfigure}[b]{0.45\textwidth}
			\centering
			\includegraphics[width=0.65\textwidth]{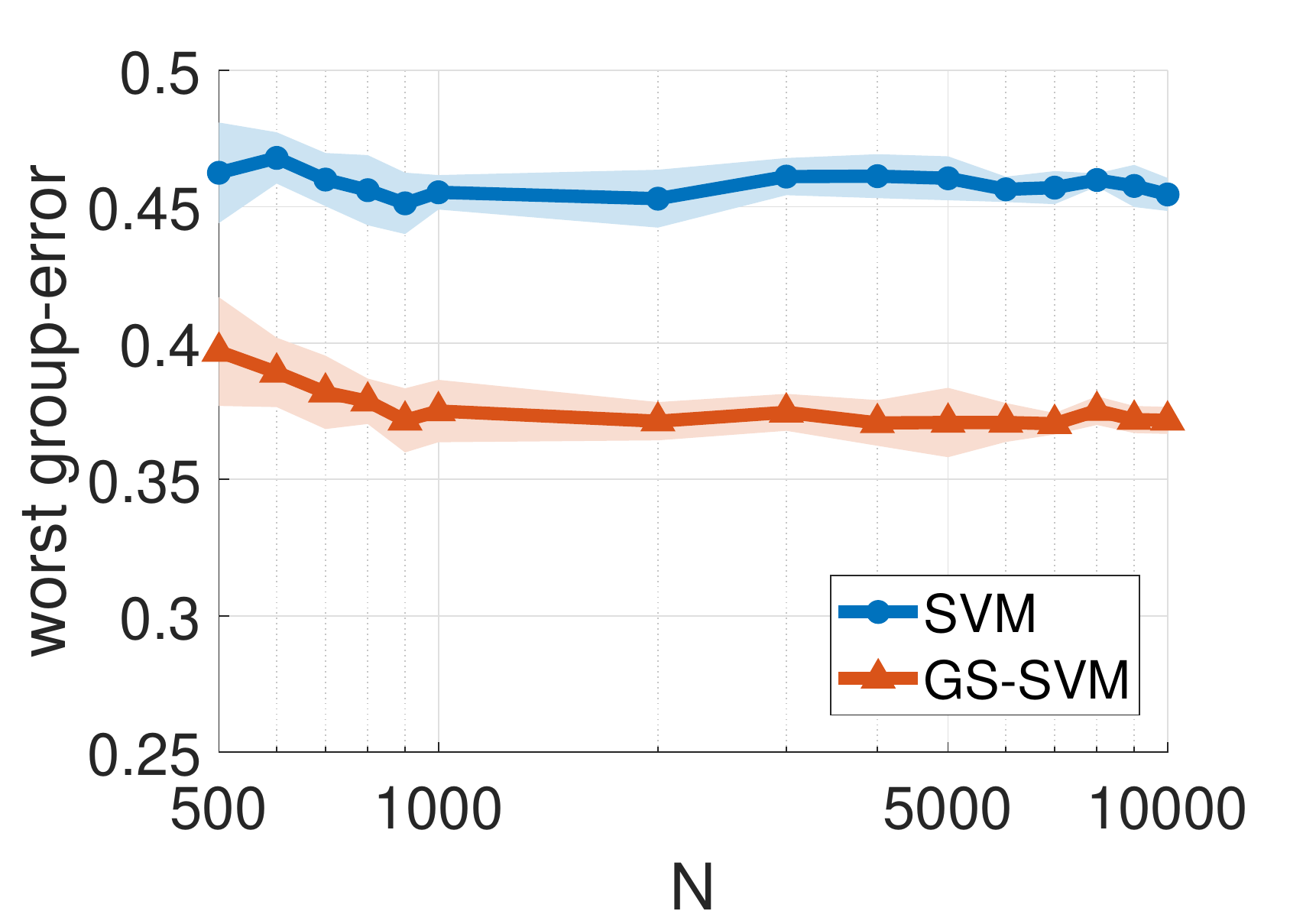}
			\caption{Worst case sub-group error}
		\end{subfigure}
		\begin{subfigure}[b]{0.45\textwidth}
			\centering
			\includegraphics[width=0.65\textwidth]{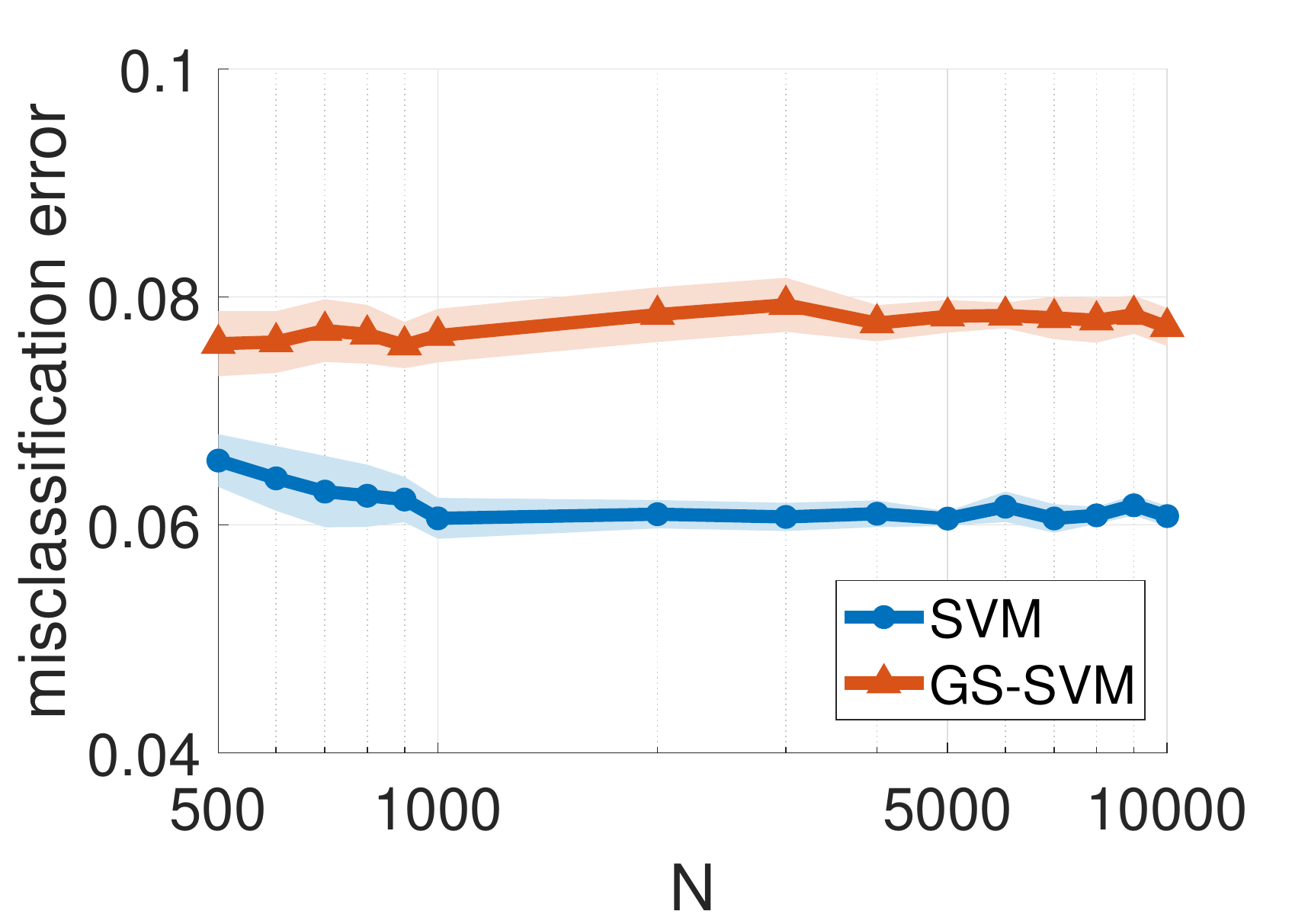}
			\caption{Misclassification error}
		\end{subfigure}
	\end{center}
	\caption{The benefit of GS-SVM (corr. Group-VS loss) compared to SVM (corr. wCE) in achieving smaller \emph{worst case sub-group error} without significant loss on the misclassification error in the Waterbirds dataset. Training a linear model with $N$-dimensional Random-feature map over pretrained ResNet-18 features as in \cite{sagawa2020investigation}. }
	\label{fig:waterbirds_expt}
\end{figure}

Section \ref{sec:GS_exp_main} demonstrated, for a deep-net model trained on the Waterbird dataset, the efficacy of the Group-VS loss compared to the CE and DRO algorithms used in \cite{sagawa2019distributionally}. Here, we follow \cite{sagawa2020investigation} who, similar to us, focused in overparameterized training in the TPT. Specifically, \cite{sagawa2020investigation} showed that wCE trained on a Random-feature model applied on top of a pretrained ResNet results in large \emph{worst-group error}  when trained in TPT. In their analysis, they observed that this is because weighted logistic loss in the separable regime behaves like SVM, which is insensitive to groups. Here, we repeat their experiment only this time we use the Group-VS loss. In line with our results thus far, Group-VS loss shows improved performance in this setting as well. 
 
%
%

\noindent\textbf{Algorithm.}~Concretely, since we are training linear models (on random feature maps), we know from Theorem \ref{propo:gd} that Group-VS loss converges to GS-SVM. Thus, for simplicity, we directly trained the following instance of  GS-SVM and compared it against SVM:
\begin{align}\label{eq:gs_svm_waterbirds}
	~~\min_{\w}~~\|\w\|_2 \qquad\text{sub. to}~~y_i(h(\x_i)^T\w+b)\geq \delta_{s_i}, ~i\in[n]. 
\end{align}
Above, $\delta_{s_i}=\delta_{(y_i,g_i)}=(\frac{1}{\hat p_{(y,g)}})^4$, $h:\Xc\rightarrow\R^N$ is the random-feature map (see Section \ref{sec:MNIST}), and
 $\x_i,i \in[n]$ are $d$-dimensional pretrained ResNet18 features (same as those used in \cite{sagawa2020investigation}). Here, $n=4795$, $N$ took a range of values from $500$ to $10000$ and $d=512$. For those values of $N$ the data are separable, thus SVM/GS-SVM are feasible.

\noindent\textbf{Experiment \#1: GS-SVM vs SVM (or, Group-VS vs wCE).}~Figure  \ref{fig:waterbirds_expt} shows worst-group and missclassification errors of GS-SVM and SVM as a function of the feature dimension $N$. The curves show averages over $10$ realizations of the random projection matrix along with standard deviations depicted using shaded error-bars. We confirm that:
\begin{itemize}
\item GS-SVM consistently outperforms standard SVM in the overparameterized regime in terms of worst-group error 
\item This gain comes without significant losses  on the misclassification error.
\end{itemize}

\begin{figure}[t]
	\begin{center}
		\begin{subfigure}[b]{0.5\textwidth}
			\centering
			\includegraphics[width=0.44\textwidth]{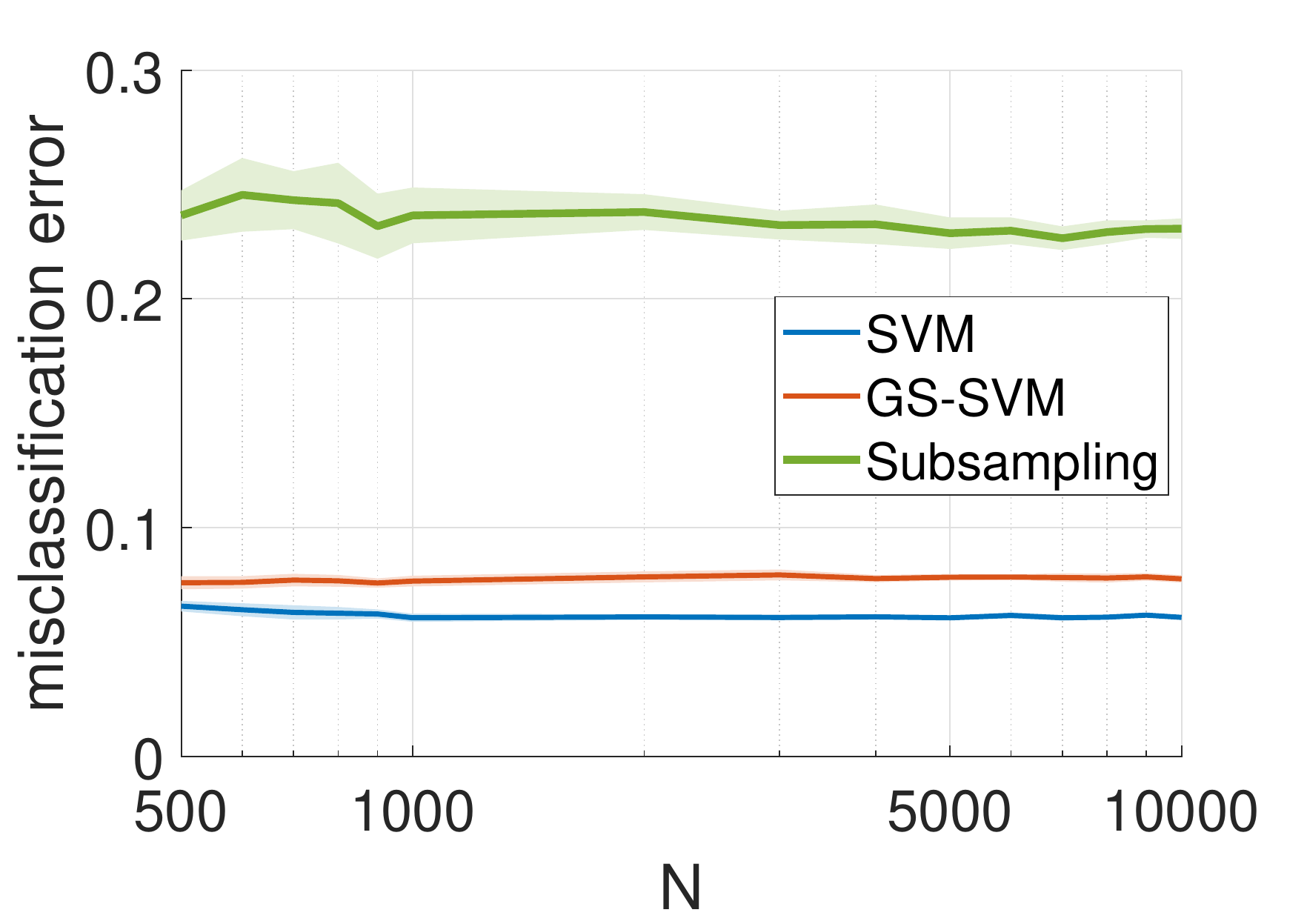}
			\caption{misclassification error}
		\end{subfigure}
		\\
		\begin{subfigure}[b]{0.24\textwidth}
			\centering
			\includegraphics[width=\textwidth]{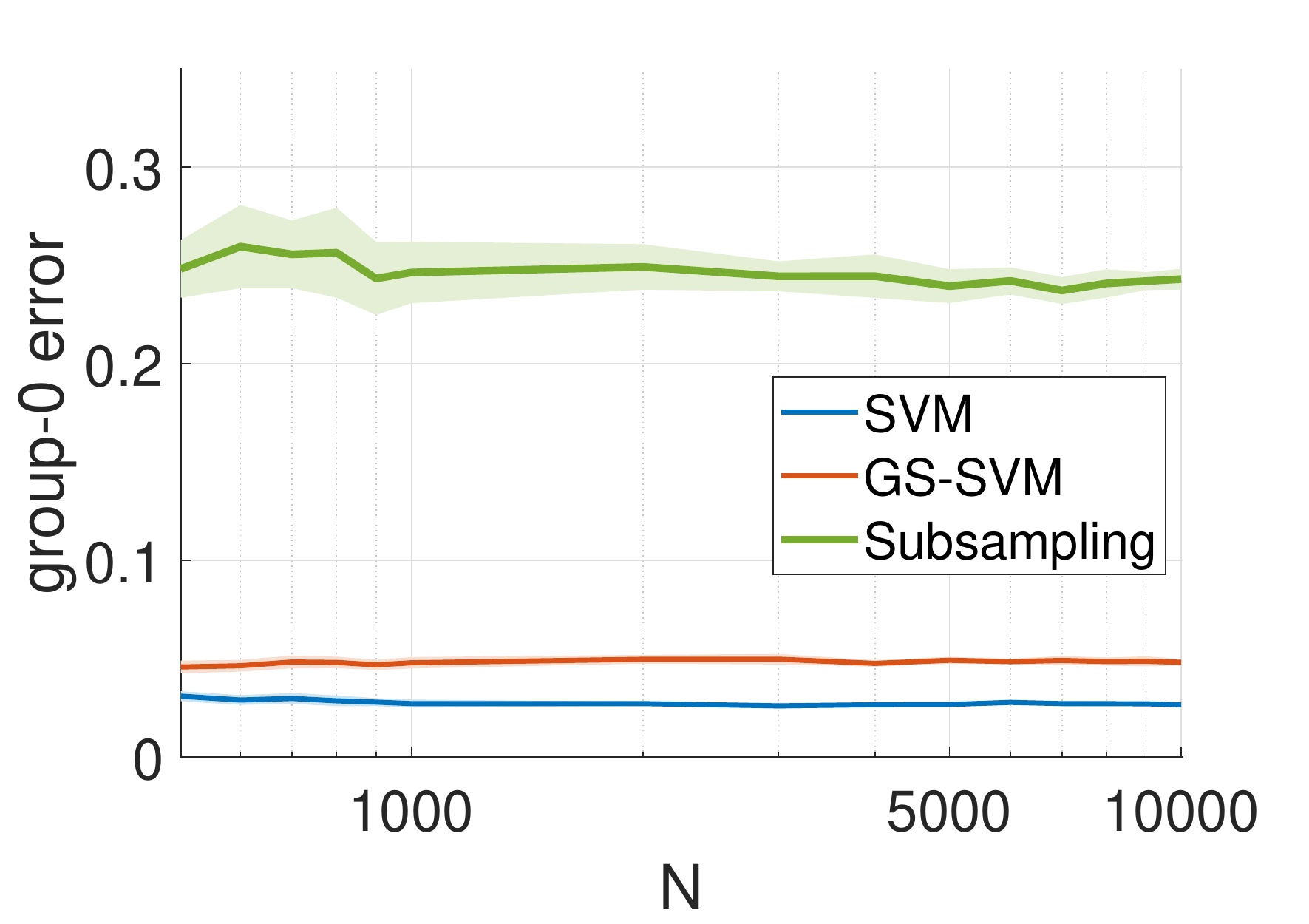}
			\caption{Sub-group-0 error}
		\end{subfigure}
		\begin{subfigure}[b]{0.24\textwidth}
			\centering
			\includegraphics[width=\textwidth]{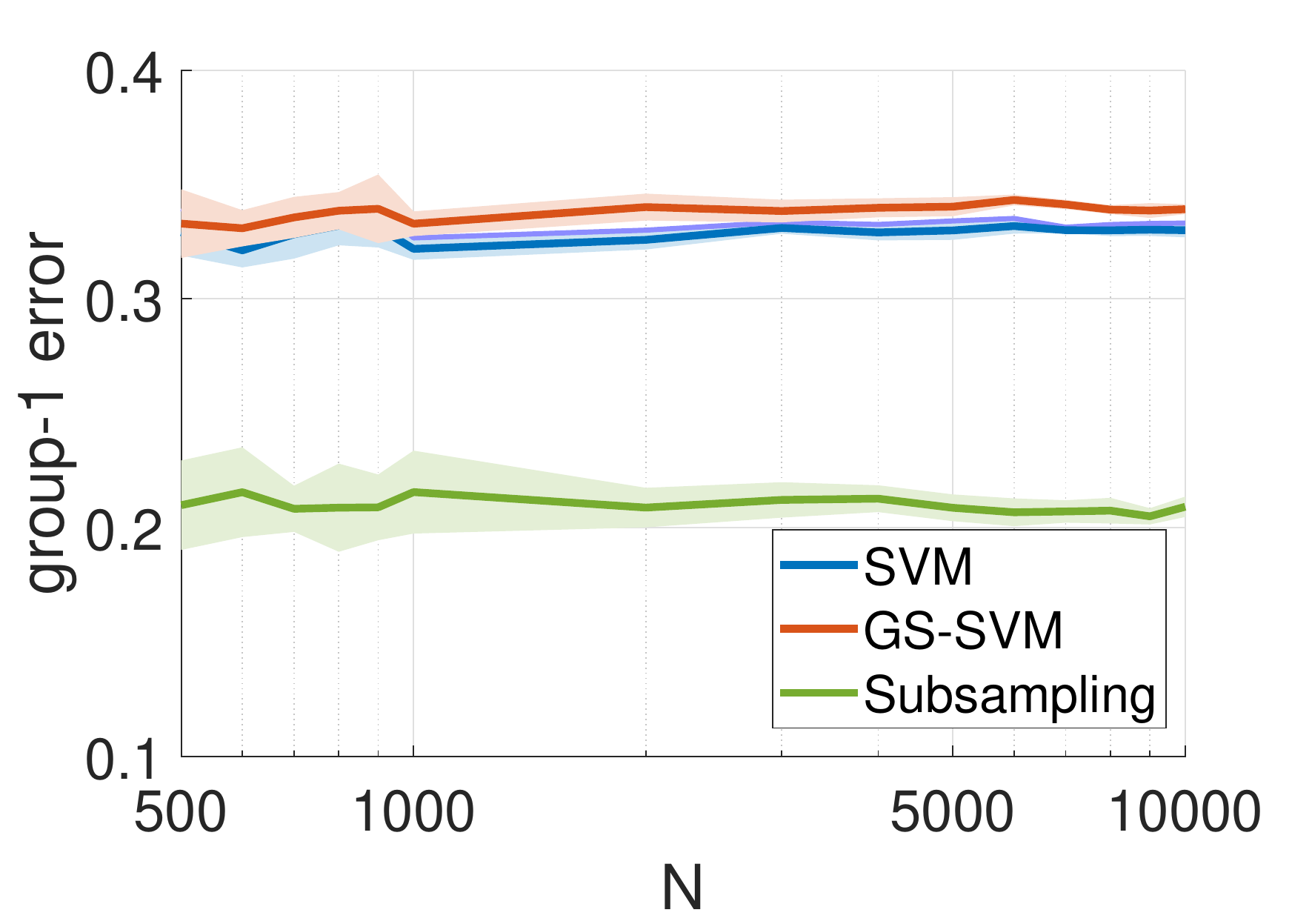}
			\caption{Sub-group-1 error}
		\end{subfigure}
		\begin{subfigure}[b]{0.24\textwidth}
			\centering
			\includegraphics[width=\textwidth]{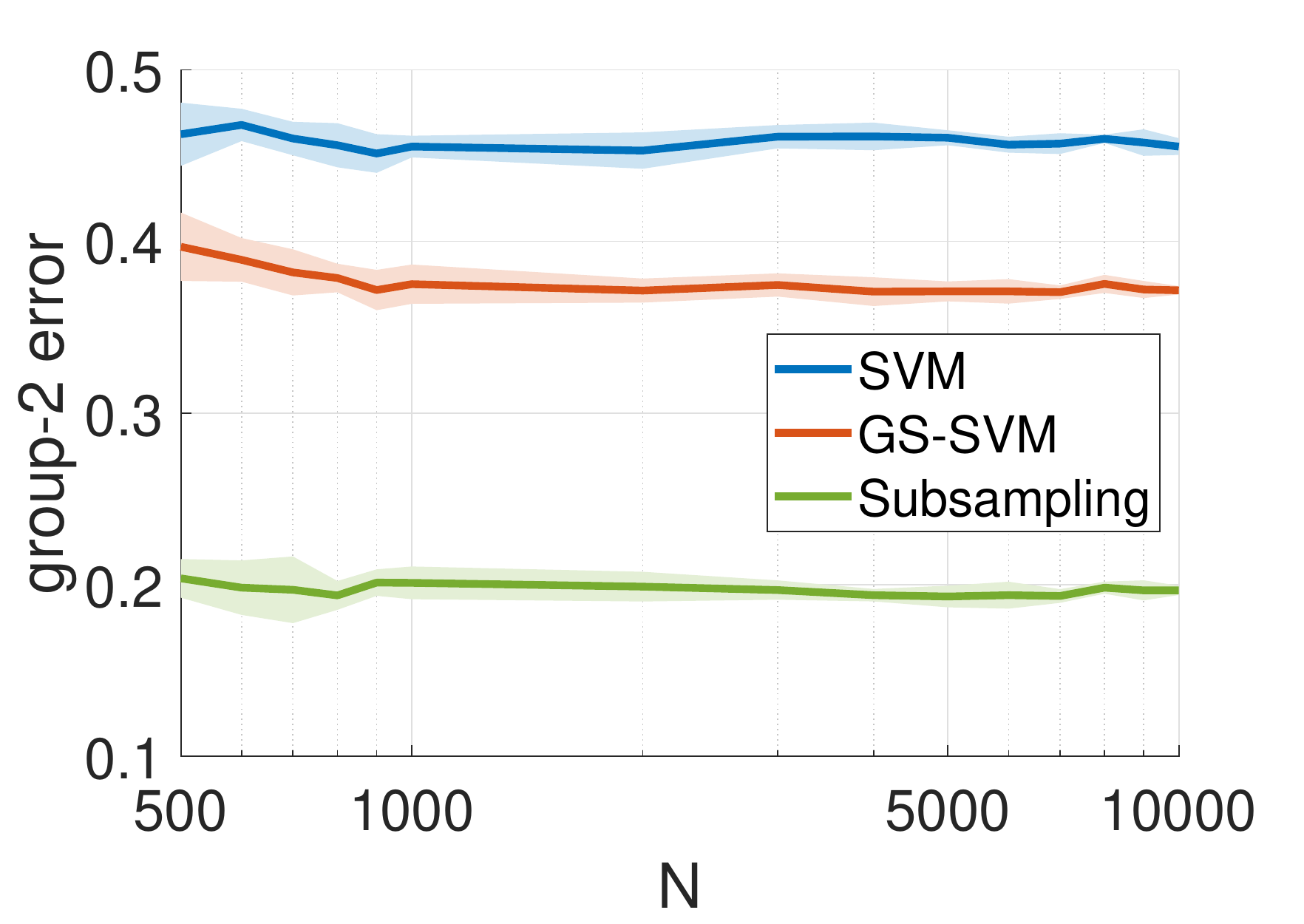}
			\caption{Sub-group-2 error}
		\end{subfigure}
		\begin{subfigure}[b]{0.24\textwidth}
			\centering
			\includegraphics[width=\textwidth]{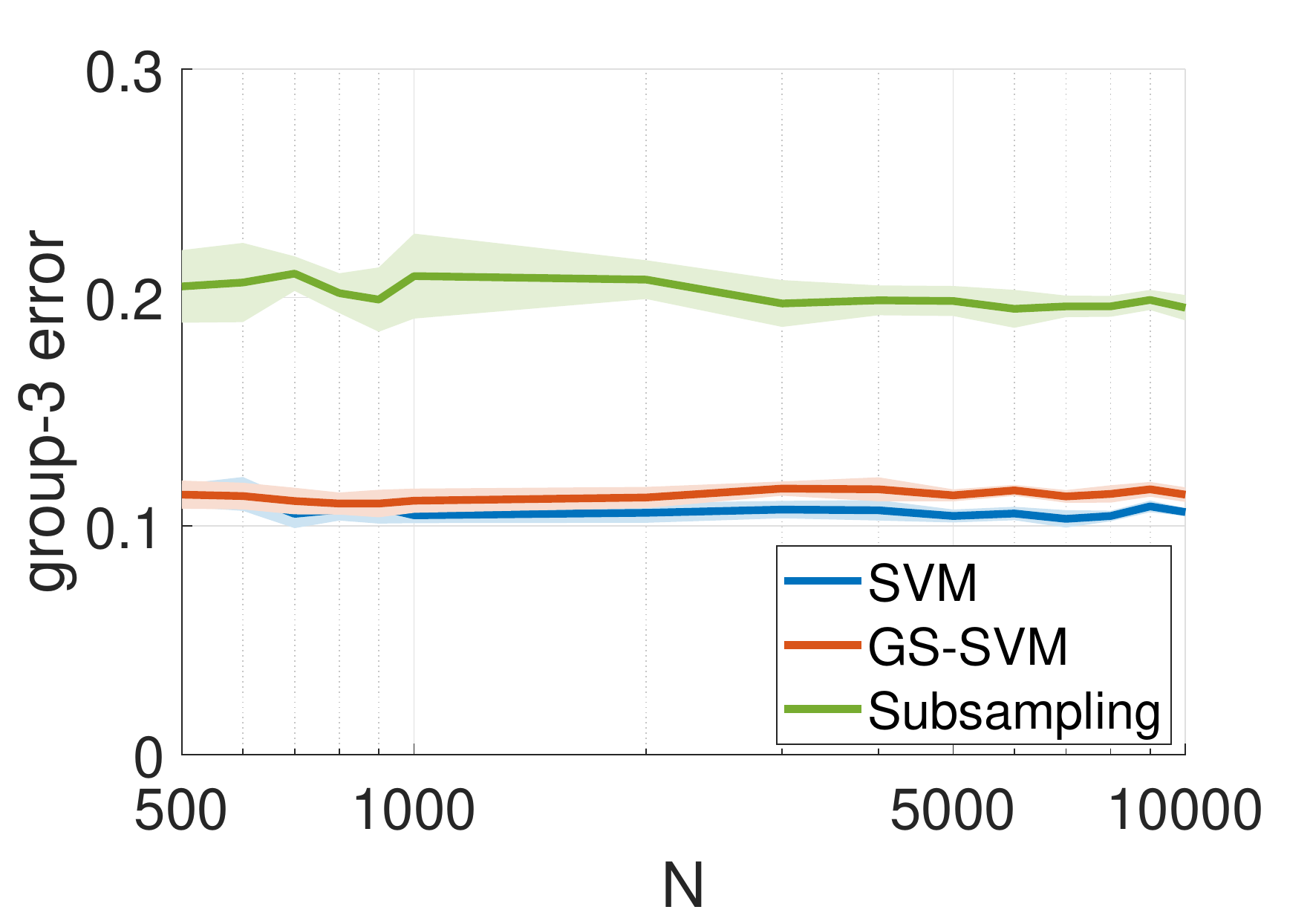}
			\caption{Sub-group-3 error}
		\end{subfigure}
	\end{center}
	\caption{Misclassification and conditional sub-group errors of SVM (blue), GS-SVM with heuristic tuning $\delta_{(y,g)}={p^{-4}_{(y,g)}}$ (red), and, SVM with subsampling (green) for the Waterbirds dataset. GS-SVM has lower worst-case error (Sub-group-2) compared to the SVM without significant increase on the misclassification error. SVM with subsampling has the best worst-group error performance, but also worst misclassification error in subfigure (a).}
	\label{fig:waterbirds_group_errors_svm_gs_ss}
\end{figure}

\noindent\textbf{Experiment \#2: GS-SVM vs Sub-sampling.}~
As a means of improving over wCE, \cite{sagawa2020investigation} proposed instead the use of \emph{CE with subsampling,} for better \emph{worst case sub-group error}. In Figure \ref{fig:waterbirds_group_errors_svm_gs_ss} we compare the performance of three algorithms: (i) SVM, (ii) GS-SVM, and (iii) SVM with subsampling (corresponding to CE with subsampling).  For the latter, 
we chose $56$ examples from every sub-group (this is the size of the smallest sub-group) and ran SVM on the resulting (smaller), now balanced, dataset. Figure \ref{fig:waterbirds_group_errors_svm_gs_ss}  reports missclassification error, as well as, \emph{conditional sub-group errors}. Recall that in the original dataset, sub-groups- 0 and 3 were the \emph{majority} with 3498 and 1057 examples respectively, while sub-groups 1 and 2 were the \emph{minorities} with 184 and 56 examples, respectively.  We find the following:
\begin{itemize}
\item Consistent with \cite{sagawa2020investigation} SVM with subsampling achieves low \emph{worst case sub-group error}, lower than both SVM and GS-SVM (at least, when tuned with $\delta_{(y,a)}=\big(\frac{1}{p_{(y,a)}}\big)^4$).
\item Specifically, note the very low errors achieved by SVM with subsampling  for minority sub-groups 2 and 3.
\item However, the gain comes at a significant cost paid for the majority sub-groups- 1 and 3 resulting in an increase of the misclassification error  by  more than $3-$ times compared to standard SVM and GS-SVM. 
\end{itemize}

We expect that, with more careful tuning of the hyper-parameters $\delta_{(y,g)}$,  GS-SVM can eventually achieve even lower \emph{sub-group errors} for the minority sub-groups without hurting the \emph{majority sub-group errors} significantly. We leave this to future work.

%% file: more_num.tex
\begin{figure}[h]
 \centering
         \includegraphics[width=0.65\textwidth,height=0.45\textwidth]{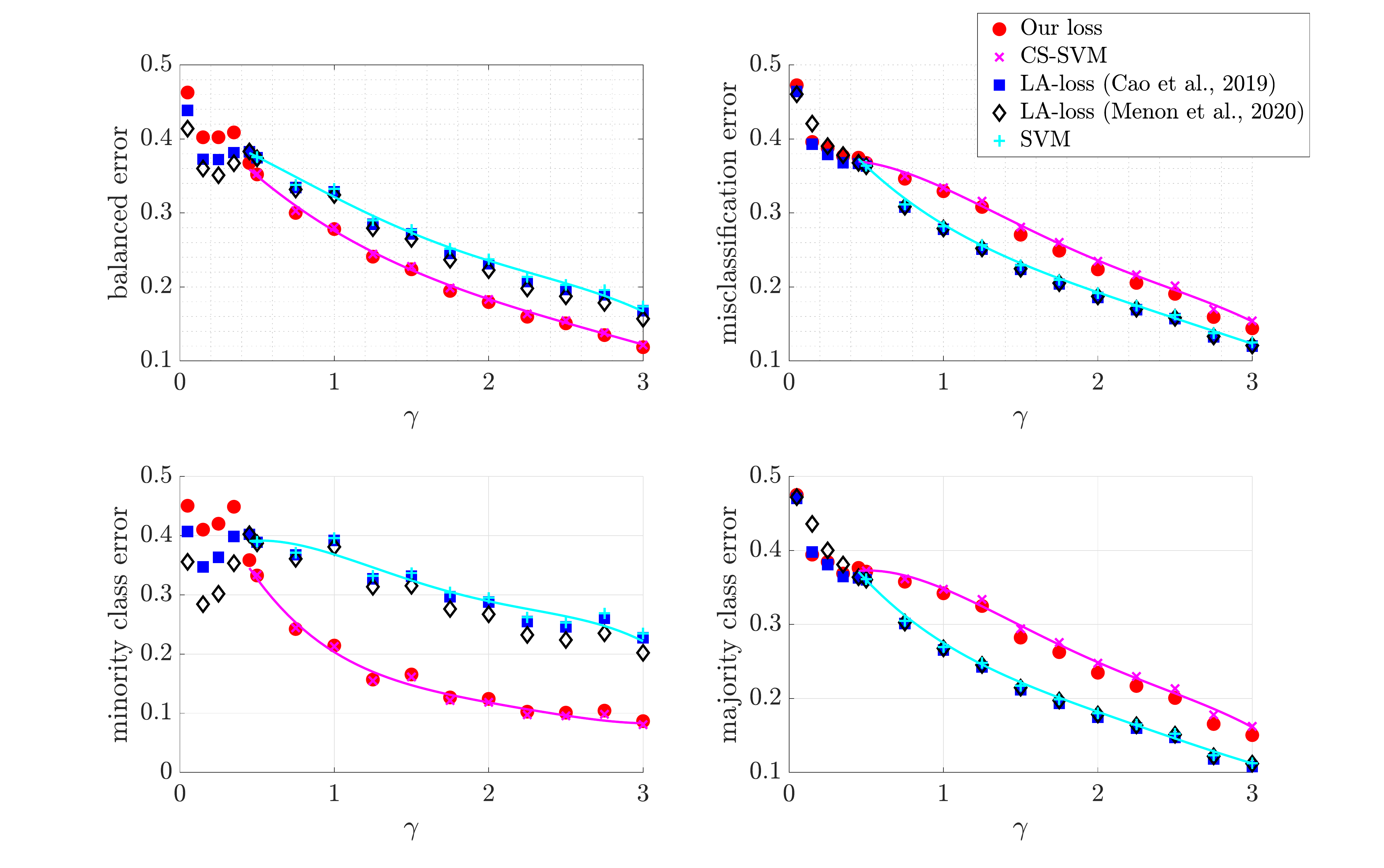}
\caption{Performance of CDT vs LA loss for label-imbalanced GMM with missing features. (Top Left) is same as Figure \ref{fig:mismatch_intro}(a). The other three plots show: (Top Right) misclassification error $\Rc$; (Bottom Left) majority class error $\Rc_-$; (Bottom Right) minority class error $\Rc_+$. Throughout, solid lines correspond to theoretical formulas obtained thanks to Theorem \ref{thm:main_imbalance}.}
\label{fig:figure_intro_more}
\end{figure} 

\section{Additional numerical results}\label{sec:more_num}


\subsection{Multiplicative vs Additive adjustments for label-imbalanced GMM 
}

In Figure \ref{fig:figure_intro_more} we show a more complete version of Figure \ref{fig:mismatch_intro}(a), where we additionally report standard and per-class accuracies. We minimized the CDT/LA losses in the separable regime with normalized gradient descent (GD), which uses  increasing learning rate appropriately normalized by the loss-gradient norm for faster convergence; refer to Figure \ref{fig:normalized_gd_experiments} and Section \ref{sec:gd_num_app} for the advantages over constant learning rate. Here,  normalized GD was ran until the norm of the gradient of the loss becomes less than $10^{-8}$. We observed empirically that the GD on the LA-loss reaches the stopping criteria faster compared to the CDT-loss. This is in full agreement with the CIFAR-10 experiments in Section \ref{sec:insights} and theoretical findings in Section \ref{sec:insights}.

In all cases, we reported both the results of Monte Carlo simulations, as well as, the theoretical formulas predicted by Theorem \ref{thm:main_imbalance}. As promised, the theorem sharply predicts the conditional error probabilities of minority/majority class despite the moderate dimension of $d=300$.

As noted in Section \ref{sec:insights}, CDT-loss results in better balanced error (see `Top Left') in the separable regime (where $\Rc_{\rm train}=0$) compared to LA-loss. This naturally comes at a cost, as the role of the two losses is reversed in terms of the misclassification error (see `Top Right'). The two bottom figures better explain these, by showing that VS sacrifices the error of majority class for a significant drop in the error of the minority class. All types of errors decrease with increasing overparameterization ratio $\gamma$ due to the mismatch feature model.

Finally, while balanced-error performance of CDT-loss is clearly better  compared to the LA-loss in the separable regime, the additive offsets $\iota_y$'s improve performance in the non-separable regime. Specifically, the figure confirms experimentally the superiority of the tuning of the LA-loss in \cite{Menon} compared to that in \cite{TengyuMa} (but only in the underparameterized regime). Also, it confirms our message: VS-loss that combines the best of two worlds by using both additive and multiplicative adjustments.



\subsection{Multiplicative vs Additive adjustments with $\ell_2$-regularized GD }
In this section we shed more light on the experiments presented in Figure \ref{fig:mismatch_intro}(b,c), by studying the effect of $\ell_2$-regularization.
Specifically, we repeat here the experiment of Fig. \ref{fig:mismatch_intro}(b) with $p=d=50,n=30$. We train with CE, CDT, and LA-losses in TPT with a weight-decay implementation of  $\ell_2$-regularization, that is GD with update step:
$
	\w_{t+1} = (1-\beta)\w_t-\eta\nabla_{\w}\Lc(\w_t),
$
where $\beta$ is the weight-decay factor and we used $\beta\in\{0,10^{-3},10^{-2}\}$. 

For our discussion, recall our findings in Section \ref{sec:insights}: (i) CDT-loss trained without regularization in TPT converges to CS-SVM, thus achieving better balanced error than LA-loss converging to SVM; (ii) however, at the beginning of training, multiplicative adjustment of CDT-loss can hurt the balanced error; (iii) Additive adjustments on the other hand helped in the beginning of GD iterations but were not useful deep in TPT.

We now turn our focus to the behavior of training in presence of $\ell_2$-regularization. The weight-decay factor was kept small enough to still achieve zero training error. A few interesting observations are summarized below:
\begin{figure}[h]
	\begin{center}
		\begin{subfigure}[b]{0.24\textwidth}
			\centering
			\includegraphics[width=\textwidth]{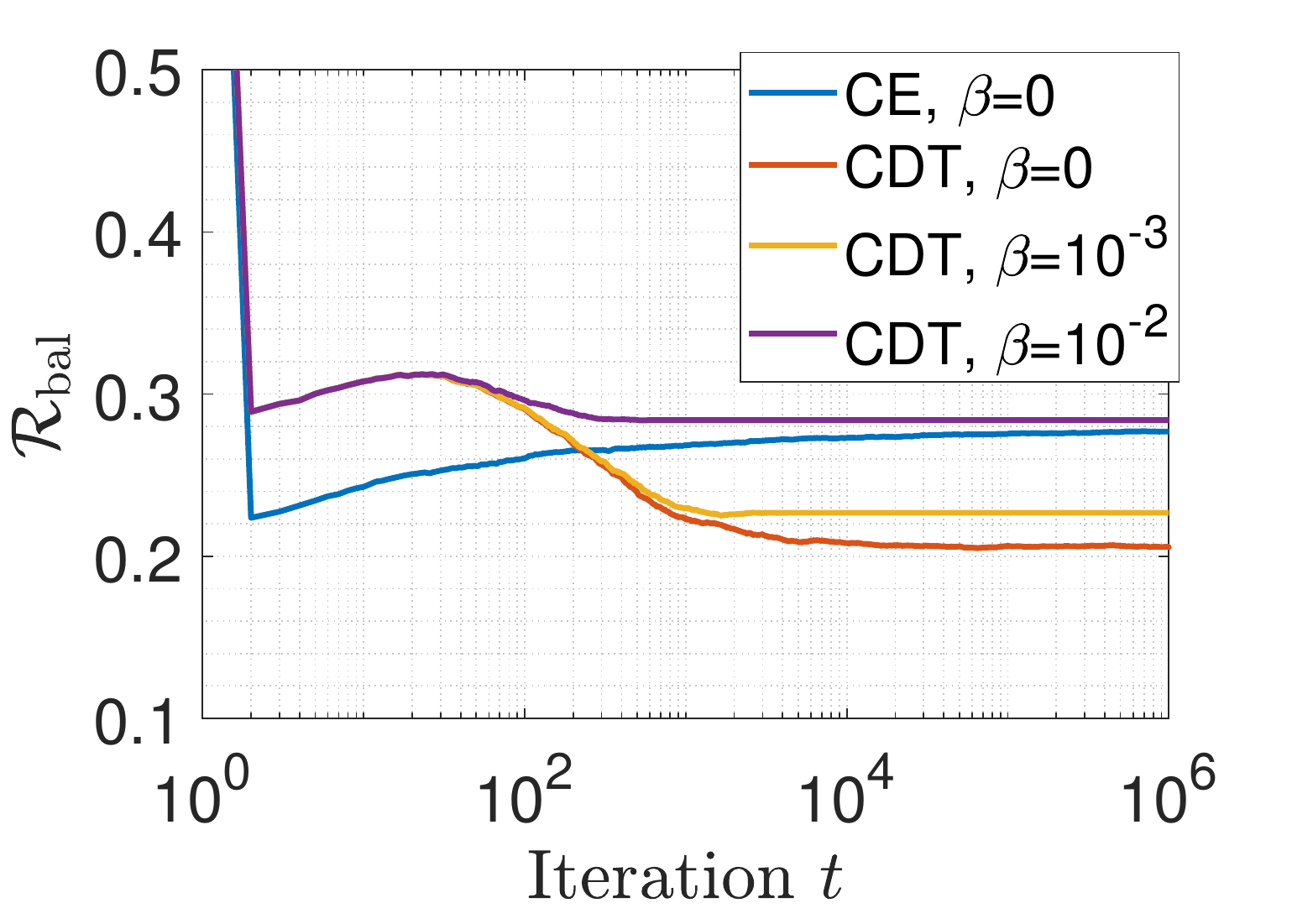}
			\caption{CDT: bal. error}
		\end{subfigure}
	\begin{subfigure}[b]{0.24\textwidth}
		\centering
		\includegraphics[width=\textwidth]{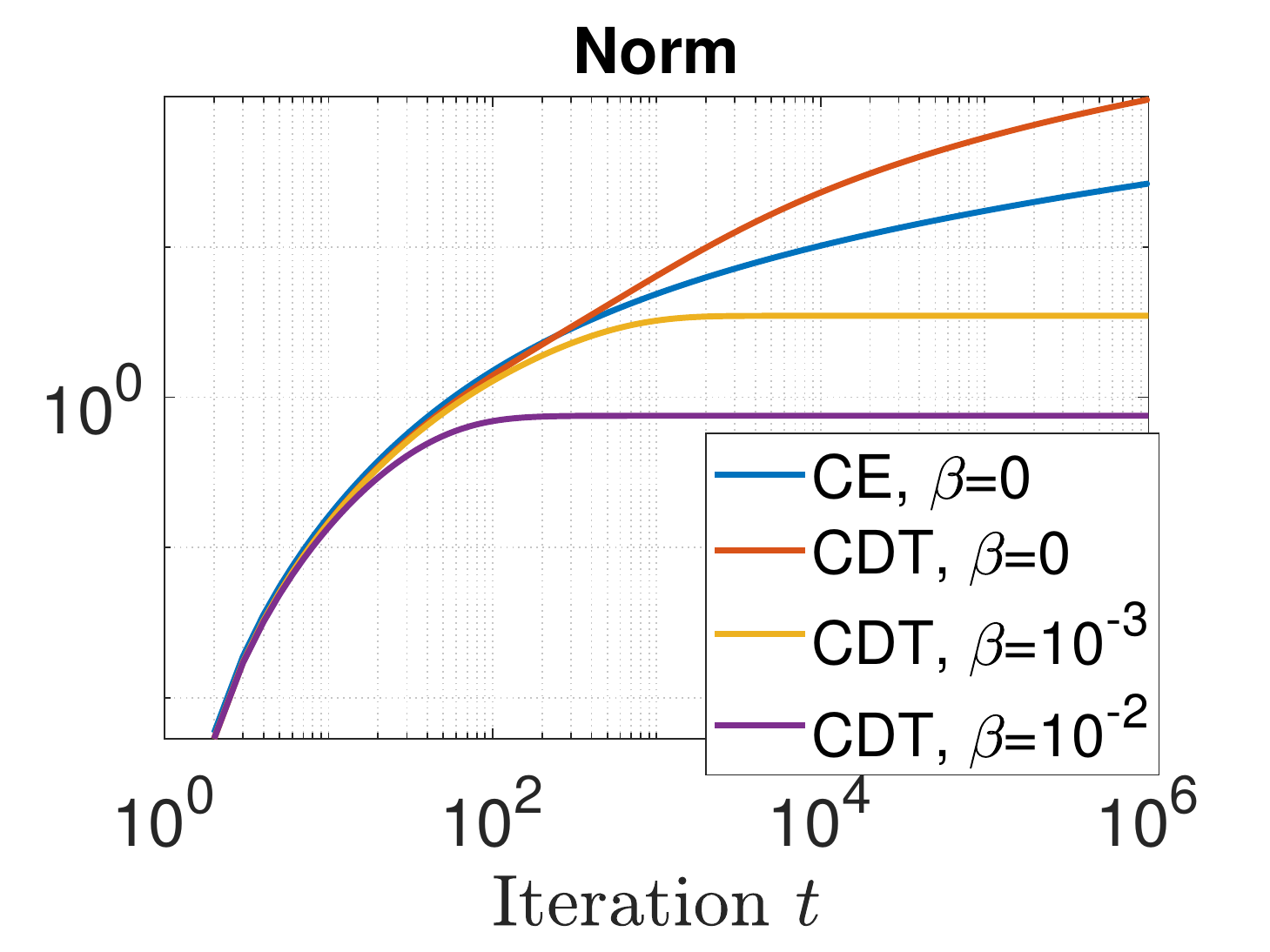}
		\caption{CDT:classifier norm}
	\end{subfigure}
		\begin{subfigure}[b]{0.24\textwidth}
			\centering
			\includegraphics[width=\textwidth]{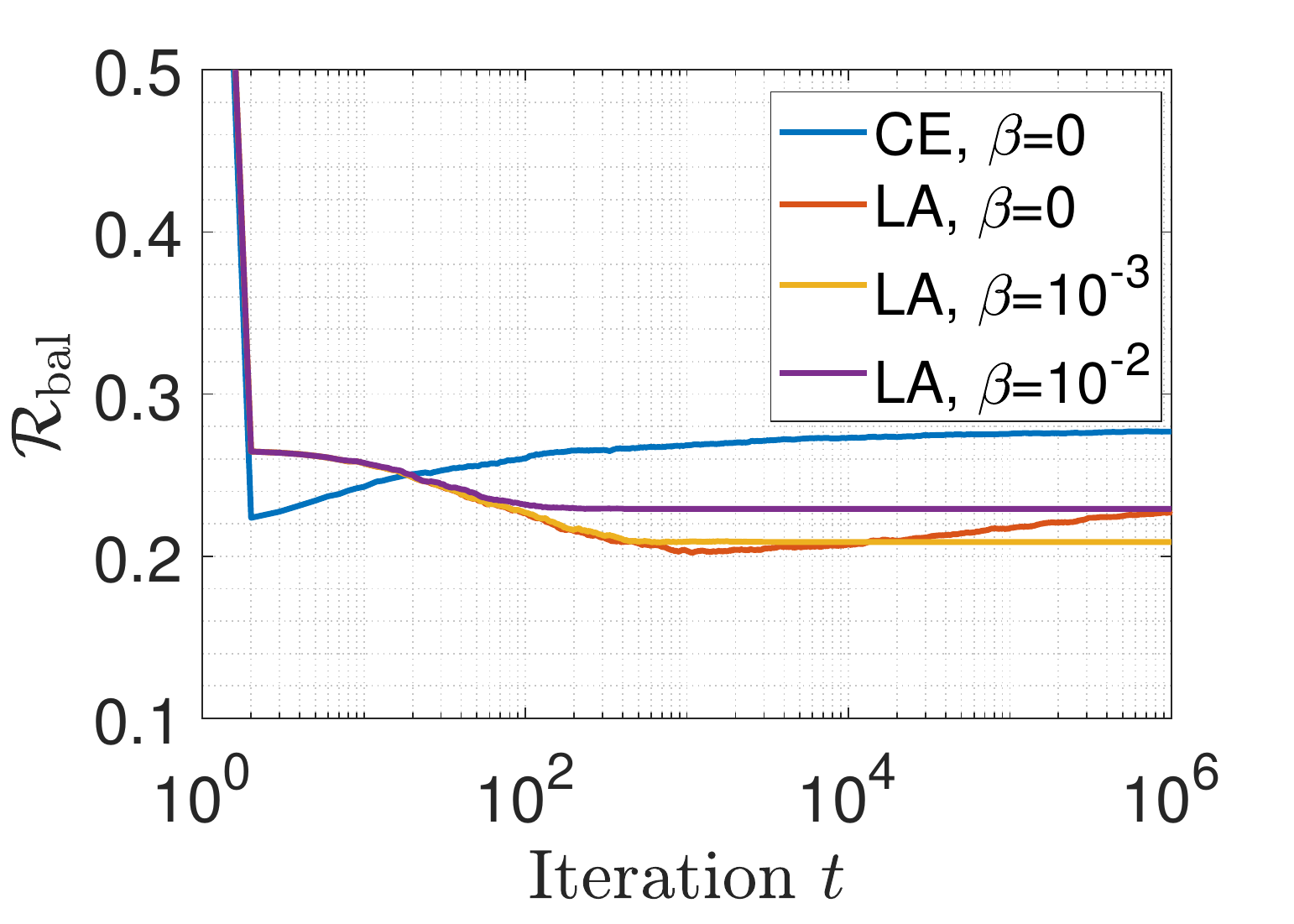}
			\caption{LA: bal. error}
		\end{subfigure}
		\begin{subfigure}[b]{0.24\textwidth}
			\centering
			\includegraphics[width=\textwidth]{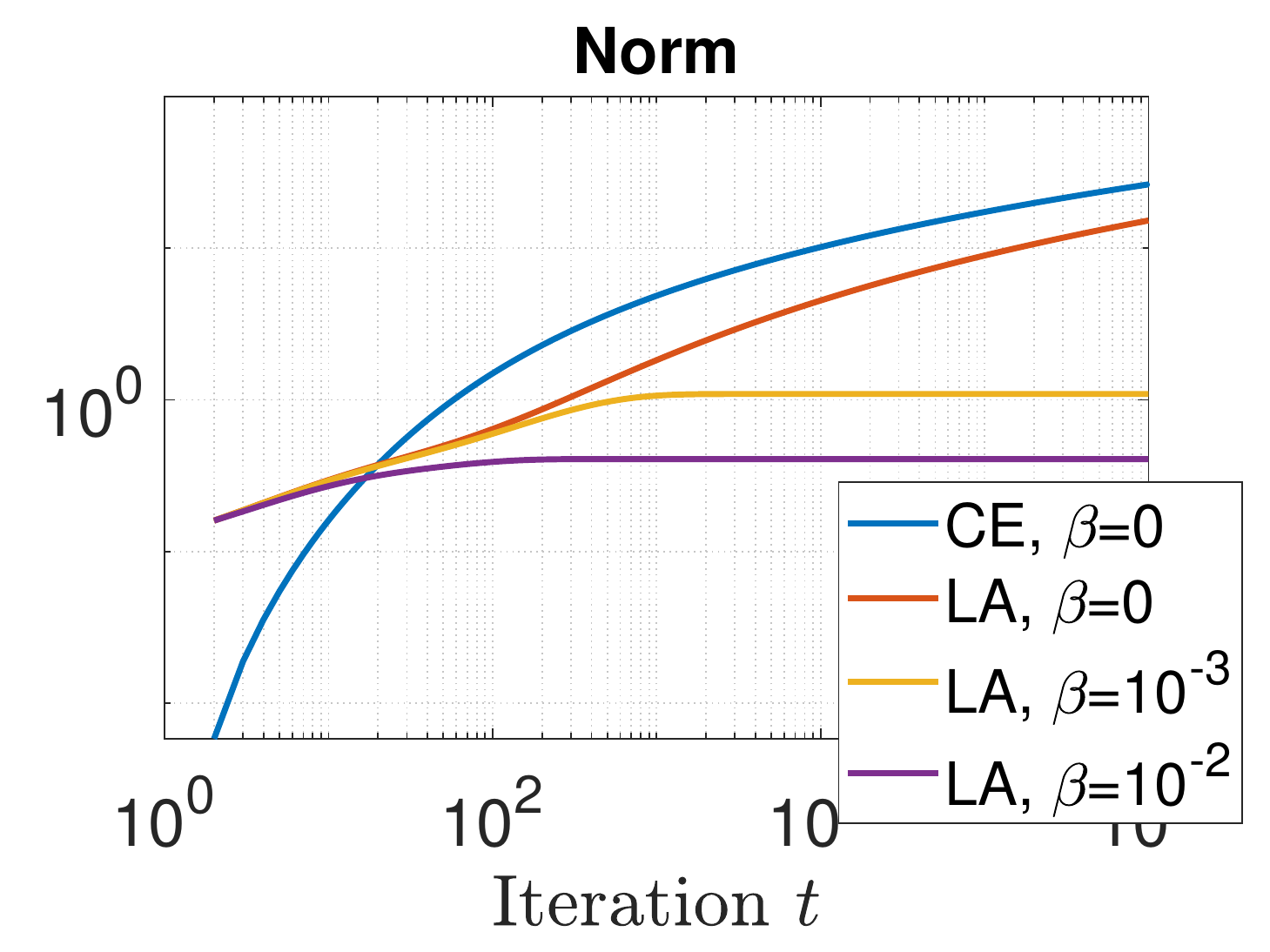}
			\caption{LA: classifier norm}
		\end{subfigure}
	\end{center}
	\caption{Training dynamics of a linear classifier trained with gradient descent on LA and CDT losses, with and without weight decay (parameter $\beta$).}
	\label{fig:gd_weight_decay}
\end{figure}
\begin{itemize}
	\item The classifier norm plateaus when trained with regularization (while it increases logarithmically without regularization; see Theorem \ref{thm:gradient_flow}). The larger the weight decay factor, the earlier the norm saturates; see Fig. \ref{fig:gd_weight_decay}(b) and (d).

%
	\item Suppose a classifier is trained with a small, but non-zero, weight decay factor in TPT, and the resulting classifier has a norm saturating at some value $\zeta>0$.
	The final balanced error performance of such a classifier closely matches the balanced error produced by a classifier trained without regularization but with training stopped early at that iteration for which the classifier-norm is equal to $\zeta$; compare for example, the value of yellow curve (CDT, $\beta=10^{-3}$) at $t=10^6$ with the value of the red curve (CDT, $\beta=0$) at around $t=300$ in Fig. \ref{fig:gd_weight_decay}(c) and (d). \footnote{See also \cite{raskutti2014early,ali2019continuous} for the connection between gradient-descent  and regularization solution paths.}
	
	\item If early-stopped (appropriately) before entering TPT, LA-loss can give better balanced performance than CDT-loss. In view of the above mentioned mapping between weight-decay and training epoch, the use of weight decay results in same behavior.
	 Overall, this supports that VS-loss, combining both additive and multiplicative adjustments is a better choice for a wide range of $\ell_2-$ regularization parameters.
\end{itemize}

\subsection{Additional information on Figures \ref{fig:mismatch_intro}(b),(c) and \ref{fig:tradeoffs}(a),(b)} \label{sec:details_intro_fig}

\noindent\textbf{Figures \ref{fig:mismatch_intro}(b,c).}~We generate data from a binary GMM with $d=50, n=30$ and $\pi=0.1$. We generate mean vectors as random iid Gaussian vector and scale their norms to $5$ and $1$, respectively. For training, we use gradient descent with constant learning rate $0.1$ and fixed number of $10^6$ iterations. The balanced test error in Figure \ref{fig:mismatch_intro}(b)  is computed by Monte Carlo on a balanced test set of $10^5$ samples. Figure \ref{fig:mismatch_intro}(c) measures the angle gap of GD outputs $\w^t$ to the solution $\hat\w_\delta$ of CS-SVM in \eqref{eq:CS-SVM} with $\delta=\delta_\star$ and $\h(\x_i)=\x_i$.

\noindent\textbf{Figures \ref{fig:tradeoffs}(a,b).}~In (a), we generated GMM data with $\|\mub_+\|=3, \mub_-=-\mub_+$  and $\pi=0.05.$  In (b), we considered the GMM of Section \ref{sec:generalization}  with $\|\mub_{y,g}\|=3,  y\in\{\pm1\}, g\in\{1,2\}$ and $\mub_{+,1}\perp\mub_{+,2} \in \R^d$, sensitive group prior $p=0.05$ and equal class priors $\pi=1/2$. 

\input{convergence_group}

\input{waterbirds_experiments}
\subsection{Validity of theoretical performance analysis} 
\begin{figure}[t]
	\begin{center}
			\includegraphics[width=0.38\textwidth,height=0.25\textwidth]{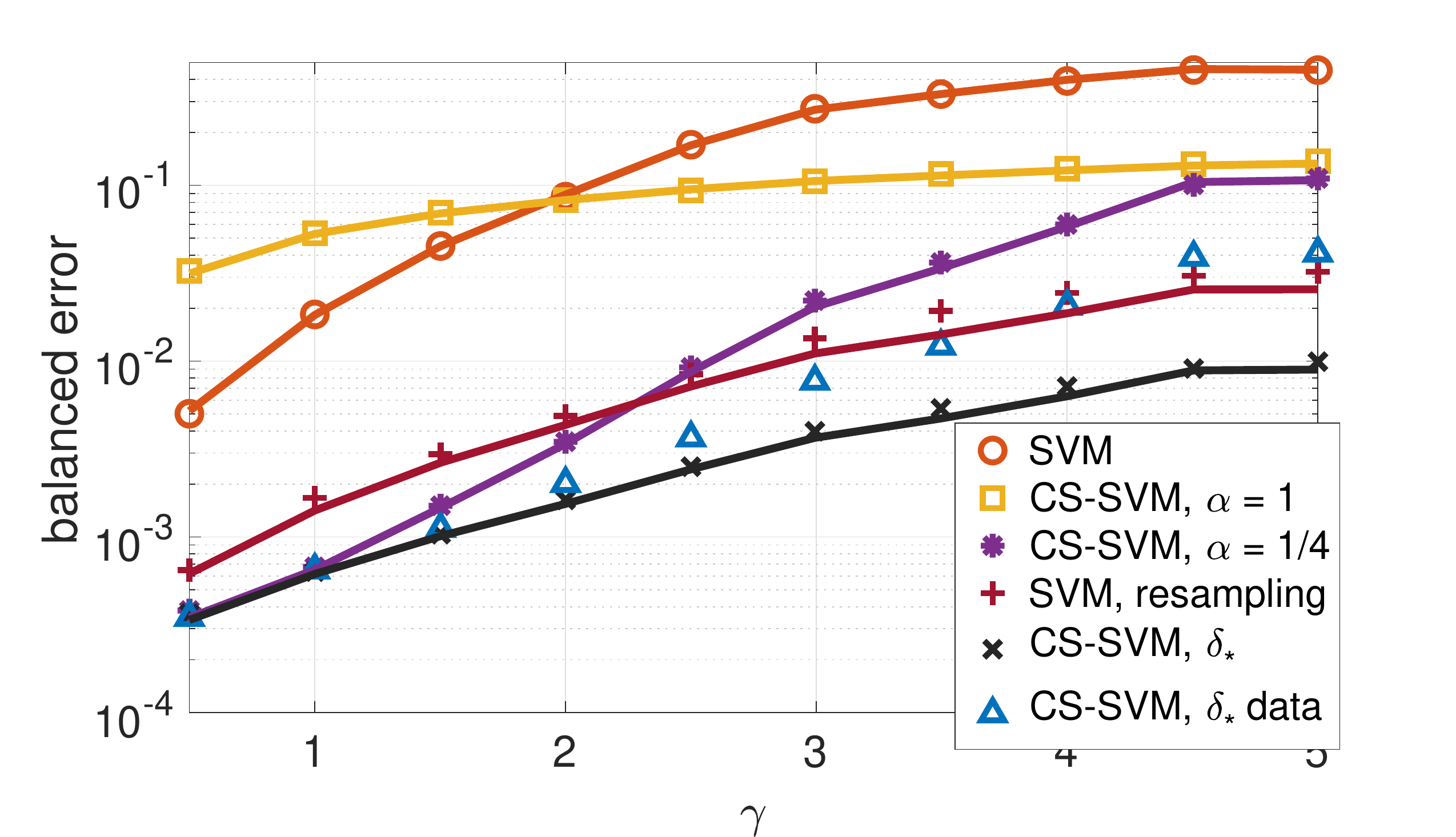}
		\includegraphics[width=0.38\textwidth,height=0.25\textwidth]{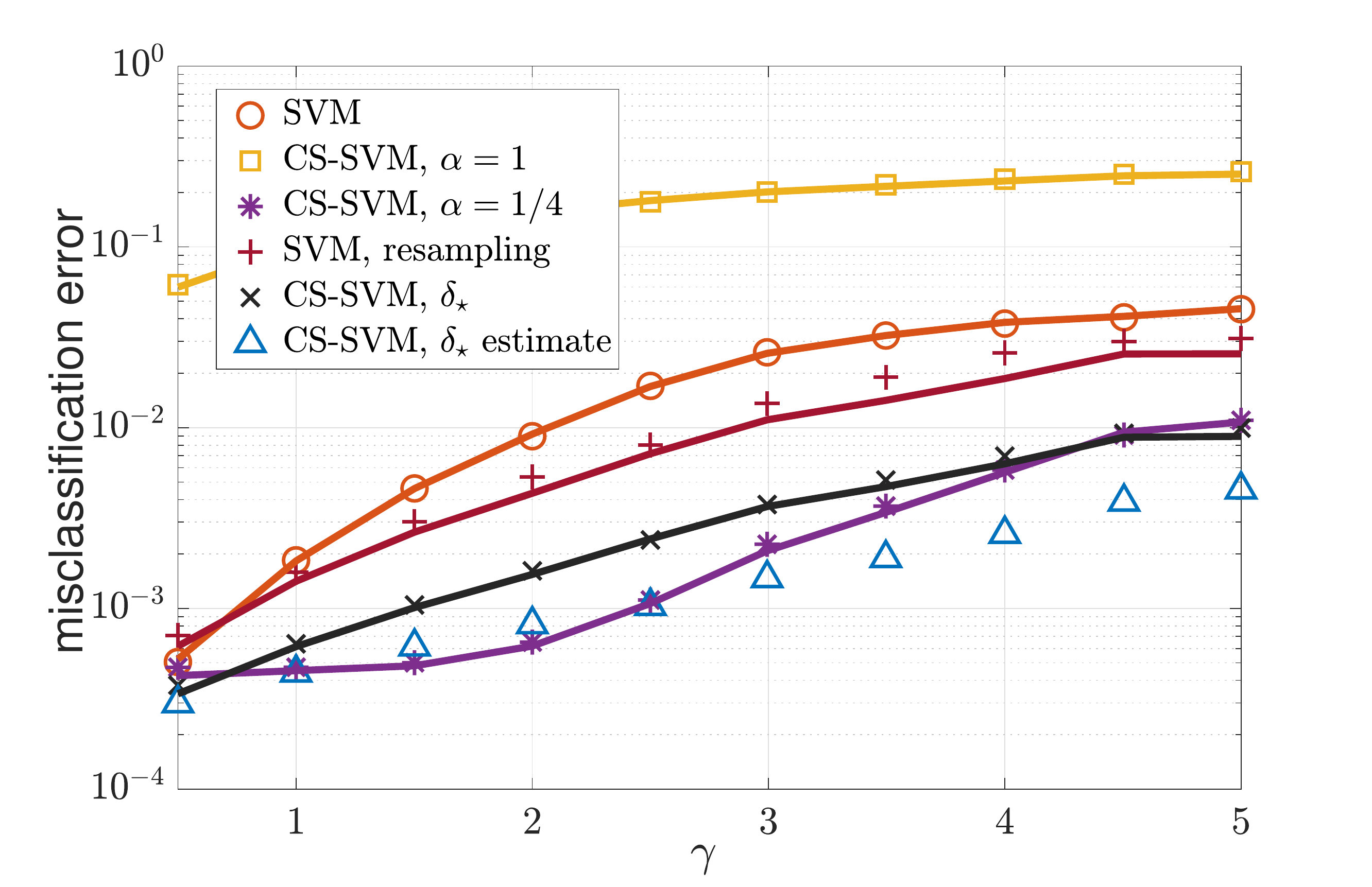}
	\end{center}
	\caption{Balanced (Left) and misclassification (Right) errors as a function of the parameterization ratio $\gamma=d/n$ for the following algorithms: SVM with and without majority class resampling, CS-SVM with different choices of $\delta=\big(\frac{1-\pi}{\pi}\big)^\alpha, \pi=0.05$ and $\delta=\delta_\star$ (cf. Eqn. \eqref{eq:delta_star}) plotted for different values of $\gamma=d/n$. {Solid lines show the theoretical values thanks to Theorem \ref{thm:main_imbalance} and the discrete markers represent empirical errors over 100 realizations of the dataset.} Data were generated from a GMM with $\mub_{+}=4\eb_1, \mub_{-}=-\mub_{+} \in\R^{500}$, and $\pi=0.05$. SVM with resampling outperforms SVM without resampling in terms of balanced error, but the optimally tuned CS-SVM is superior to both in terms of both balanced and misclassification errors for all values of $\gamma$.
	} 
	\label{fig:cls_imbalance_antipodal_misclass_errors}
\end{figure}

\begin{figure}[t!]
	\begin{center}
		\includegraphics[width=0.5\textwidth]{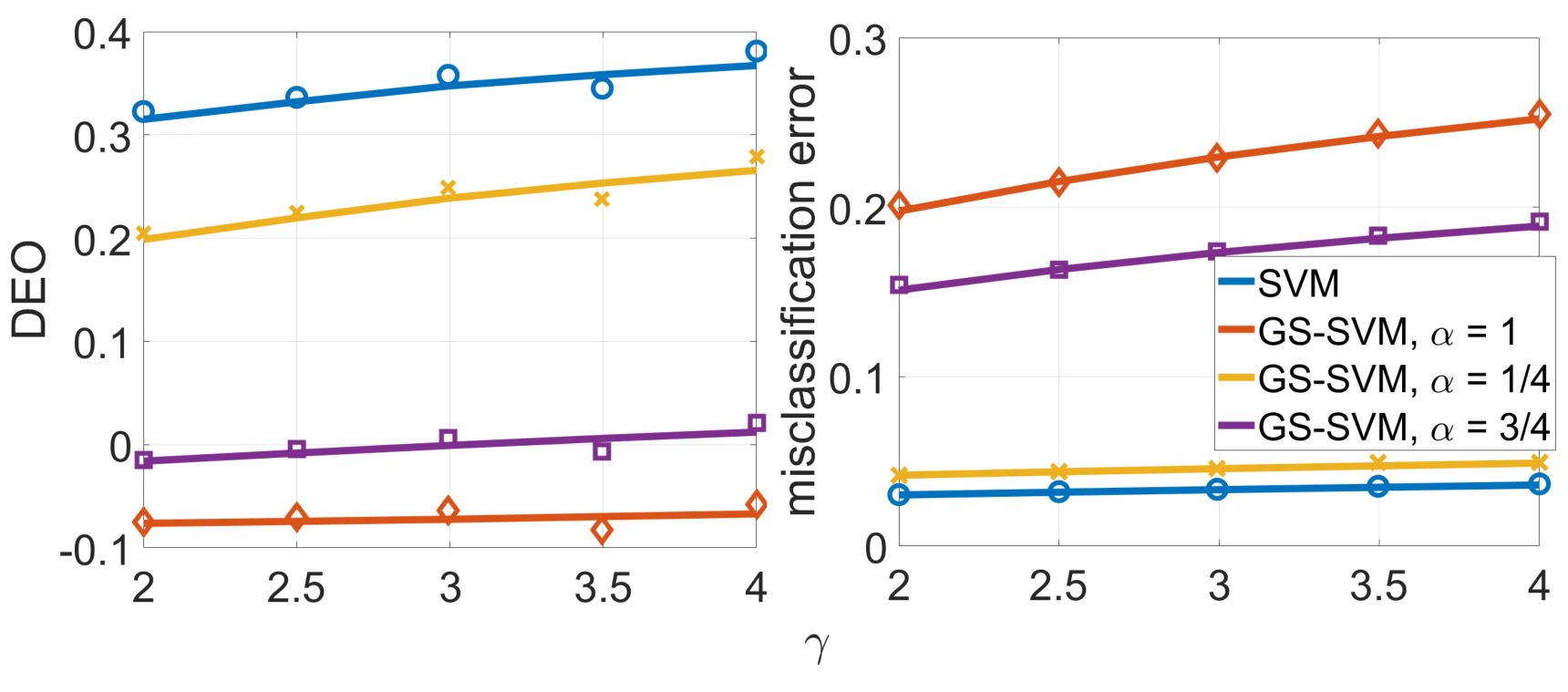}
	\end{center}
	\caption{DEO and misclassification error of SVM and GS-SVM with different choices of $\delta=\big(\frac{1-p}{p}\big)^\alpha$ for minority group prior $p=0.05$ plotted against  $\gamma=d/n$. {Solid lines show the theoretical values and the discrete markers represent empirical errors over 100 realizations of the dataset.} Data generated from a GMM with $\mub_{+,1}=3\eb_1, \mub_{+,2}=3\eb_2 \in\R^{500}$. While SVM has the least misclassification error, it suffers from a high DEO. By trading off misclassification error, it is possible to tune GS-SVM (specifically, $\alpha=0.75$) so that it achieves DEO close to 0 for all the values of $\gamma$ considered here. } 
	\label{fig:group_fairness_deo_misclas}
\end{figure}

Figures \ref{fig:cls_imbalance_antipodal_misclass_errors} and \ref{fig:group_fairness_deo_misclas} demonstrate that our Theorems \ref{thm:main_imbalance} and \ref{thm:main_group_fairness} provide remarkably precise prediction of the GMM performance even when dimensions are in the order of hundreds. Moreover, both figures show the clear advantage of CS/GS-SVM over regular SVM and naive resampling strategies in terms of balanced error and equal opportunity, respectively.

The reported values for the misclassification error and the balanced error / DEO were computed over $10^5$ test samples drawn from the same distribution as the training examples.


Additionally, Figure \ref{fig:cls_imbalance_antipodal_misclass_errors} validates the explicit formula that we derive in Equation \eqref{eq:delta_star} for $\delta_\star$ minimizing the balanced error. 
Specifically, observe that  CS-SVM with $\delta=\delta_\star$ (`$\times$' markers)  not only minimizes balanced error (as predicted in Section \ref{sec:delta_star_proof}), but also leads to better misclassification error compared to SVM for all depicted values of $\gamma$. The figure also shows the performance of our data-dependent heuristic of computing $\delta_\star$ introduced in Section \ref{sec:delta_star_estimate}. The heuristic appears to be accurate for small values of $\gamma$ and is still better in terms of balanced error compared to the other two heuristic choices of $\delta=(\frac{1-\pi}{\pi})^{\alpha}, \alpha=1/4,1$. Finally,  we also evaluated the SVM+subsampling algorithm; see Section \ref{sec:undersampling} below for the algorithm's description and performance analysis.  Observe that SVM+resampling outperforms SVM without resampling in terms of balanced error, but the optimally tuned CS-SVM is superior to both.

\subsubsection{Max-margin SVM with random majority class undersampling}\label{sec:undersampling}

For completeness, we briefly discuss here SVM combined with undersampling, a popular technique that first randomly undersamples majority examples and only then trains max-margin SVM. The asymptotic performance of this scheme under GMM can be analyzed using Theorem \ref{thm:main_imbalance} as explained below. 

Suppose the majority class is randomly undersampled to ensure equal size of the two classes. This increases the effective overparameterization ratio by a factor of $\frac{1}{2\pi}$ (in the asymptotic limits). In particular, the conditional risks converge as follows:
\begin{align}
	\Rc_{+,\text{undersampling}}(\gamma,\pi)&\rP \Rco_{+,\text{undersampling}}(\gamma,\pi)=\Rco_+\big(\frac{\gamma}{2\pi},0.5\big)\nonumber\\\Rc_{-,\text{undersampling}}(\gamma,\pi)&\rP\Rco_{-,\text{undersampling}}(\gamma,\pi)=\Rco_{+,\text{undersampling}}(\gamma,\pi).
\end{align} 
Above, $\Rc_{+,\text{undersampling}}$ and $\Rc_{-,\text{undersampling}}$ are the class-conditional risks of max-margin SVM after random undersampling of the majority class to ensure equal number of training examples from the two classes. The risk $\Rco_+\big(\frac{\gamma}{2\pi},0.5\big)$ is the asymptotic conditional risk of a \emph{balanced} dataset with overparameterization ratio $\frac{\gamma}{2\pi}$. This is computed as instructed in Theorem \ref{thm:main_imbalance} for the assignments $\gamma\leftarrow \frac{\gamma}{2\pi}$ and $\pi\leftarrow 1/2$ in the formulas therein.

Our numerical simulations in Figure \ref{fig:cls_imbalance_antipodal_misclass_errors} verify the above formulas.

%% file: convergence_group.tex
\begin{figure}[t]
	\begin{center}
		\begin{subfigure}[b]{0.45\textwidth}
			\centering
			\includegraphics[width=0.85\textwidth,height=0.45\textwidth]{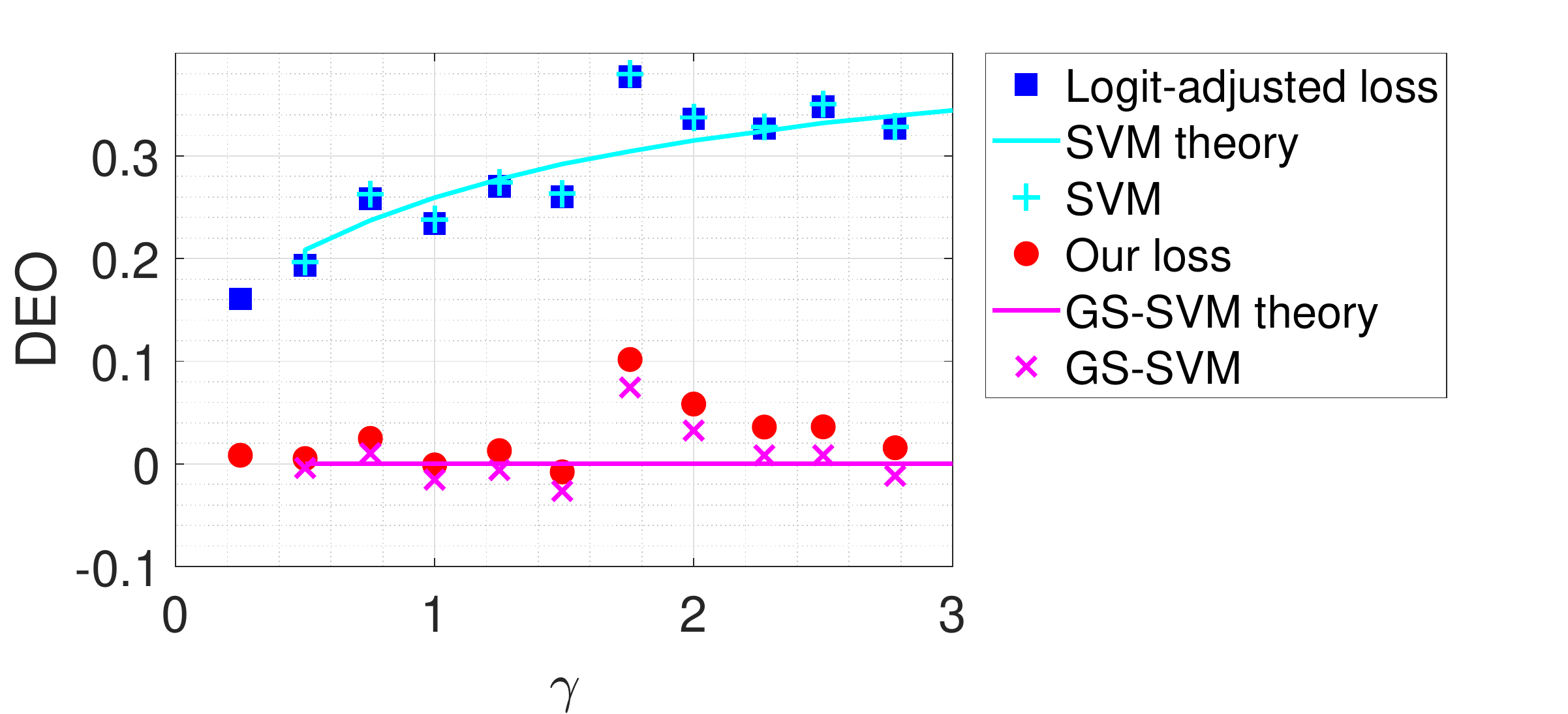}
			\caption{DEO}
		\end{subfigure}
		\begin{subfigure}[b]{0.45\textwidth}
			\centering
			\includegraphics[width=0.85\textwidth,height=0.45\textwidth]{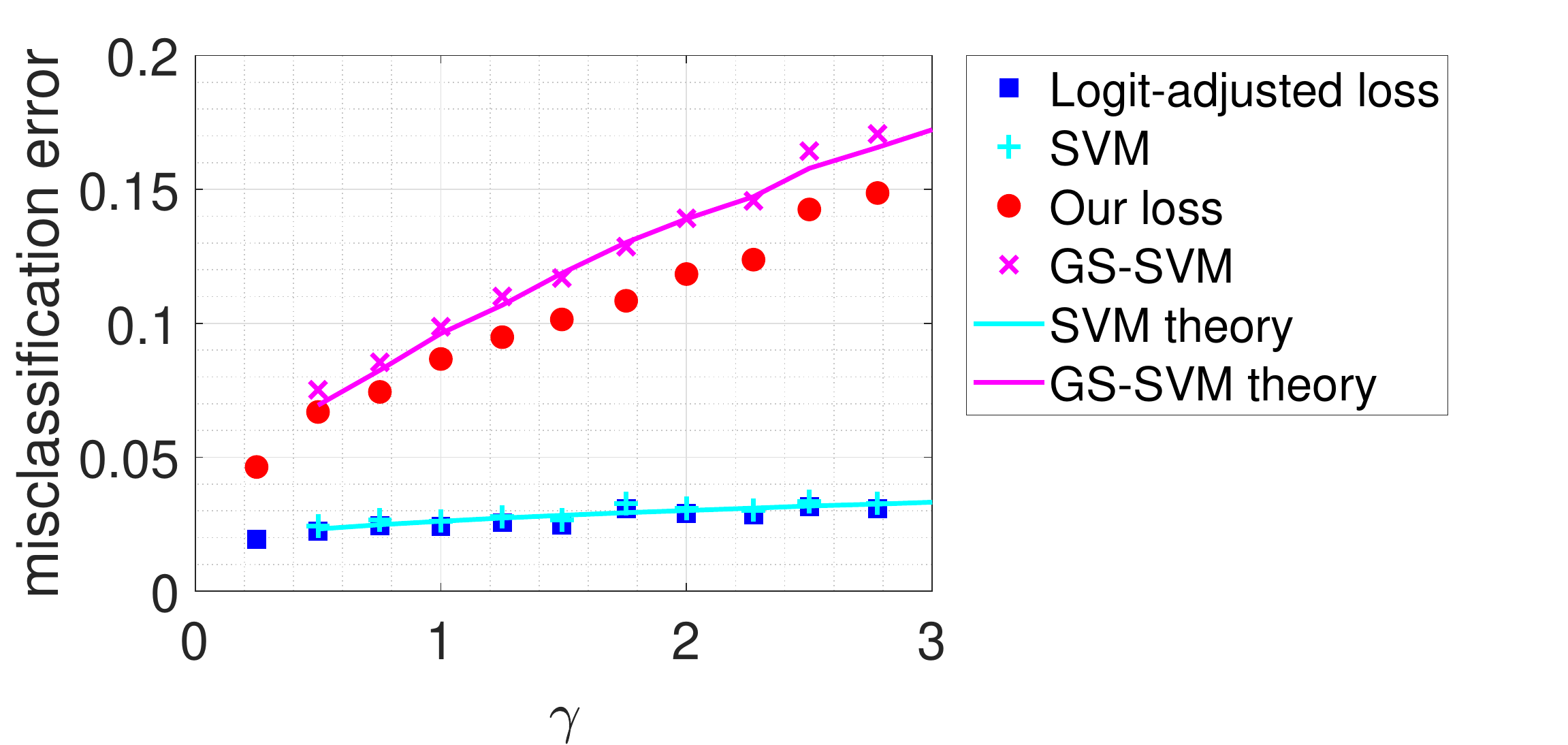}
			\caption{Misclassification error}
		\end{subfigure}
		\begin{subfigure}[b]{0.45\textwidth}
		\centering
		\includegraphics[width=0.5\textwidth]{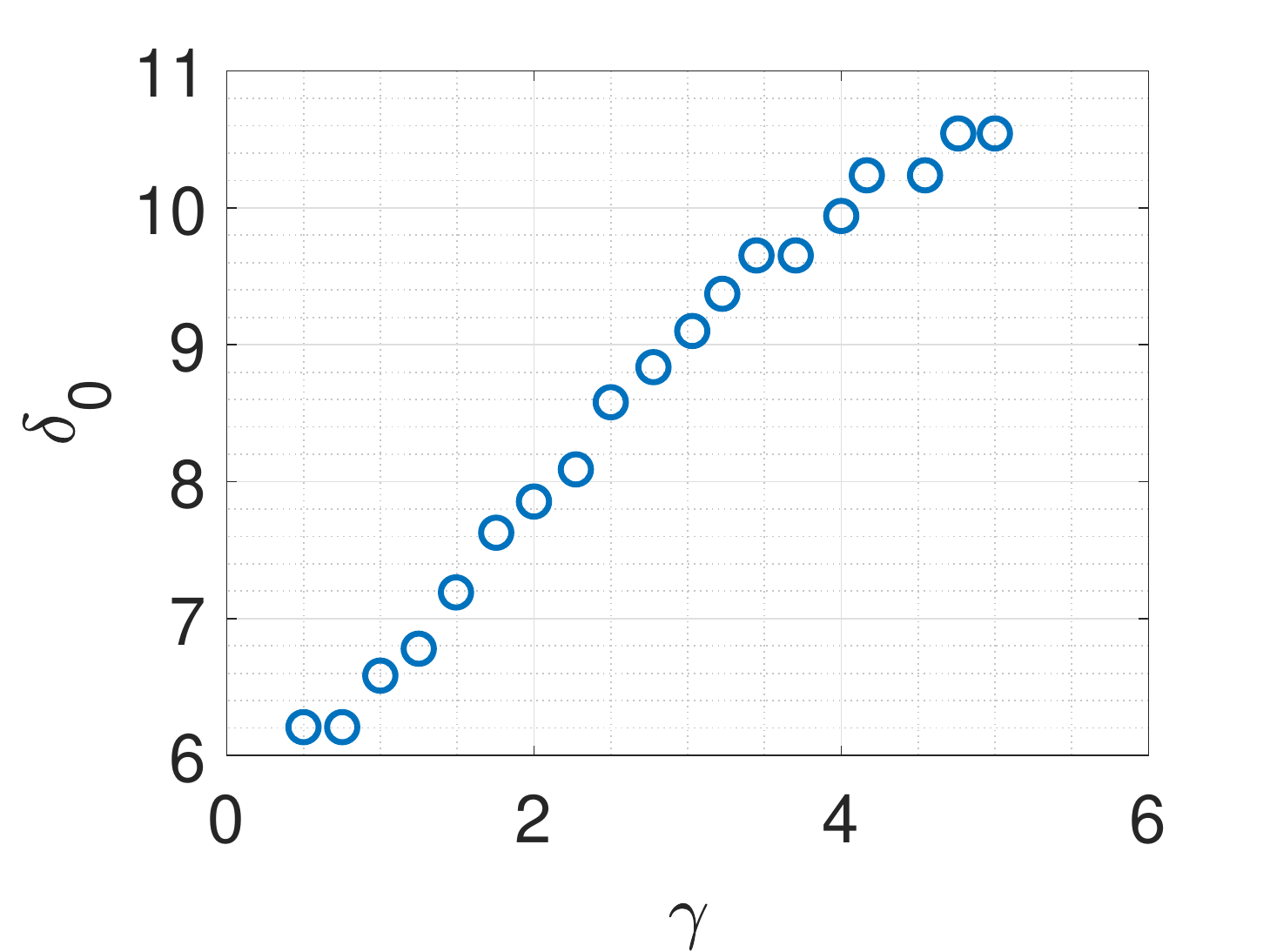}
		\caption{$\delta_0$}
	\end{subfigure}
	\end{center}
	\caption{This figure highlights the benefits of our theory-inspired VS-loss and GS-SVM over regular SVM and logit-adjusted loss in a group-sensitive classification setting. We trained a linear model with varying number $n$ of examples in $\R^{d=100}$, of a binary Gaussian-mixture dataset with two groups. $x$-axis is the parameterization ratio $d/n$. Data were generated from a GMM with prior $p=0.05$ for the minority group. For $\gamma>0.5$, we train additionally using SVM (cyan plus marker) and group-sensitive SVM (magenta cross). The plot (c) displays the parameter $\delta=\delta_0$ that we used to tune the VS-loss and GS-SVM. These values were obtained through a grid search from the theoretical prediction such that the theoretical $\Rdeo$ (cf. Theorem \ref{thm:main_group_fairness}) produced by the corresponding GS-SVM is $0$. The solid lines depict theoretical predictions obtained by Theorem \ref{thm:main_group_fairness}. The empirical probabilities were computed by averaging over 25 independent realizations of the training and test data.}
	\label{fig:gs_deo_gd_convergence}
\end{figure}

\subsection{VS-loss vs LA-loss for a group-sensitive GMM}

In Figure \ref{fig:gs_deo_gd_convergence} we test the performance of our theory-inspired VS-loss against the logit-adjusted (LA)-loss in a group-sensitive classification setting with data from a Gaussian mixture model with a minority and and a majority group. Specifically, we generated synthetic data from the model with class prior $\pi=1-\pi=1/2$, minority group membership prior $p=0.05$ (for group $g=1$) and $\mub_{1}=3\eb_1, \mub_{2}=3\eb_2 \in\R^{500}$. We trained homogeneous linear classifiers based on a varying number of training sample $n=d/\gamma$. For each value of $n$ (eqv. $\gamma$) we ran normalized gradient descent (see Sec. \ref{sec:gd_num_app}) on 
\begin{itemize}
\item CDT-loss $\ell(y,\w^T\x,g):=\log(1+e^{-\Delta_g y(\w^T\x)})$ with $\Delta_g=\delta_0\ind{g=1}+\ind{g=2}$.
\item the  LA-loss modified for group-sensitive classification $\ell(y,\w^T\x,g):=\log(1+e^{\iota_g}e^{ y(\w^T\x)})$ with $\iota_g=p^{-1/4}\ind{g=1}+(1-p)^{-1/4}\ind{g=2}$. This value for $\iota$ is inspired by \cite{TengyuMa}, but that paper only considered applying the LA-loss in label-imbalanced settings.
\end{itemize}
For $\gamma>0.5$ where data are necessarily separable, we also ran the standard SVM and the GS-SVM with $\delta=\delta_0$.

Here, we chose the parameter $\delta_0$ such that the GS-SVM achieves zero DEO. To do this, we used the theoretical predictions of Theorem \ref{thm:main_group_fairness} for the DEO of GS-SVM for any value of $\delta$ and performed a grid-search giving us the desired $\delta_0$; see Figure \ref{fig:gs_deo_gd_convergence} for the values of $\delta_0$ for different values of $\gamma$. 

Figure \ref{fig:gs_deo_gd_convergence}(a) verifies that the GS-SVM achieves DEO (very close to) zero on the generated data despite the finite dimensions in the simulations. On the other hand, SVM has worse DEO performance. In fact, the DEO of SVM increases with $\gamma$, while that of GS-SVM stays zero by appropriately tuning $\delta_0$.

The figure further confirms the message of Theorem \ref{thm:implicit_loss_gen}:
%
%
In the separable regime, GD on logit-adjusted loss converges to the standard SVM performance, whereas GD on our VS-loss converges to the corresponding GS-SVM solution, thus allowing to tune a suitable $\delta$ that can trade-off misclassification error to smaller DEO magnitudes. The stopping criterion of GD was a tolerance value on the norm of the gradient. The match between empirical values and the theoretical predictions improves with increase in the dimension, more Monte-Carlo averaging and a stricter stopping criterion for GD.

%% file: imbalanced_proofs.tex
\section{Margin properties and implicit bias of VS-loss}\label{sec:propo_proof}

\subsection{A more general version and proof of Theorem \ref{propo:gd}}\label{sec:gen_gd}

We will state and prove a more general theorem to which Theorem \ref{propo:gd} is a corollary. The new theorem also shows that the group-sensitive adjusted VS-loss in \eqref{eq:loss_ours_bin_group} converges to  the GS-SVM.

\begin{remark} Theorem \ref{propo:gd} and the content of this section are true for \emph{arbitrary} linear models $\f_\w(\x)=\inp{h(\x)}{\w}$ and  feature maps $h:\Xc\rightarrow\R^p.$ To lighten notation in the proofs, we assume for simplicity that $h$ is the identity map, that is $\h(\x)=\x.$ For the general case, just substitute the raw features $\x_i\in\Xc$ below with their feature representation $h(\x_i)\in\R^p$.
\end{remark}



Consider the VS-loss empirical risk minimization (cf. \eqref{eq:loss_ours_bin} with $f(\x)=\w^T\x$):
\begin{align}\label{eq:loss_app}
\Lc(\w):=\sum_{i\in[n]}\ell(y_i,\w^T\x_i,g_i):=\om_i\log\left(1+e^{\iota_i}\cdot e^{- \Del_i y_i(\w^T\x_i)}\right).
\end{align}
for strictly positive (but otherwise arbitrary) parameters $\Delta_i, \omega_i > 0$ and arbitrary $\iota_{i}$. For example, setting $\om_i=\omega_{y_i,g_i}, \Del_i = \Delta_{y_i,g_i}$ and $\iota_i=\iota_{y_i,g_i}$ recovers the general form of our binary VS-loss in \eqref{eq:loss_ours_bin_group}.

Also, consider the following general cost-sensitive SVM (to which both the CS-SVM and the GS-SVM are special instances)
\begin{align}\label{eq:CS-SVM_app}
\wh:= \arg\min_\w \|\w\|_2\quad\text{subject to}~~y_i(\w^T\x_i)\geq {1}\big/{\Del_i}, \forall i\in[n].
\end{align}

First, we state the following simple facts about the cost-sensitive max-margin classifier in \eqref{eq:CS-SVM_app}. The proof of this claim is rather standard and is included in Section \ref{sec:CS_svm_easy_claim} for completeness. 

\begin{lemma}\label{lem:CS_svm_easy_claim}
Assume that the training dataset is linearly separable, i.e. $\exists \w$ such that $y_i(\w^T\x_i)\geq 1$ for all $i\in[n]$. Then, \eqref{eq:CS-SVM_app} is feasible.  Moreover, letting $\wh$ be the solution of \eqref{eq:CS-SVM_app}, it holds that 
\begin{align}\label{eq:CS_svm_easy_claim}
 \frac{\hat\w}{\|\hat\w\|_2} = \arg\max_{\|\w\|_2=1}\,\min_{i\in[n]}\, \Delta_{i}y_i\x_i^T\w.
\end{align}
\end{lemma}

Next, we state the main result of this section connecting the VS-loss in \eqref{eq:loss_app} to the max-margin classifier in \eqref{eq:CS-SVM_app}. After its statement, we show how it leads to Theorem \ref{propo:gd}; its proof is given later in Section \ref{sec:proof_gd}.  

{
\begin{theorem}[Margin properties of VS-loss: General result]\label{thm:implicit_loss_gen}
Define the norm-constrained optimal classifier
\begin{align}\label{eq:con_ERM}
\w_R:=\arg\min_{\|{\w}\|_2\leq R}\Lc(\w),
\end{align}
with the loss $\Lc$ as defined in \eqref{eq:loss_app} for positive (but otherwise arbitrary) parameters $\Delta_i, \omega_i>0$ and arbitrary $\iota_i$. 
Assume that the training dataset is linearly separable
and let $\wh$ be the solution of \eqref{eq:CS-SVM_app}.
Then, it holds that 
\begin{align}\label{eq:implicit_2_prove}
\lim_{R\rightarrow\infty}\frac{\w_R}{\|\w_R\|_2} = \frac{\hat\w}{\|\hat\w\|_2} \,.
\end{align}
\end{theorem}
}

\subsubsection{Proof of Theorem \ref{propo:gd}}
Theorem \ref{propo:gd} is a corollary of Theorem \ref{thm:implicit_loss_gen} by setting $\om_i=\om_{y_i}$, $\iota_i=\iota_{y_i}$ and $\Delta_i=\Delta_{y_i}$. Indeed for this choice the loss in Equation \eqref{eq:loss_app} reduces to that in Equation \eqref{eq:loss_ours_bin}. Also, \eqref{eq:CS-SVM_app} reduces to \eqref{eq:CS-SVM}. The latter follows from the equivalence of the following two optimization problems:
\begin{align*}
&\big\{\,\,\arg\min_\w \|\w\|_2\quad\text{subject to}~~\w^T\x_i \begin{cases}\geq 1/\Del_+ & y_i=+1 \\ \leq -1/\Del_- & y_i=-1\end{cases}\,\, \big\}~~
\\
&\qquad\qquad\qquad=~~\big\{\,\,\arg\min_\vb \|\vb\|_2\quad\text{subject to}~~\vb^T\x_i \begin{cases} \geq\Del_-/\Del_+ & y_i=+1 \\ \leq-1 & y_i=-1\end{cases} \,\,\big\},
\end{align*}
which can be verified simply by a change of variables $\vb/\Del_-\leftrightarrow\w$ and $\Del_->0$.

\vp
\noindent{\textbf{The case of group-sensitive VS-loss.}~}
As another immediate corollary of Theorem \ref{thm:implicit_loss_gen} we get an analogue of Theorem \ref{propo:gd} for a group-imbalance data setting with $K=2$ and balanced classes. Then, we may use the VS-loss in \eqref{eq:loss_app} with margin parameters $\Delta_{i}=\Delta_g, g=1,2$.
From Theorem \ref{thm:implicit_loss_gen}, we know that in the separable regime and in the limit of increasing weights, the classifier $\w_R$ (normalized) will converge to the solution of the GS-SVM with $\delta=\Delta_2/\Delta_1.$

\subsubsection{Proof of Theorem \ref{thm:implicit_loss_gen}}\label{sec:proof_gd}


{First, we will argue that for any $R>0$ the solution to the constrained VS-loss minimization is on the boundary, i.e. 
\begin{align}\label{eq:boundary}
\|\w_R\|_2=R.
\end{align}
 We will prove this by contradiction. Assume to the contrary that $\w_R$ is a point in the strict interior of the feasible set. It must then be by convexity that $\nabla \Lc(\w_R)=0$. Let $\wt$ be any solution feasible in \eqref{eq:CS-SVM_app} (which exists as shown above) such that $y_i(\x_i^T\wt)\geq 1/\Del_i$. On one hand, we have $\wt^T\nabla \Lc(\w_R)=0$. On the other hand, by positivity of $\om_i,\Del_i, \forall i\in[n]$: 
\begin{align}\label{eq:no_finite_cp}
\wt^T\nabla \Lc(\w_R) = \sum_{i\in[n]}\underbrace{\frac{-\omega_{i} \Delta_{i}e^{-\Delta_i y_i\x_i^T\w_R+\iota_i}}{ 1+ e^{\iota_{i}}e^{-\Delta_i y_i\x_i^T\w_R}}}_{<0}\, \underbrace{y_i\wt^T\x_i}_{>0} <0,
\end{align}
which leads to a contradiction.}

Now, suppose that \eqref{eq:implicit_2_prove} is not true. This means that there is some $\eps_0>0$ such that there is always an arbitrarily large $R>0$ such that $\frac{\w_R^T\wh}{\|\w_R\|_2\|\wh\|_2}\leq 1-\eps_0$. Equivalently, (in view of \eqref{eq:boundary}):
\begin{align}\label{eq:corr_gd}
\frac{\w_R^T\wh}{R\|\wh\|_2}\leq 1-\eps_0.
\end{align}

 Towards proving a contradiction, we will show that, in this scenario using $\wh_R=R\frac{\wh}{\tn{\wh}}$ yields a strictly smaller VS-loss (for sufficiently large $R>0$), i.e. 
  \begin{align}
 \Lc(\wh_R) < \Lc(\w_R),\qquad\text{for sufficiently large $R$}.
 \end{align}

We start by upper bounding $\Lc(\wh_R)$. To do this, we first note from definition of $\wh_R$ the following margin property: 
\begin{align}\label{eq:key_margin}
y_i\wh_R^T\x_i = \frac{R}{\|\hat\w\|_2} y_i\wh^T\x_i \geq  \frac{R}{\|\hat\w\|_2} (1/\Del_i)=:\frac{\bar{R}}{\Del_i},
\end{align}
where the inequality follows from feasibility of $\wh$ in \eqref{eq:CS-SVM_app} and we set $\bar{R}:=R/\|\wh\|_2$. Then, using \eqref{eq:key_margin} it follows immediately that
\begin{align}
\Lc(\wh_R)&=\sum_{i=1}^n  \omega_{i}\log\left(1+ e^{\iota_i}e^{-\Delta_{i}y_i\wh_R^T\x_i}\right) \nn\\
&\leq \sum_{i=1}^n  \omega_{i}\log\left(1+ e^{\iota_i}e^{-\frac{\bar{R}}{\Delta_i}\Delta_{i}}\right)\nn\\
&=\sum_{i=1}^n  \omega_{i}\log\left(1+ e^{\iota_i}e^{-\bar{R}}\right) \nn\\
&\leq \omega_{\max}n e^{\iota_{\max}-\bar{R}}.\label{eq:ub}
\end{align}
In the first inequality above we used \eqref{eq:key_margin} and non-negativity of $\omega_i, \Delta_i \geq 0$. In the last line, we have called $\omega_{\max} := \max_{i\in[n]}\omega_{i}>0$ and $\iota_{\max} := \max_{i\in[n]}\iota_{i}>0$\,.

Next, we lower bound $\Lc(\w_R)$. To do this, consider the vector $$\bar{\w}=\frac{\tn{\wh}}{R}\w_R=\w_R/\bar{R}.$$ By feasibility of $\w_R$ (i.e. $\|\w_R\|_2\leq R$), note that $\|\bar\w\|_2 \leq \|\wh\|_2$. Also, from \eqref{eq:corr_gd}, {we know that $\bar\w\neq\wh$}. {Indeed, if it were $\bar\w=\wh \iff \wh/\|\wh\|_2 = \w_R/R$, then
$$
\frac{\wh^T\w_R}{R\|\wh\|_2} =  1,
$$
which would contradict \eqref{eq:corr_gd}. Thus, it must be that $\bar\w\neq\wh$.
From these and strong convexity of the objective function in \eqref{eq:CS-SVM_app}, it follows that $\bar\w$ must be \emph{infeasible} for \eqref{eq:CS-SVM}. Thus, there exists at least one example $\x_j,~j\in[n]$ and $\eps>0$ such that
\[
 y_j\bar{\w}^T\x_j \leq(1-\eps)(1/\Delta_j).
\]
}
But then
\begin{align}
 y_j\w_R^T\x_j \leq \bar{R}(1-\eps)(1/\Del_j),\label{eq:point2}
\end{align}
which we can use to lower bound $\Lc(\w_R)$ as follows:
\begin{align}
\Lc(\w_R)&\geq \omega_{j} \log\left(1+ e^{\iota_{j}-\Delta_{j}y_j\w_R^T\x_j}\right)\nn\\
&\geq \omega_{j} \log\left(1+ e^{\iota_{y_j}-\bar{R}\Delta_{j}\frac{(1-\eps)}{\Delta_j}}\right)\nn\\
&\geq \omega_{\min}\log\left(1+ e^{\iota_{\min}-\bar{R}(1-\eps)}\right).\label{eq:lb}
\end{align}
The second inequality follows fron \eqref{eq:point2} and non-negativity of $\Delta_{\pm},\omega_{\pm}.$ 

To finish the proof we compare \eqref{eq:lb} against \eqref{eq:ub}.   If $\eps\geq 1$, clearly $\Lc(\wh_R)<\Lc(\w_R)$ for sufficiently large $R$. Otherwise $e^{-\bar{R}(1-\eps)}\rightarrow 0$ with $R\rightarrow\infty$. Hence,
\[
\Lc(\w_R)\geq \omega_{\min}\log\left(1+ e^{\iota_{\min}-\bar{R}(1-\eps)}\right)\geq 0.5 \omega_{\min}e^{\iota_{\min}-\bar{R}(1-\eps)}.
\]
Thus, again
\[
\Lc(\wh_R)< \Lc(\w_R)\impliedby \omega_{\max}n e^{\iota_{\max}-\bar{R}}<0.5 \omega_{\min}e^{\iota_{\min}-\bar{R}(1-\eps)}\iff e^{\bar{R}\eps}>\frac{2n\omega_{\max}}{\omega_{\min}}e^{\iota_{\max}-\iota_{\min}},
\]
because the right side is true by picking  $R$ arbitrarily large.


\subsubsection{Proof of Lemma \ref{lem:CS_svm_easy_claim}}\label{sec:CS_svm_easy_claim}
The proof of Lemma \ref{lem:CS_svm_easy_claim} is standard, but included here for completeness. The lemma has two statements and we prove them in the order in which they appear.

\vp
\noindent\textbf{Linear separability $\implies$ feasibility of \eqref{eq:CS-SVM_app}.} Assume  $\w$ such that $y_i(\w^T\x_i)\geq 1$ for all $i\in[n]$, which exists by assumption. Define $M:=\max_{i\in[n]}\frac{1}{\Del_i}>0$ and consider $\wt=M\w$. Then, we claim that $\wt$ is feasible for \eqref{eq:CS-SVM_app}. To check this, note that
\begin{align*}
y_i = +1 &~\implies~ \x_i^T\wt=M(\x_i^T\w)\geq M\geq 1/\Delta_i\quad\text{since}~\x_i^T\w\geq 1,\\
y_i = -1 &~\implies~ \x_i^T\wt=M(\x_i^T\w)\leq -M\leq -1/\Delta_i\quad\text{since}~\x_i^T\w\leq -1.
\end{align*}
Thus, $y_i(\x_i^T\wt)\geq 1/\Del_i$ for all $i\in[n]$, as desired.

\vp
\noindent\textbf{Proof of \eqref{eq:CS_svm_easy_claim}.} For the sake of contradiction let $\wt\neq  \frac{\hat\w}{\|\hat\w\|_2}$ be the solution to the max-min optimization in the RHS of \eqref{eq:CS_svm_easy_claim}. Specifically, this means that $\|\wt\|_2=1$ and
$$
\tilde m:= \min_{i\in[n]} \Delta_iy_i\x_i^T\wt > \min_{i\in[n]} \Delta_iy_i\x_i^T \frac{\hat\w}{\|\hat\w\|_2}=: m.
$$
We will prove that the vector $\w^\prime:=\wt/\tilde m$ is feasible in \eqref{eq:CS-SVM_app} and has smaller $\ell_2$-norm than $\hat\w$ contradicting the optimality of the latter. First, we check feasibility. Note that, by definition of $\tilde m$, for any $i\in[n]$:
$$
\Delta_iy_i\x_i^T\w^\prime = \frac{\Delta_iy_i\x_i^T\wt}{\tilde m} \geq 1,
$$
Second, we show that $\|\w^\prime\|_2< \|\hat \w\|_2$:
$$
\|\w^\prime\|_2 = \frac{\|\wt\|_2}{\tilde m} = \frac{1}{\tilde m} < \frac{1}{m} = \frac{\|\hat\w\|_2}{\min_{i\in[n]} \Delta_iy_i\x_i^T\hat\w} \leq \|\hat\w\|_2,
$$
where the last inequality follows by feasibility of $\hat\w$ in \eqref{eq:CS-SVM_app}. This completes the proof of the lemma.

\subsection{Multiclass extension}
In this section, we present a natural extension of Theorem \ref{thm:implicit_loss_gen} to the multiclass VS-loss in \eqref{eq:loss_ours}. Here, let we let the label set $\Yc=\{1,2,\ldots,C\}$ for a $C$-class classification setting and consider the cross-entropy VS-loss:
\begin{align}\label{eq:loss_app_multi}
\Lc(\W):=\sum_{i\in[n]}\ell(y_i,\w_1^T\x_i,\ldots,\w_K^T\x_i) = \sum_{i\in[n]}\om_{y_i}\log\Big(1+\sum_{\substack{y'\in[C]\\y'\neq y_i}}e^{\iota_{y'}-\iota_{y_i}}e^{- (\Del_{y_i}\w_{y_i}^T\x_i-\Del_{y'}\w_{y'}^T\x_i)}\Big),
\end{align}
where $\W=[\w_1,\ldots,\w_C]\in\R^{C\times d}$ and $\w_y$ is the classifier corresponding to class $y\in[C]$. We will also consider the following multiclass version of the CS-SVM in \eqref{eq:CS-SVM_app}:
\begin{align}\label{eq:CS-SVM_app_multi}
\hat\W = \arg\min_{\W} \|\W\|_F\quad\text{subject to }\x_i^T(\Delta_{y_i}\w_{y_i}-\Delta_{y'}\w_{y'})\geq 1, ~\forall y'\neq y_i\in[C]~\text{ and }~\forall i\in[n].
\end{align}


Similar to Lemma \ref{lem:CS_svm_easy_claim},  it can be easily checked that \eqref{eq:CS-SVM_app_multi} is feasible provided that the training data are separable, in the sense that 
\begin{align}\label{eq:sep_multi}
\exists \W=[\w_1,\ldots,\w_K] \text{ suc that } \x_i^T(\w_{y_i}-\w_{y'})\geq1, \forall y'\in[C], y'\neq y_i~\text{ and }~\forall i\in[n].
\end{align}
Moreover, it holds that
$$
\hat\W/\|\hat\W\|_F = \arg\max_{\|\W\|_F=1}\min_{i\in[n]}\min_{y'\neq y_i} \,\x_i^T(\Delta_{y_i}\w_{y_i}-\Delta_{y'}\w_{y'}).
$$

The theorem below is an extension of Theorem \ref{thm:implicit_loss_gen} to multiclass classification. 

{
\begin{theorem}[Margin properties of VS-loss: Multiclass]\label{thm:implicit_loss_multi}
Consider a $C$-class classification problem and define the norm-constrained optimal classifier
\begin{align}\label{eq:con_ERM_multi}
\W_R=\arg\min_{\|{\W}\|_F\leq R}\Lc(\W),
\end{align}
with the loss $\Lc$ as defined in \eqref{eq:loss_app_multi} for positive (but otherwise arbitrary) parameters $\Delta_y, \omega_y>0, y\in[C]$ and arbitrary $\iota_{y}, y\in[C]$. 
Assume that the training dataset is linearly separable as in \eqref{eq:sep_multi}
and let $\hat\W$ be the solution of \eqref{eq:CS-SVM_app_multi}.
Then, it holds that 
\begin{align}\label{eq:implicit_2_prove_multi}
\lim_{R\rightarrow\infty}\frac{\W_R}{\|\W_R\|_F} = \frac{\hat\W}{\|\hat\W\|_2} \,.
\end{align}
\end{theorem}
}

\begin{proof}
The proof follows the same steps as in the proof of Theorem \ref{thm:implicit_loss_gen}. Thus, we skip some details and outline only the basic calculations needed.

It is convenient to introduce the following notation, for $\ell\in[C]$:
$$
p(\ell | \x, y, \W):= \frac{e^{\iota_y}e^{\Delta_{y} \x^T\w_\ell}}{\sum_{y'\in[C]}e^{\iota_{y'}}e^{\Delta_{y'} \x^T\w_{y'}}}.
$$
In this notation,
$\Lc(\W)=-\sum_{i\in[n]}\log\big(p(y_i|\x_i,y_i,\W)\big)$ and 
 for all  $\ell\in[C]$ it holds that
$$
\nabla_{\w_\ell}\Lc(\W) = \sum_{i\in[n]} \omega_{y_i}\Delta_{y_i} \left(p(\ell | \x_i, y_i, \W) - \ind{y_i=\ell} \right) \x_i.
$$
Thus, for any $\wte\W$ that is feasible in \eqref{eq:CS-SVM_app_multi} 
\begin{align*}
\sum_{\ell\in[C]}\wte\w_\ell^T\nabla_{\w_\ell}\Lc(\W) &= \sum_{i\in[n]}\sum_{\ell\in[C]}\omega_{y_i}\Delta_{y_i} \left(p(\ell | \x_i, y_i, \W) - \ind{y_i=\ell} \right) \x_i^T\wte\w_\ell\\
&= \sum_{i\in[n]}\sum_{\ell\neq y_i}\omega_{y_i}\Delta_{y_i} p(\ell | \x_i, y_i, \W) \,\x_i^T\wte\w_\ell - \omega_{y_i}\Delta_{y_i}\left(1 - p(y_i | \x_i, y_i, \W)  \right) \x_i^T\wte\w_{y_i}\\
&= \sum_{i\in[n]}\underbrace{-\omega_{y_i} \big(\sum_{\ell\neq y_i}p(\ell | \x_i, y_i, \W)\big)}_{<0}\,\underbrace{\Delta_{y_i}\x_i^T\left(\wt_\ell-\wt_{y_i}\right)}_{>0}~<~0,
\end{align*}
where in the third line we used that $\sum_{\ell\in[C]}p(\ell | \x, y, \W)=1$. With the above it can be shown following the exact same argument as in the proof of \eqref{eq:boundary} for the binary case that $\|\W_R\|_F=R$, the minimizer of \eqref{eq:con_ERM_multi} satisfies the constraint with equality.

The proof continues with a contradiction argument similar to the binary case. Assume the desired \eqref{eq:implicit_2_prove_multi} does not hold. We will then show that for $\hat\W_R=\frac{R}{\|\hat\W\|_F}\hat\W$ and sufficiently large $R>0$:
$\Lc(\hat\W_R)<\Lc(\W_R)$. 

Using feasibility of $\hat\W$ in \eqref{eq:CS-SVM_app_multi} and defining $\omega_{\max}:=\max_{y\in[C]}\omega_y$ and $\iota_{\max}=\max_{y\neq y'\in[C]}\iota_{y'}-\iota_y$, it can be shown similar to \eqref{eq:ub}  that
\begin{align}\label{eq:multi_proof_1}
\Lc(\hat\W_R)&=\sum_{i\in[n]}\om_{y_i}\log\Big(1+\sum_{\substack{y'\in[C]\\y'\neq y_i}}e^{\iota_{y'}-\iota_{y_i}}e^{-(R/\|\W\|_F) (\Del_{y_i}\wh_{y_i}^T\x_i-\Delta_{y'}\wh_{y'}^T\x_i)}\Big),\nn
\\
& \leq n \omega_{\max} \log\Big(1+(K-1)e^{\iota_{\max}}e^{-R/\|\hat\W\|_F}\Big)\leq n(K-1) e^{\iota_{\max}}e^{-R/\|\hat\W\|_F}.
\end{align}
Next, by contradiction assumption and strong convexity of \eqref{eq:CS-SVM_app_multi}, for $\bar\W=\frac{\|\hat\W\|_2}{R}\W_R$, there exist $\eps>0$ and at least one $j\in[n]$ and $y'\neq y_j$ such that $\x_j^T(\Delta_{y_j}\bar\w_{j} -\Delta_{y'} \bar\w_{y'})\leq (1-\eps)$. With this, we can show similar to \eqref{eq:lb} that
\begin{align}\label{eq:multi_proof_2}
\Lc(\W_R) \geq \log\Big(1+e^{\iota_{y'}-\iota_{y_j}}e^{R/\|\hat\W\|_F(1-\eps)}\Big).
\end{align}
The proof is complete by showing that for sufficiently large $R$ the RHS of \eqref{eq:multi_proof_2} is larger than the LHS of \eqref{eq:multi_proof_1} leading to a contradiction. We omit the details for brevity.\end{proof}

%% file: gradient_flow.tex
\subsection{Implicit bias of Gradient flow with respect to VS-loss}\label{sec:gradient_flow}
Theorem \ref{thm:implicit_loss_gen} does not consider the effect of the optimization algorithm. Instead here, we study gradient flow (the limit of gradient descent for infinitesimal step-size) and characterize its implicit bias when applied to the VS-loss. Similar, to Theorem \ref{thm:implicit_loss_gen}, we find that the iterations of gradient flow converge to the solution of a corresponding CS-SVM. For simplicity, we consider a VS-type adjusted exponential loss $\ell(t)=e^{-t}$, rather than logistic loss $\ell(t)=\log(1+e^{-t})$. Recent work makes it clear that both loss functions have similar implicit biases and similar lines of arguments are used to analyze the convergence properties \cite{ji2021characterizing,ji2018risk}. Thus, one would expect that insights also apply to logistic loss.
\begin{theorem}[Implicit bias of the gradient flow]\label{thm:gradient_flow}
Consider the gradient flow iteration
$
\dot{\w}_t=-\nabla\Lc(\w_t),
$
on the exponential VS-loss $\Lc(\w)=\sum_{i\in[n]}\omega_i\exp(-\Delta_iy_i\x_i^T\w+\iota_i).$ Recall that $\hat\w$ is the solution to the CS-SVM in \eqref{eq:CS-SVM_app}.
For almost every dataset which is linearly separable and any starting point $\w_0$ the gradient flow iterates will behave as $\w(t)=\hat\w\log(t)+\rhob_t$ with a bounded residual $\rhob_t$ so that $\lim_{t\rightarrow\infty}\frac{\w_t}{\|\w_t\|_2}=\frac{\hat\w}{\|\hat\w\|_2}.$
\end{theorem}

Note that \cite{soudry2018implicit} previously studied the implicit bias of the gradient flow on standard CE or exponential loss. The theorem above studies the gradient flow applied to the VS-loss and its proof is similar to \cite{soudry2018implicit}. 

\begin{proof} Let $\Sc\subset[n]$ be the set of indices such that $\forall i\in\Sc: \Delta_iy_i\x_i^T\hat\w=1$, i.e. the set of support vectors of the CS-SVM.  By KKT conditions (eg. see Equation \eqref{eq:KKT_2}), there exist $\eps_i>0$ such that $\hat\w=\sum_{i\in\Sc}\eps_iy_i\x_i.$ Moreover, by \cite[Lemma 12]{soudry2018implicit}, for almost all datasets it is true that $|\Sc|\leq d$ and $i\in\Sc\implies\eps_i>0$. Thus, for almost all datasets we can define vector $\wt$ satisfying the following equation
$\omega_i\Delta_i\exp(-\Delta_iy_i\x_i^T\wt+\iota_i) = \eps_i, \forall i\in[S].$
Note then that
\begin{align}\label{eq:kkt_GF}
\wh=\sum_{i\in[S]}\omega_i\Delta_ie^{-\Delta_iy_i\x_i^T\wt+\iota_i}y_i\x_i
\end{align}

Let us define
$\rb_t=\rhob_t-\wt=\w_t-\log(t)\hat\w-\wt.$ It suffices to show that $\|\rb(t)\|_2$ is bounded, since that would automatically give $\rhob_t$ is bounded. By the gradient flow equation, we have that $$
\dot\rb_t=-\nabla\Lc(\w_t)-\frac{\hat\w}{t}=\sum_{i\in[n]}\omega_i\Delta_iy_ie^{-\Delta_iy_i\x_i^T\w_t+\iota_i}\x_i-\frac{\hat\w}{t}.
$$ Therefore,
\begin{align}
\frac{1}{2}\frac{\mathrm d}{\mathrm{d} t}\|\rb_t\|_2^2&=\dot\rb_t^T\rb_t
=\sum_{i\in[n]}\omega_i\Delta_iy_ie^{-\Delta_iy_i\x_i^T\w_t+\iota_i}\x_i^T\rb_t-\frac{1}{t}\hat\w^T\rb_t\nn\\
&=\underbrace{\sum_{i\in\Sc}\omega_i\Delta_iy_ie^{-\Delta_iy_i\x_i^T\w_t+\iota_i}\x_i^T\rb_t-\frac{1}{t}\hat\w^T\rb_t}_{:=A} + \underbrace{\sum_{i\not\in\Sc}\omega_i\Delta_iy_ie^{-\Delta_iy_i\x_i^T\w_t+\iota_i}\x_i^T\rb_t}_{:=B}\label{eq:implicit_step1}
\end{align}
We now study the two terms $A$ and $B$ separately. In doing so, recall that $\w_t=\rb_t+\log(t)\hat\w+\wt.$ Hence, using the fact that ${\Delta_iy_i\x_i^T\wh}={\begin{cases}=1 &i\in\Sc \\ \geq m>1 &i\not\in\Sc\end{cases}}$, it holds that
\begin{align*}
\exp\big({-\Delta_iy_i\x_i^T\w_t+\iota_i}\big)
&\begin{cases}
=\frac{1}{t}\cdot\exp\big({-\Delta_iy_i\x_i^T\rb_t}\big)\cdot\exp\big({-\Delta_iy_i\x_i^T\wt}+\iota_i\big) &i\in\Sc \\
\leq\frac{1}{t^m}\cdot\exp\big({-\Delta_iy_i\x_i^T\rb_t}\big)\cdot\exp\big({-\Delta_iy_i\x_i^T\wt}+\iota_i\big) &i\not\in\Sc
\end{cases}
\end{align*}
Using this and \eqref{eq:kkt_GF}, the term $A$ becomes
\begin{align*}
A&=\frac{1}{t}\sum_{i\in[S]}\omega_ie^{-\Delta_iy_i\x_i^T\wt+\iota_i}\cdot e^{-\Delta_iy_i\x_i^T\rb_t} \Delta_iy_i\x_i^T\rb_t-\frac{1}{t}\sum_{i\in[S]}\omega_i\Delta_ie^{-\Delta_iy_i\x_i^T\wt+\iota_i}y_i\x_i^T\rb_t\\
&=\frac{1}{t}\sum_{i\in[S]}\omega_ie^{-\Delta_iy_i\x_i^T\wt+\iota_i}\cdot\Big(e^{-\Delta_iy_i\x_i^T\rb_t} \Delta_iy_i\x_i^T\rb_t-\Delta_iy_i\x_i^T\rb_t\Big)\leq 0,
\end{align*}
since $\forall x, x\geq xe^{-x}.$ 

Similarly, for term $B$:
\begin{align}
B\leq\frac{1}{t^m}\sum_{i\not\in\Sc}\omega_i e^{-\Delta_iy_i\x_i^T\wt+\iota_i}\cdot e^{-\Delta_iy_i\x_i^T\rb_t}\cdot \Delta_iy_i\x_i^T\rb_t \leq
\frac{1}{t^m}\sum_{i\not\in\Sc}\omega_i e^{-\Delta_iy_i\x_i^T\wt+\iota_i}\,,\label{eq:m>1}
\end{align}
since $\forall x, xe^{-x}\leq 1$.

To finish the proof it only takes now using the above bounds on $A, B$ and integrating both sides of Equation \eqref{eq:implicit_step1}. This gives that for all $t_0, t>t_0$, there exists finite constant $C$ such that $\|\rb_t\|^2\leq \|\rb_{t_0}\|^2+C$ where it was critical that $m>1$ in \eqref{eq:m>1} for the corresponding integral to be finite. This proves that $\|\rb_t\|_2$ is bounded as desired. 
\end{proof}

We note that the above proof is a straightforward extension of \cite{soudry2018implicit} for analysis of CDT, with a simple rescaling of the features of the training set according to the labels, however the analysis for VS-loss with additive logit-adjustments (although similar) cannot be obtained as a special case of \cite{soudry2018implicit}.

\subsection{Numerical illustrations of Theorems \ref{propo:gd} and \ref{thm:gradient_flow}}\label{sec:gd_num_app}

\begin{figure}[t]
\begin{center}
\begin{subfigure}[b]{0.45\textwidth}
         \centering
         \includegraphics[width=\textwidth]{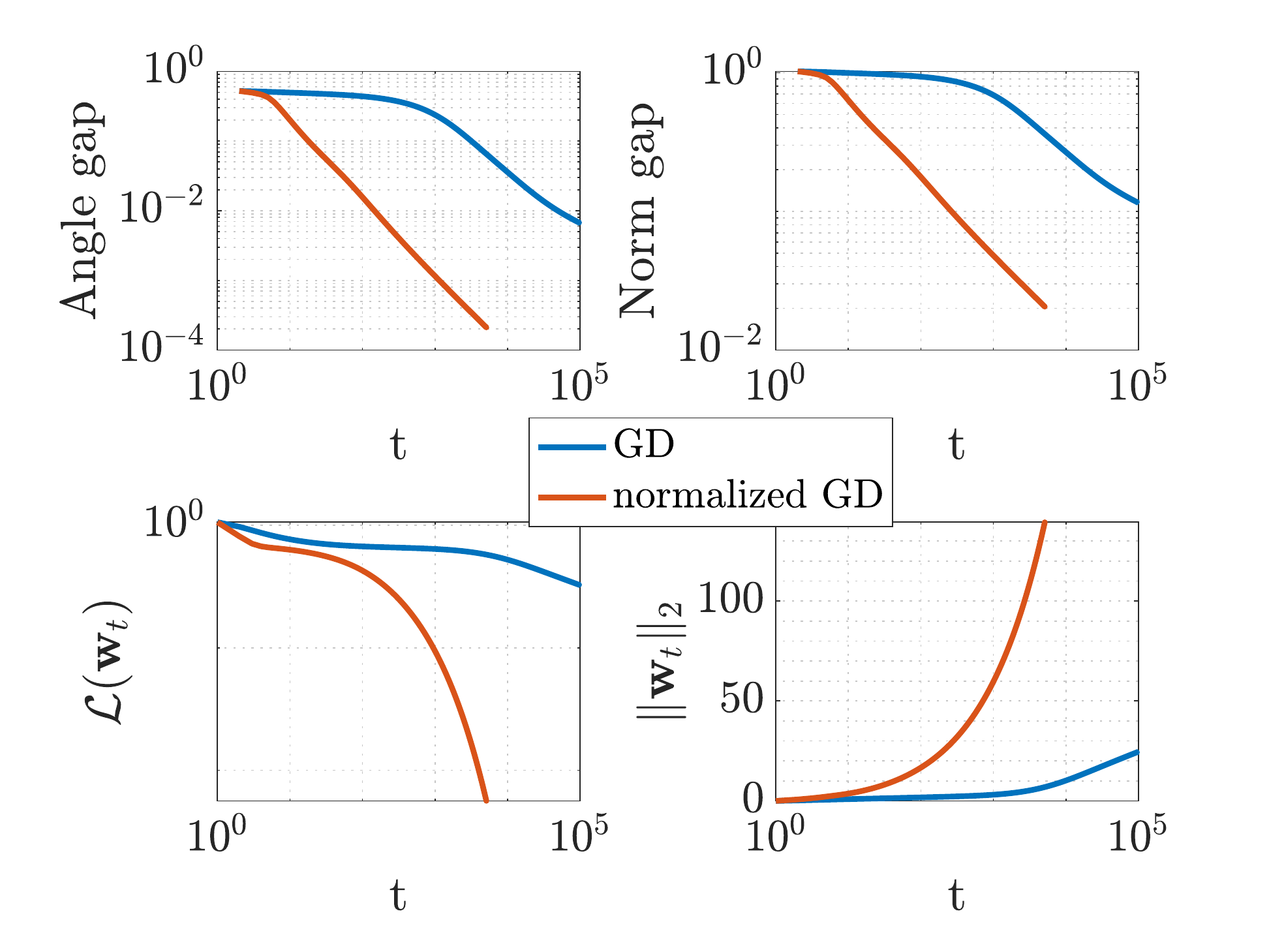}
                  \caption{CDT-loss vs CS-SVM}
\end{subfigure}
\begin{subfigure}[b]{0.45\textwidth}
         \centering
         \includegraphics[width=\textwidth]{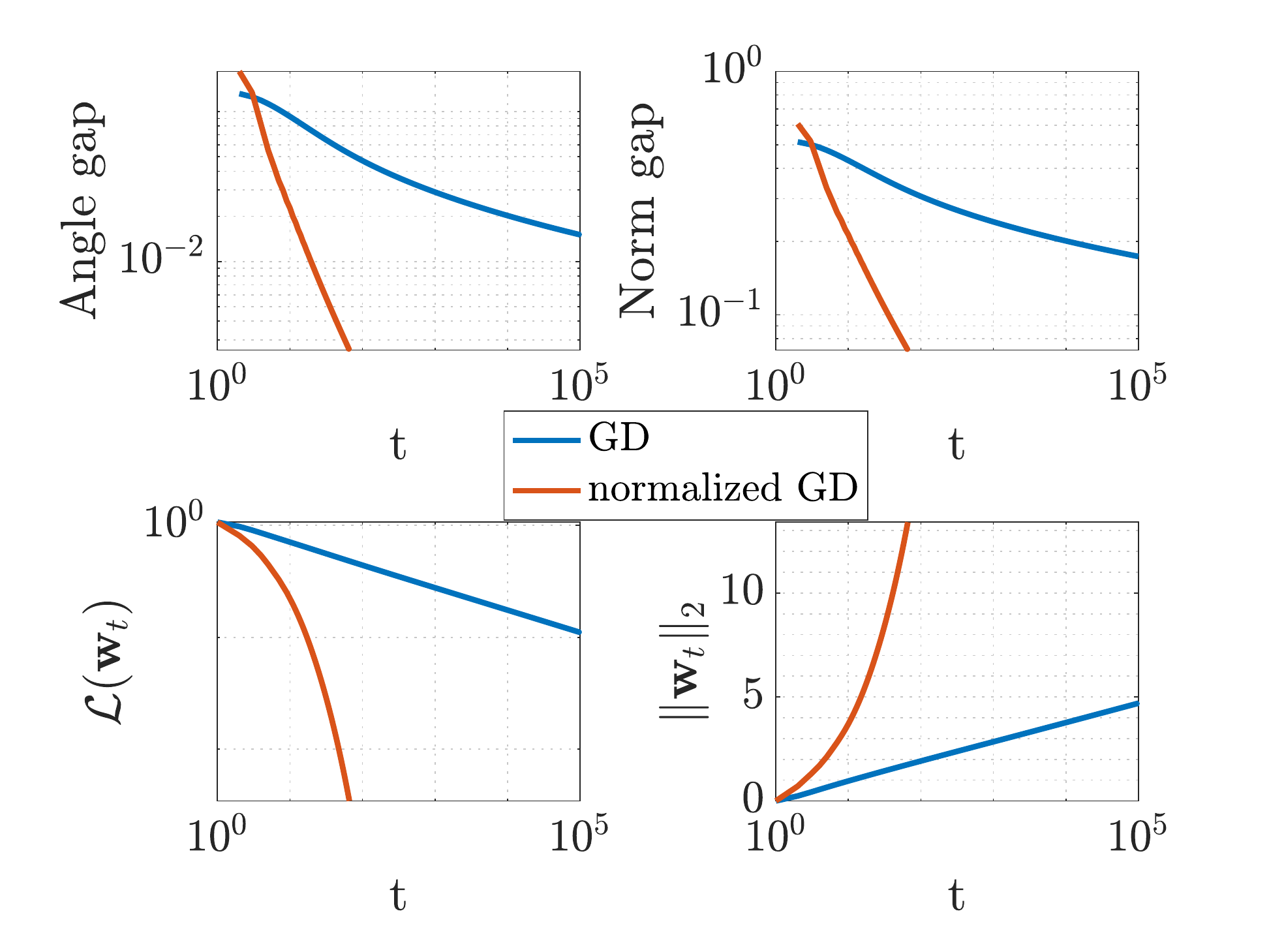}
         \caption{LA-loss vs SVM}
\end{subfigure}
\end{center}
\caption{
Convergence properties of GD (blue) and normalized GD (red) iterates $\w_t, t\geq 1$ on VS-loss with $f_\w(x)=\w^T\x$ for two set of parameter choices: (a) $\omega_y=1,\iota_y=0,\Delta_y=\delta\ind{y=1}+\ind{y=-1}$ (aka CDT-loss) with $\delta=20$; (b) $\omega_y=1,\iota_y=\pi^{-1/4}\ind{y=1}+(1-\pi)^{-1/4}\ind{y=-1},\Delta_y=1$ (aka LA-loss).  
 We plotted the angle gap $1-\frac{\wh^T\w_t}{\|\w_t\|_2\|\wh\|_2}$  
and norm gap $\|\frac{\w_t}{\|\w_t\|_2}-\frac{\wh}{\|\wh\|_2}\|_2$ 
of $\w_t$  to $\wh$, for two values of $\wh$ for the two subfigures as follows: (a) $\wh$ is the CS-SVM solution in \eqref{eq:CS-SVM} with parameter $\delta$; (b) $\wh$ is the standard SVM solution. Data were generated from a Gaussian mixture model with $\mub_1=2\eb_1, \mub_2=-3\eb_1 \in\R^{220}$, $n=100$ and $\pi=0.1$. For (standard) GD we used a constant rate $\eta_t=0.1$. For normalized GD, we used $\eta_t=\frac{1}{\sqrt{t}\|\nabla\Lc(\w_t)\|_2}$ as suggested in \cite{nacson2019convergence}.}
\label{fig:normalized_gd_experiments}
\end{figure}


Figure \ref{fig:normalized_gd_experiments} numerically demonstrate the validity of Theorems \ref{propo:gd} and \ref{thm:gradient_flow}. Here, we solved the VS-loss  in Equation \eqref{eq:loss_ours_bin} using gradient descent (GD) for GMM data with class imbalance $\pi=0.1$. We ran two experiments for two choices of parameters in \eqref{eq:loss_ours_bin} corresponding to CDT-loss (with non-trivial multiplicative weights) and the LA-loss (with non-trivial additive weights); see the figure's caption for details. For each iterate outcome $\w_t$ of GD, we report the (i) angle and (ii) vector-norm gap to CS-SVM and SVM for the VS-loss and LA-loss, respectively, as well as, the (iii) value of the loss $\Lc(\w_t)$ and the (iv) norm of the weights $\|\w_t\|_2$ at current iteration. Observe that the loss $\Lc(\w_t)$ is driven to zero and the norm of the weights $\|\w_t\|_2$ increases to infinity with increasing $t$. 

The experiment confirms that the VS-loss converges (aka angle/norm gap vanishes) to the CS-SVM solution, while the LA-loss converges to the SVM. 

In Figure \ref{fig:normalized_gd_experiments}, we also study (curves in red) the convergence properties of \emph{normalized GD}. Following \cite{nacson2019convergence}, we implemented a version of normalized GD that uses a variable learning rate $\eta_t$ at iteration $t$  normalized by the gradient of the loss as follows: $\eta_t=\frac{1}{\|\nabla\Lc(\wt)\|_2\sqrt{t+1}}$. \cite{nacson2019convergence} (see also \cite{ji2021characterizing}) demonstrated that this normalization speeds up the convergence of standard logistic loss to SVM. Figure \ref{fig:normalized_gd_experiments} suggests that the same is true for convergence of the VS-loss to the CS-SVM.

%% file: optimal_tuning.tex
\section{Optimal tuning of CS-SVM}\label{sec:opt_tune_app}

\subsection{An explicit formula for optimal tuning}
The parameter $\delta$ in the CS-SVM constraints in \eqref{eq:CS-SVM} aims to shift the decision space towards the majority class so that it better balances the conditional errors of the two classes. But, how to best choose $\delta$ to achieve that? That is, how to find $\arg\min_\delta \Rc_+(\delta)+\Rc_-(\delta)$ where $\Rc_\pm(\delta):=\Rc_\pm\big((\hat\w_\delta,\hat b_\delta)\big)$? Thanks to Theorem \ref{thm:main_imbalance}, we can substitute this hard, data-dependent parameter optimization problem with an analytic form that only depends on the problem parameters $\pi, \gamma$ and $\M$. Specifically, we seek to solve the following optimization problem
\begin{align}\nn
&\arg\min_{\delta>0}~ Q(\eb_1^T\Vb\Sb\rhob_\delta+b_\delta/q_\delta) + Q(-\eb_2^T\Vb\Sb\rhob_\delta-b_\delta/q_\delta)\\
&~~\text{sub. to~~~ $(q_\delta,\rhob_\delta,b_\delta)$ defined as \eqref{eq:eta_eq_main}.}\label{eq:delta_tune_1}
\end{align}
Compared to the original data-dependent problem, the optimization above has the advantage that it is explicit in terms of the problem parameters. However, as written, the optimization is still cumbersome as even a grid search over possible values of $\delta$ requires solving the non-linear equation \eqref{eq:eta_eq_main} for each candidate value of $\delta$. Instead, we can exploit a structural property of CS-SVM (see Lemma \ref{lem:1_to_delta} in Section \ref{sec:lemma1_proof}) to rewrite \eqref{eq:delta_tune_1} in a more convenient form. Specifically, we will show in Section \ref{sec:delta_star_proof} that \eqref{eq:delta_tune_1} is equivalent to the following \emph{explicit minimization}:
\begin{align}\label{eq:delta_tune_2}
\hspace{-0.08in}\arg\min_{\delta>0}~ Q\Big(\ell_{+} +\big(\frac{\delta-1}{\delta+1}\big){q_1^{-1}}\Big) + Q\Big(\ell_{-} -\big(\frac{\delta-1}{\delta+1}\big){q_1^{-1}}\Big),
\end{align}
where we defined $\ell_{+}:=\eb_1^T\Vb\Sb\rhob_1+b_1/q_1,~\ell_-:=-\eb_2^T\Vb\Sb\rhob_1-b_1/q_1$, and, $(q_1,\rhob_1,b_1)$ are as defined in Theorem \ref{thm:main_imbalance} for $\delta=1$. In other words, $(q_1,\rhob_1,b_1)$ are the parameters related to the standard hard-margin SVM, for which the balanced error is then given by $\left(Q(\ell_+)+Q(\ell_-)\right)\big/2$. To summarize, we have shown that  one can optimally tune $\delta$ to minimize the \emph{asymptotic} balanced error by minimizing the objective in \eqref{eq:delta_tune_2} that only depends on the parameters $(q_1,\rhob_1,b_1)$ characterizing the asymptotic performance of SVM. In fact, we obtain explicit formulas for the optimal value $\delta_\star$ in \eqref{eq:delta_tune_2} as follows
\begin{align}\label{eq:delta_star}
\delta_\star := ({\ell_- -\ell_++2q_1^{-1}})\big/{\left(\ell_+-\ell_-+2q_1^{-1}\right)_+},
\end{align}
where it is understood that when the denominator is zero (i.e. $\ell_+-\ell_-+2q_1^{-1}\leq0$) then $\delta_\star\rightarrow\infty$.  When $\ell_+-\ell_-+2q_1^{-1}>0$,
setting $\delta=\delta_\star$ in \eqref{eq:CS-SVM} not only achieves minimum balanced error among all other choices of $\delta$, but also it achieves perfect balancing between the conditional errors of the two classes, i.e. $\Rc_+=\Rc_-=Q(\frac{\ell_-+\ell_+}{2}).$

Formally, we have the following result.

\begin{theorem}[Optimal tuning of CS-SVM]\label{propo:delta_star}
Fix $\gamma>\gamma_\star$. Let $\Rcbalinf(\delta)$ denote the asymptotic balanced error of the CS-SVM with  margin-ratio parameter $\delta>0$ as specified in Theorem \ref{thm:main_imbalance}. Further let $(q_1,\rhob_1,b_1)$ the solution to \eqref{eq:eta_eq_main} for $\delta=1$. Finally, define
$$
\ell_{+}:=\eb_1^T\Vb\Sb\rhob_1+b_1/q_1,\quad\ell_-:=-\eb_2^T\Vb\Sb\rhob_1-b_1/q_1,$$
 Then, for all $\delta> 0$ it holds that 
$$
\Rcbalinf(\delta) \geq \Rcbalinf(\delta_\star)
$$
where $\delta_\star$ is defined as 
\begin{align}\label{eq:delta_star_app}
\delta_\star = \begin{cases}
\frac{\ell_- -\ell_++2q_1^{-1}}{\ell_+-\ell_-+2q_1^{-1}} & \text{if } \ell_++\ell_-\geq0 \text{ and } \ell_+-\ell_-+2q_1^{-1}> 0,\\
\rightarrow\infty & \text{if } \ell_++\ell_-\geq0 \text{ and } \ell_+-\ell_-+2q_1^{-1}\leq 0, \\
\rightarrow 0 & \text{if } \ell_++\ell_-<0.
\end{cases}
\end{align}
Specifically, if $\ell_+ + \ell_- \geq 0$ and $\ell_+-\ell_-+2q_1^{-1}>0$ hold, then the following two hold: (i) 
$\Rcbalinf(\delta_\star)=Q\left(\left({\ell_-+\ell_+}\right)\big/{2}\right)$, and, (ii)  the asymptotic conditional errors are equal, i.e.
$
\Rc_+(\delta_\star)=\Rc_-(\delta_\star).
$ 
\end{theorem}

\begin{figure}[t]
     \centering
     \begin{subfigure}[b]{0.3\textwidth}
         \centering
         \includegraphics[width=\textwidth]{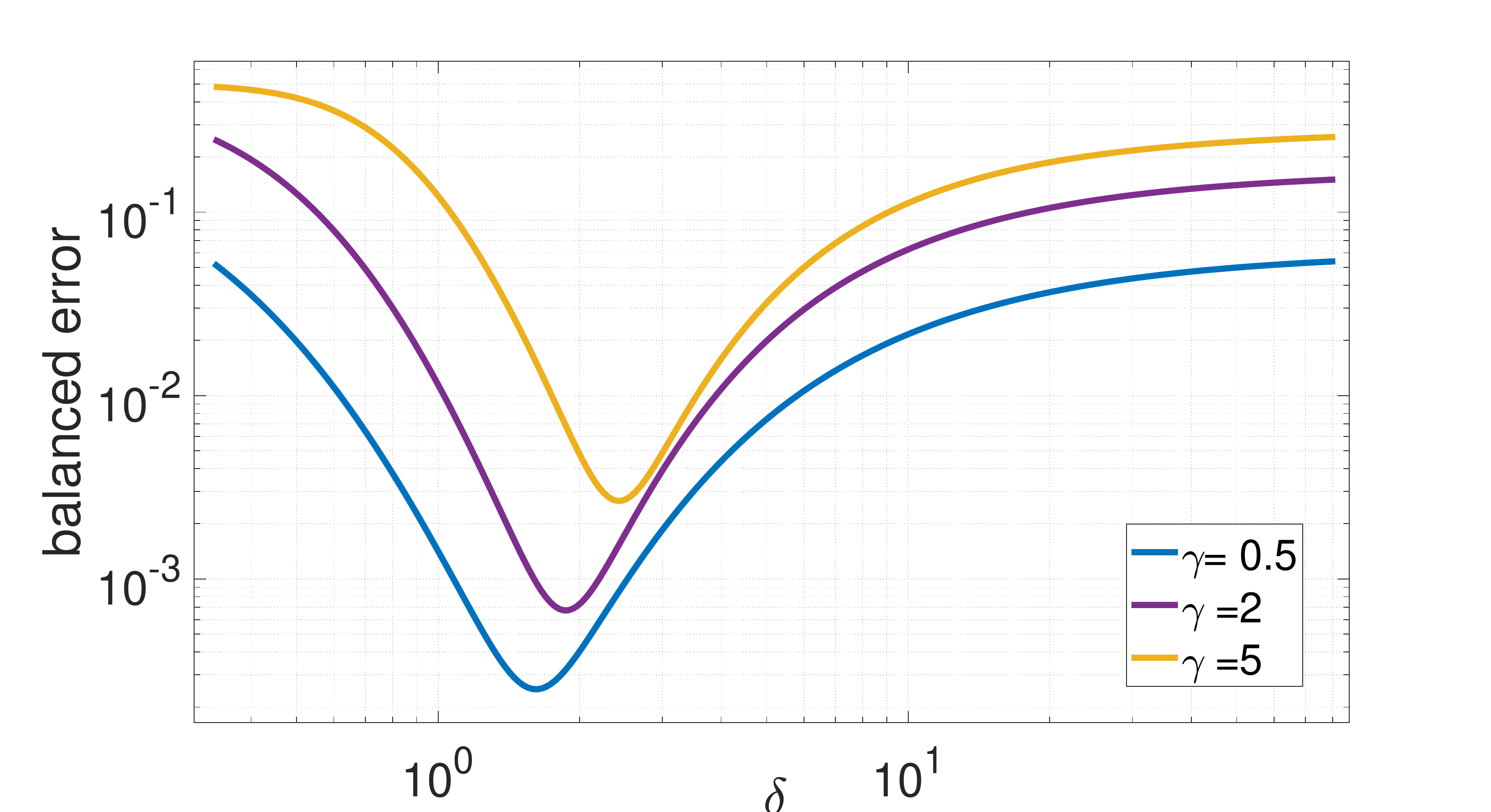}
         \caption{$\pi=0.1$}
         \label{}
     \end{subfigure}
     \hfill
     \begin{subfigure}[b]{0.3\textwidth}
         \centering
         \includegraphics[width=\textwidth]{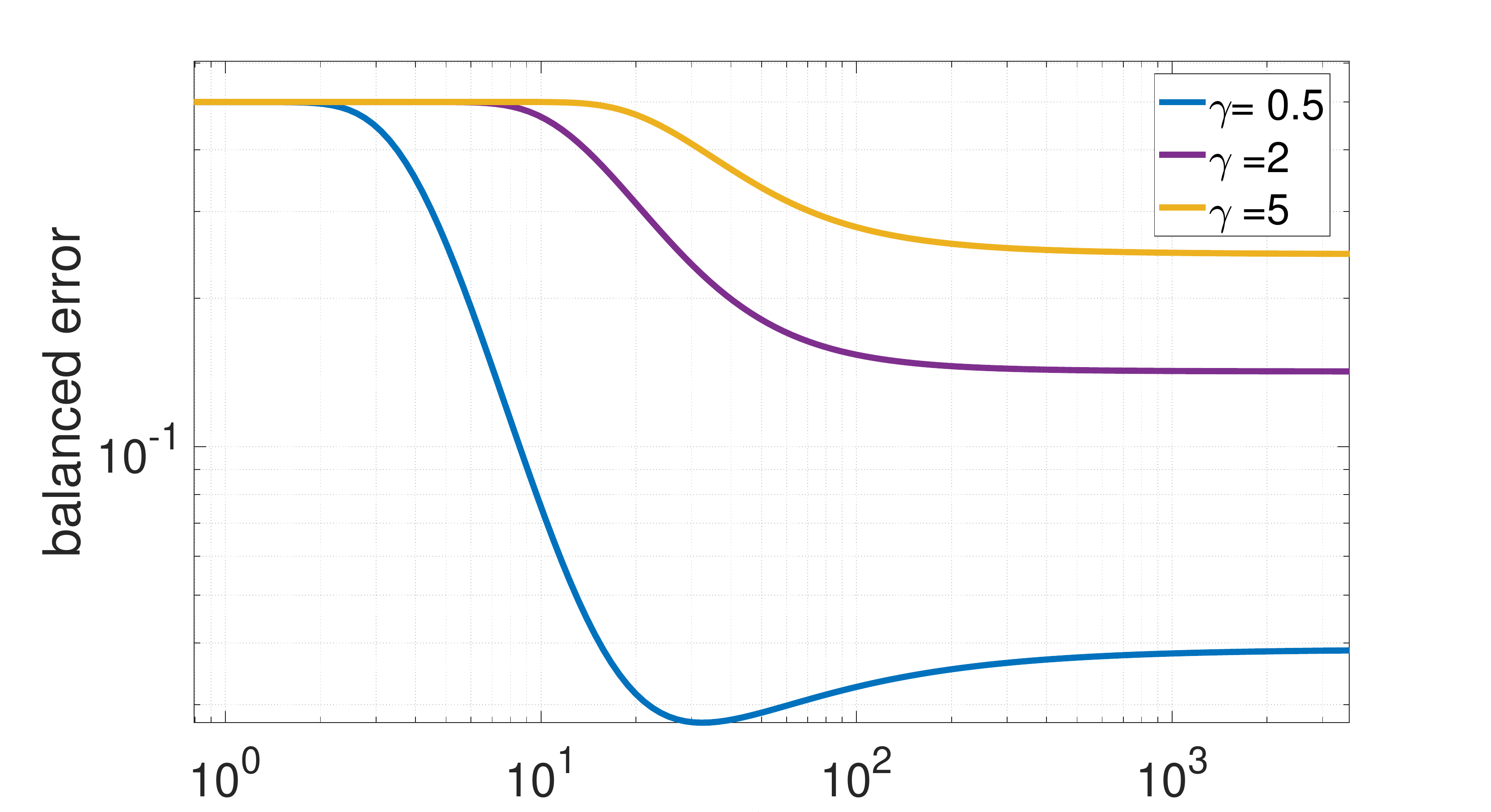}
         \caption{$\pi=0.0025$}
         \label{}
     \end{subfigure}
     \hfill
     \begin{subfigure}[b]{0.3\textwidth}
         \centering
         \includegraphics[width=\textwidth]{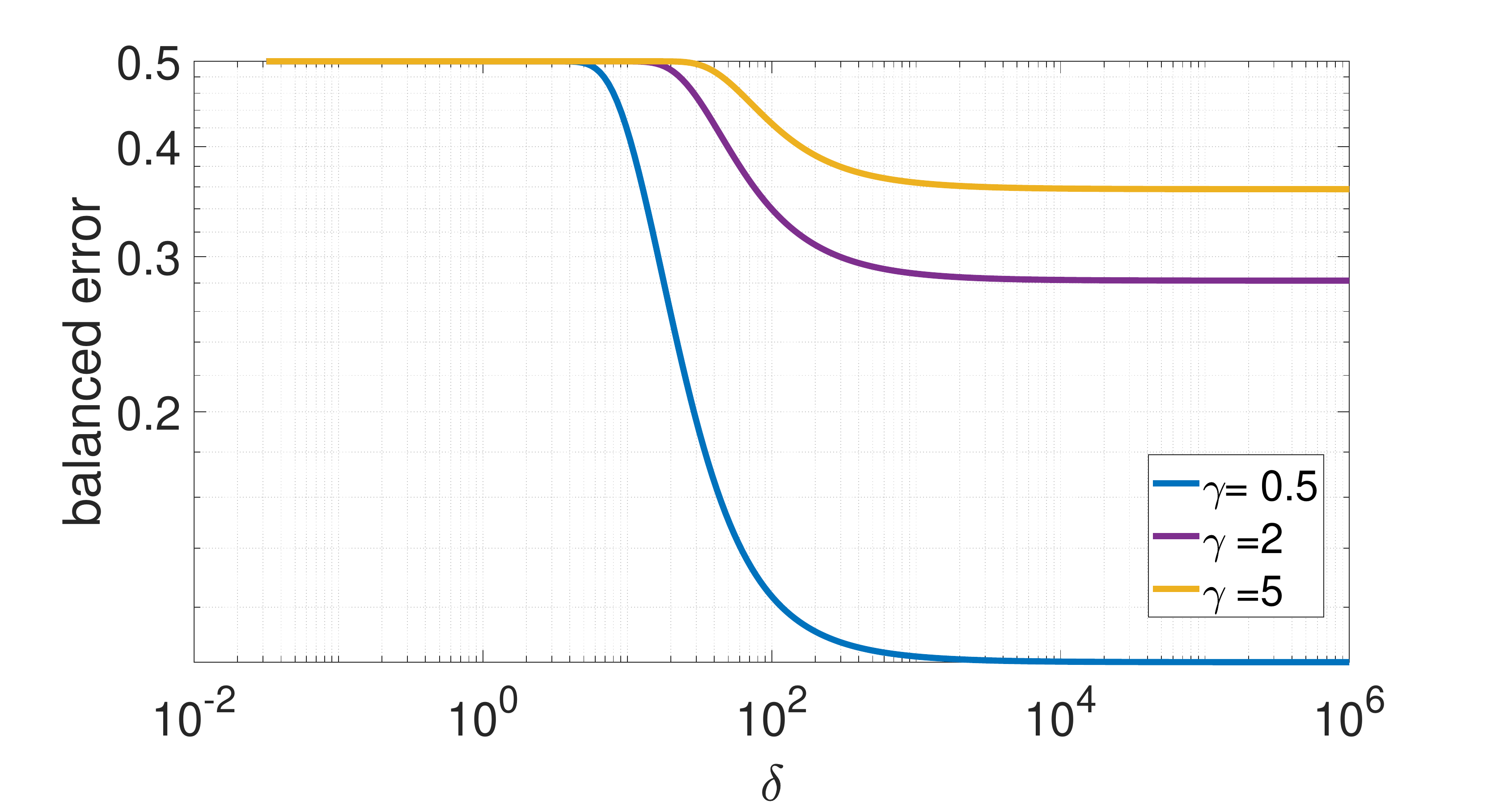}
         \caption{$\pi=0.0001$}
         \label{fig:five over x}
     \end{subfigure}
        \caption{Graphical illustration of the result of Theorem \ref{propo:delta_star}:  Balanced errors of CS-SVM against the margin-ratio parameter $\delta$ for a GMM of antipodal means with $\|\mu_+\|=\|\mu_-\|=4$ and different minority class probabilities $\pi$. The balanced error is computed using the formulae of Theorem \ref{thm:main_imbalance}. For each case, we studied three different values of $\gamma$. The value $\delta_\star$ at which the curves attain (or approach) their minimum are predicted by Theorem \ref{propo:delta_star}. Specifically, note the following for the three different priors. (a) For all values of $\gamma$, the minimum is attained (cf. first branch of \eqref{eq:delta_star_app}). (b) For $\gamma=2,5$ the minimum is approached in the limit $\delta\rightarrow\infty$ (cf. second branch of \eqref{eq:delta_star_app}), but it is attained for $\gamma=0.5$ (c) The minimum is always approached as $\delta_\star\rightarrow\infty$. }
        \label{fig:delta_star_theory_formula}
\end{figure}

\begin{figure}[t]
		\centering
		\includegraphics[width=0.3\textwidth]{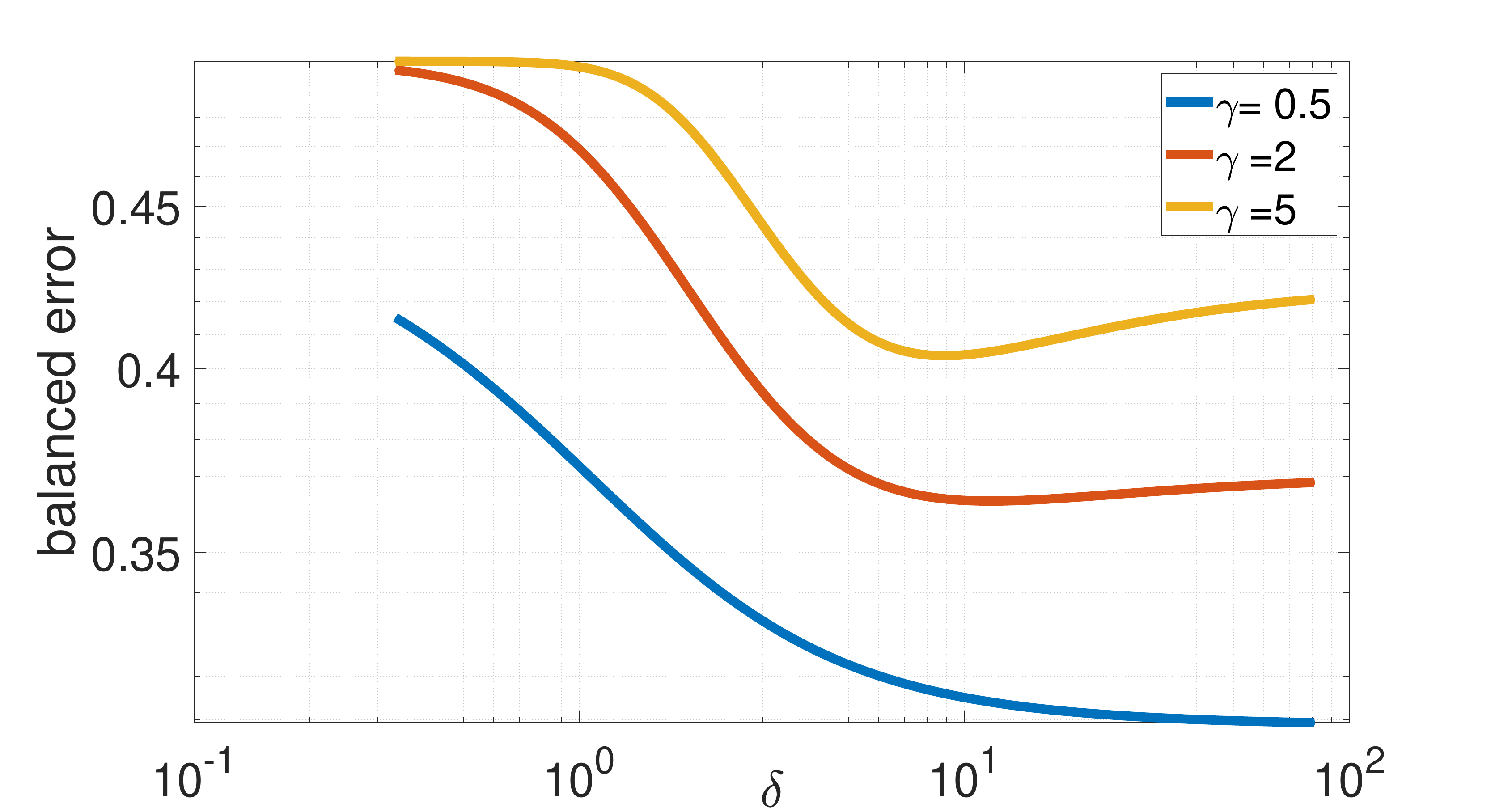}
		
	\caption{An example showing the dependence of $\delta_\star$ on the data geometry. The above figure is similar to Fig \ref{fig:delta_star_theory_formula} but with a smaller $\|\mu_+\|=\|\mu_-\|=1$, and for $\pi=0.1$. While in Fig \ref{fig:delta_star_theory_formula}, the value of $\delta_\star$, whenever finite, can be seen to increase with increase in $\gamma$, for the current setting, it is observed to decrease. Note also that $\delta_\star\rightarrow\infty$ for $\gamma=0.5$, but finite for $\gamma=2,5$.}
	\label{fig:delta_star_theory_formula_2}
\end{figure}

See Figures \ref{fig:five over x} and \ref{fig:delta_star_theory_formula_2} for numerical illustrations of the formula in Theorem \ref{propo:delta_star}, specifically how $\delta_\star$ depends on $\pi$ and $\gamma$. 


\subsubsection{Data-dependent heuristic to estimate $\delta_\star$}\label{sec:delta_star_estimate}


It is natural to ask if formula \eqref{eq:delta_star_app} can be used for tuning in practice. To answer this, observe that evaluating the formula requires knowledge of the true means, which are typically unknown. In this section, we propose a \emph{data-dependent heuristic} to estimate $\delta_\star$. More generally, tuning $\delta$ (or $\Delta_y$ in VS-loss) requires a train-validation split by creating a balanced validation set from the original training data which would help assess balanced risk. Since there is only a single hyperparameter we expect this approach to work well with fairly small validation data (without hurting the minority class sample size).

Recall from Equation \eqref{eq:delta_star} that  
$
\delta_\star := ({\ell_- -\ell_++2q_1^{-1}})\big/{\left(\ell_+-\ell_-+2q_1^{-1}\right)_+},
$
where $\ell_{+}:=\eb_1^T\Vb\Sb\rhob_1+b_1/q_1$ and $\ell_-:=-\eb_2^T\Vb\Sb\rhob_1-b_1/q_1$. Also, according to Theorem \ref{thm:main_imbalance} and for $\delta=1$ it holds that
\begin{align}\label{eq:remember_mnist}
(\|\hat\w_1\|_2, {\wh_1^T\mub_+}/{\|\wh_1\|_2}, {\wh_1^T\mub_-}/{\|\wh_1\|_2} , \hat b_1) \rP (q_1,\eb_1^T\Vb\Sb\rhob_1,\eb_2^T\Vb\Sb\rhob_1, b_1).
\end{align}
The first key observation here is that $\wh_1,\hat b_1$ are the solutions to SVM, thus they are data-dependent quantities to which we have access to. Hence, we can simply run SVM and estimate $q_1$ and $b_1$ using Equation \eqref{eq:remember_mnist}. Unfortunately, to further estimate $\rhob_1$ we need knowledge of the data means. When this is not available, we propose approximating the data means by a simple average of the features, essentially pretending that the data follow a GMM.

Concretely, our recipe for approximating the optimal $\delta$ is as follows. First, using the training set we calculate the empirical means for the two classes, $\tilde{\mub}_+$ and $\tilde{\mub}_-$. (Ideally, this can be done on a balanced validation set.) Then, we train standard SVM on the same set of data and keep track of the coefficients $\wh_1$ and the intercept $\hat b_1$. Then, we can reasonably approximate the optimal $\delta$ as:
\begin{align}
\tilde\delta_\star := \frac{\tilde{\ell_-} -\tilde{\ell}_++2\|\hat\w_1\|_2^{-1}}{\left(\tilde{\ell_+}-\tilde{\ell_-}+2\|\hat\w_1\|_2^{-1}\right)_+}, \text{ with }
\tilde{\ell_{+}}:=\frac{\wh_1^T\tilde{\mub}_+ + \hat b_1}{\|\wh_1\|_2}, \quad \tilde{\ell_{-}}:=-\frac{\wh_1^T\tilde{\mub}_- + \hat b_1}{\|\wh_1\|_2}.
\end{align}

We expect this data-dependent theory-driven heuristic to perform reasonably well on data that resemble the GMM. For example, this is confirmed by our experiments in Figures \ref{fig:mnist_delta_tuning} and \ref{fig:cls_imbalance_antipodal_misclass_errors}. More generally, we propose tuning $\delta$ with a train-validation split by creating a balanced validation set from the original training data which would help assess balanced risk. Since there is only a single hyperparameter we expect this approach to work well with a fairly small validation data (without hurting the minority class sample size).

\subsection{CS-SVM as post-hoc weight normalization}\label{sec:lemma1_proof}

We need the lemma below to prove Theorem \ref{propo:delta_star}. But the results is interesting on its own right as it allows us to view CS-SVM as an appropriate ``post-hoc weight normalization"-approach. 

\begin{lemma}\label{lem:1_to_delta}
Let $(\hat\w_1,\hat b_1)$ be the hard-margin SVM solution. Fix any $\delta>0$ in \eqref{eq:CS-SVM} and define: 
$
\hat\w_\delta:=\big(\frac{\delta+1}{2}\big)\,{\hat\w_1}$
and $\hat b_\delta:=\big(\frac{\delta+1}{2}\big)\,\hat b_1 + \big(\frac{\delta-1}{2}\big).
$
Then, $(\hat\w_\delta,\hat b_\delta)$ is optimal in \eqref{eq:CS-SVM}.
\end{lemma}
Thus, classification using \eqref{eq:CS-SVM} is equivalent to the following. First learn $(\hat\w_1,\hat b_1)$ via standard hard-margin SVM, and then simply predict:
$
\hat y = {\rm sign}\big( (\hat\w_1^T\x + \hat b_1) + \frac{\delta-1}{\delta+1} \big).
$
The  term $\frac{\delta-1}{\delta+1}$ can be seen as an additive form of post-hoc weight normalization to account for class imbalances. In the literature this post-hoc adjustment of the threshold $b$ of standard SVM is often referred to as boundary-movement SVM (BM-SVM) \cite{shawe1999optimizing,wu2003class}. Here, we have shown the equivalence of CS-SVM to BM-SVM for a specific choice of the boundary shift. The proof of Lemma \ref{lem:1_to_delta} presented in Appendix \ref{sec:lemma1_proof} shows the desired using the KKT conditions of \eqref{eq:CS-SVM}.

\begin{proof}
From optimality of $(\hat\w_1,\hat b_1)$, convexity of \eqref{eq:CS-SVM} and the KKT-conditions, there exist dual variables $\beta_i, i\in[n]$ such that:
\begin{align}\label{eq:KKT_1}
&\hat \w_1 = \sum_{i\in[n]}y_i\beta_i \x_i,\quad\sum_{i\in[n]}y_i\beta_i = 0,\\
&\forall i\in[n]~:~\beta_i\big(\x_i^T\hat \w_1+\hat b_1\big)=\beta_iy_i\nn,\quad \beta_i\geq 0.
\end{align}
Let $(\hat\w_\delta,\hat b_\delta)$ defined as in the statement of the lemma and further define $\epsilon_i:=\big(\frac{\delta+1}{2}\big)\beta_i,~i\in[n]$. Then, it only takes a few algebra steps using \eqref{eq:KKT_1} to check that the following conditions hold:
\begin{align}\label{eq:KKT_2}
&\hat \w_\delta = \sum_{i\in[n]}y_i\epsilon_i \x_i,\quad\sum_{i\in[n]}y_i\epsilon_i = 0,\\
&\forall i\in[n]~:~\epsilon_i\big(\x_i^T\hat \w_\delta+\hat b_\delta\big)=\epsilon_i\cdot\begin{cases}\delta &,\text{if } y_i=+1\\ -1 &,\text{if } y_i=-1\end{cases}\nn,\quad \epsilon_i\geq 0.
\end{align}
It can also be verified that \eqref{eq:KKT_2} are the KKT conditions of the CS-SVM with parameter $\delta$. This proves that $(\hat\w_\delta,\hat b_\delta)$ is optimal in \eqref{eq:CS-SVM} as desired.
\end{proof}

\subsection{Proof of Theorem \ref{propo:delta_star}}\label{sec:delta_star_proof}

As discussed in the section above the proof proceeds in two steps:

(i) First, starting from \eqref{eq:delta_tune_1}, we prove \eqref{eq:delta_tune_2}.

(ii) Second, we analytically solve \eqref{eq:delta_tune_2} to derive the explicit expression for $\delta_\star$ in \eqref{eq:delta_star_app}.

\noindent\textbf{Proof of \eqref{eq:delta_tune_2}.}~
Fix any $\delta>0.$ From Lemma \ref{lem:1_to_delta},
\begin{align}\label{eq:tune_proof_0}
\wh_\delta= \big(\frac{\delta+1}{2}\big) {\wh_1}\quad\text{and}\quad \bh_\delta=\big(\frac{\delta+1}{2}\big)\bh_1+\big(\frac{\delta-1}{2}\big).
\end{align}
Recall from Theorem \ref{thm:main_imbalance} that $\|\wh_\delta\|_2\rP q_\delta$, $\|\wh_1\|_2\rP q_1$, $\bh_\delta\rP b_\delta$, $\bh_1\rP b_1$, and, for $i=1,2$: $\frac{\wh_\delta^T\mub_i}{\|\wh_\delta\|_2}\rP\eb_i^T\Vb\Sb\rhob_\delta$ and $\frac{\wh_1^T\mub_i}{\|\wh_1\|_2}\rP\eb_i^T\Vb\Sb\rhob_1$. Here, $q_\delta,\rho_\delta,b_\delta$ and $q_1,\rho_1,b_1$ are as defined in Theorem \ref{thm:main_imbalance}. Thus, from \eqref{eq:tune_proof_0} we find that
\begin{align}
\rhob_\delta = \rhob_1,\qquad q_\delta=\big(\frac{\delta+1}{2}\big) q_1\qquad \text{and} \qquad b_\delta=\big(\frac{\delta+1}{2}\big)b_1 + \big(\frac{\delta-1}{2}\big).
\end{align}
Hence, it holds:
$$
 Q\left(\eb_1^T\Vb\Sb\rhob_\delta+b_\delta/q_\delta\right) =  Q\Big(\underbrace{\eb_1^T\Vb\Sb\rhob_\delta+b_1/q_1}_{=\ell_+}+\frac{\delta-1}{\delta+1}q_1^{-1}\Big).
$$
A similar expression can be written for the conditional error of class $-1$. Putting these together shows \eqref{eq:delta_tune_2}, as desired.

\noindent\textbf{Proof of \eqref{eq:delta_star_app}.}~Recall from \eqref{eq:delta_tune_2} that we now need to solve the following constrained minimization where for convenience we call $a=\ell_+$, $b=\ell_-$ and $c=q_1^{-1}$:
$$
\min_{\delta>0}~Q\left(a+\frac{\delta-1}{\delta+1}c\right) + Q\left(b-\frac{\delta-1}{\delta+1}c\right).
$$
We define a new variable $x=\frac{\delta-1}{\delta+1}c$. The constraint $\delta>0$ then writes $x\leq c$. This is because the function $\delta\in(0,\infty)\mapsto\frac{\delta-1}{\delta+1}$ is onto the interval $(-1,1).$


Thus, we equivalently need to solve
$$
\min_{-c<x< c}~f(x):=Q(a+x) + Q(b-x).
$$

Define function $f(x) = Q(a+x)+Q(b-x)$ for some $a,b\in\R$. 
Direct differentiation gives
$
\frac{{\rm d}f}{{\rm d}x} = \frac{1}{\sqrt{2\pi}}\left(e^{-(b-x)^2/2}-e^{-(a+x)^2/2}\right).
$
Furthermore, note that $\lim_{x\rightarrow\pm\infty}f(x)=1$. 
With thes and some algebra it can be checked that $f(\cdot)$ behaves as follows depending on the sign of $a+b.$ Denote $x_\star = (b-a)/2$.

\begin{itemize}
\item If $a+b\geq 0$, then $1>f(x)\geq f(x_\star)$ and $x_\star$ is the unique minimum.
\item If $a+b< 0$, then $1<f(x)\leq f(x_\star)$ and $x_\star$ is the unique maximum.
\end{itemize}

Thus, we conclude with the following:
\begin{align*}
\arg\inf_{-c<x< c}~f(x) = \begin{cases}
x_\star & \text{if } a+b\geq0 \text{ and } b-a< 2c,\\
c & \text{if } a+b\geq0 \text{ and } b-a\geq 2c, \\
-c & \text{if } a+b<0.
\end{cases}
\end{align*}

Equivalently,
\begin{align*}
\arg\inf_{\delta>-1}~Q\left(\ell_++\frac{\delta-1}{\delta+1}q_1^{-1}\right) + Q\left(\ell_--\frac{\delta-1}{\delta+1}q_1^{-1}\right) = \begin{cases}
\frac{\ell_- -\ell_++2q_1^{-1}}{\ell_+-\ell_-+2q_1^{-1}} & \text{if } \ell_++\ell_-\geq0 \text{ and } \ell_+-\ell_-+2q_1^{-1}> 0,\\
\infty & \text{if } \ell_++\ell_-\geq0 \text{ and } \ell_+-\ell_-+2q_1^{-1}\leq 0, \\
 0 & \text{if } \ell_++\ell_-<0.
\end{cases}
\end{align*}

This shows \eqref{eq:delta_star_app}. The remaining statement of the theorem is easy to prove requiring simple algebra manipulations.

%% file: asymptotic.tex
\section{Asymptotic analysis of CS-SVM}\label{sec:cs-svm_proof}

%

\subsection{Preliminaries}
The main goal of this appendix is proving Theorem \ref{thm:main_imbalance}.  For fixed $\delta>0$, let $(\wh,\hat b)$ be the solution to the CS-SVM in \eqref{eq:CS-SVM}. (See also \eqref{eq:CS-SVM_con} below.) In the following sections, we will prove the following convergence properties for the solution of the CS-SVM:
\begin{align}\label{eq:2prove_app}
(\|\hat\w\|_2, \frac{\wh^T\mub_+}{\|\wh\|_2}, \frac{\wh^T\mub_-}{\|\wh\|_2} , \hat b) \rP (q_\delta,\eb_1^T\Vb\Sb\rhob_\delta,\eb_2^T\Vb\Sb\rhob_\delta, b_\delta).
\end{align}
where the triplet $(q_\delta,\rhob_\delta,b_\delta)$ is as defined in the theorem's statement, that is, the unique triplet satisfying
\begin{align}\label{eq:eta_eq_main}
\eta_\delta(q_\delta,\rhob_\delta,b_\delta)= 0\quad\text{and}\quad
(\rhob_\delta,b_\delta):=\arg\min_{\substack{\|\rhob\|_2\leq 1, b\in\R}}\eta_\delta(q_\delta,\rhob,b).
\end{align}
%

 In this section, we show how to use \eqref{eq:2prove_app} to derive the asymptotic limit of the conditional class probabilities.

Consider the class conditional
$\Rc_+=\Pro\left\{ (\x^T\wh+b)<0 \,|\, y=+1 \right\}$. Recall that conditioned on $y=+1$, we have $\x=\mub_++\z$ for $\z\sim\Nn(\mathbf{0},\mathbf{I}).$ Thus, the class conditional can be expressed explicitly in terms of the three summary quantities on the left hand side of \eqref{eq:2prove_app} as follows:
\begin{align*}
\Rc_+ &=\Pro\left\{ (\x^T\wh+\hat b)<0 \,|\, y=+1 \right\} = \Pro\left\{ \z^T\wh+ \mub_+^T\wh+\hat b<0 \,|\, y=+1 \right\}\nn
\\
&=\Pro\left\{ \z^T\wh > \mub_+^T\wh+\hat b \right\} 
\\
&= \Pro_{G\sim\Nn(0,1)}\left\{ G\|\wh\|_2 > \mub_+^T\wh+\hat b \right\} = \Pro_{G\sim\Nn(0,1)}\left\{ G > \frac{\mub_+^T\wh}{\|\wh\|_2}+\frac{\hat b}{\|\wh\|_2} \right\}\\
&= Q\left(\frac{\mub_+^T\wh}{\|\wh\|_2}+\frac{\hat b}{\|\wh\|_2}\right).
\end{align*}
Then, the theorem's statement follows directly by applying \eqref{eq:2prove_app} in the expression above.

In order to prove the key convergence result in \eqref{eq:2prove_app} we rely on the convex Gaussian min-max theorem (CGMT) framework. We give some necessary background before we proceed with the proof.

\subsection{Background and related literature}\label{sec:HD_lit}

\noindent\textbf{Related works:}~Our asymptotic analysis of the CS-SVM fits in the growing recent literature on sharp statistical performance asymptotics of convex-based estimators, e.g. \cite{bean2013optimal,montanari13,thrampoulidis2018precise,thrampoulidis2018symbol} and references therin. The origins of these works trace back to the study of sharp phase transitions in compressed sensing, e.g. see \cite{thrampoulidis2018precise} for historical remarks 
 and performance analysis of the LASSO estimator for sparse signal recovery. That line of work led to the development of two analysis frameworks: (a) the approximate  message-passing (AMP) framework \cite{BayMon,AMPmain}, and, (b) the convex Gaussian min-max theorem (CGMT) framework \cite{StojLAS,thrampoulidis2015regularized}. More recently, these powerful tools have proved very useful for the analysis of linear classifiers \cite{salehi2019impact,montanari2019generalization,deng2019model,gkct_dd2020,mignacco2020role,liang2020precise,taheri2020sharp,candes2020phase,aubin2020generalization,taheri2020fundamental}. Theorems \ref{thm:main_imbalance} and \ref{thm:main_group_fairness} rely on the CGMT and contribute to this line of work. Specifically, our results are most closely related to \cite{deng2019model} who first studied max-margin type classifiers together with \cite{montanari2019generalization}. 

\vp
\noindent\textbf{CGMT framework:}~Specifically, we rely on the CGMT framework. Here, we only 
%
%
%
 summarize the framework's essential ideas and refer the reader to \cite{thrampoulidis2015regularized,thrampoulidis2018precise} for more details and precise statements. Consider the following two Gaussian processes:
\begin{subequations}\label{eq:POAO}
\begin{align}
X_{\w,\ub} &:= \ub^T \A \w + \psi(\w,\ub),\\
Y_{\w,\ub} &:= \norm{\w}_2 \h_n^T \ub + \norm{\ub}_2 \h_d^T \w + \psi(\w,\ub),\label{eq:AO_obj}
\end{align}
\end{subequations}
where: $\A\in\mathbb{R}^{n\times d}$, $\h_n \in \mathbb{R}^n$, $\h_d\in\mathbb{R}^d$, they all have entries iid Gaussian; the sets $\mathcal{S}_{\w}\subset\R^d$ and $\mathcal{S}_{\ub}\subset\R^n$ are compact; and, $\psi: \mathbb{R}^d\times \mathbb{R}^n \to \mathbb{R}$. For these two processes, define the following (random) min-max optimization programs, which are refered to as the \emph{primary optimization} (PO) and the \emph{auxiliary optimization} (AO) problems:
\begin{subequations}
\begin{align}\label{eq:PO_loc}
\Phi(\A)&=\min\limits_{\w \in \mathcal{S}_{\w}} \max\limits_{\ub\in\mathcal{S}_{\ub}} X_{\w,\ub},\\
\label{eq:AO_loc}
\phi(\h_n,\h_d)&=\min\limits_{\w \in \mathcal{S}_{\w}} \max\limits_{\ub\in\mathcal{S}_{\ub}} Y_{\w,\ub}.
\end{align}
\end{subequations}

According to the first statement of the CGMT Theorem 3 in \cite{thrampoulidis2015regularized} (this is only a slight reformulation of Gordon's original comparison inequality \cite{Gor2}), for any $c\in\R$, it holds:
\begin{equation}\label{eq:gmt}
\mathbb{P}\left\{ \Phi(\A) < c\right\} \leq 2\, \mathbb{P}\left\{  \phi(\h_n,\h_d) < c \right\}.
\end{equation}
In other words, a high-probability lower bound on the AO is a high-probability lower bound on the PO. The premise is that it is often much simpler to lower bound the AO rather than the PO. 
However, the real power of the CGMT comes in its second statement, which asserts that if the PO is \emph{convex} then the AO in  can be used to tightly infer properties of the original PO, including the optimal cost and the optimal solution.
More precisely, if the sets $\mathcal{S}_{\w}$ and $\mathcal{S}_{\ub}$ are convex and \emph{bounded}, and $\psi$ is continuous \emph{convex-concave} on $\mathcal{S}_{\w}\times \mathcal{S}_{\ub}$, then, for any $\nu \in \mathbb{R}$ and $t>0$, it holds \cite{thrampoulidis2015regularized}:
\begin{equation}\label{eq:cgmt}
\mathbb{P}\left\{ \abs{\Phi(\A)-\nu} > t\right\} \leq 2\, \mathbb{P}\left\{  \abs{\phi(\h_n,\h_d)-\nu} > t \right\}.
\end{equation}
In words, concentration of the optimal cost of the AO problem around $q^\ast$ implies concentration of the optimal cost of the corresponding PO problem around the same value $q^\ast$.  Asymptotically, if we can show that $\phi(\h_n,\h_d)\rP q^\ast$, then we can conclude that $\Phi(\A)\rP q^\ast$. 

In the next section, we will show that we can indeed express the CS-SVM in \eqref{eq:CS-SVM} as a PO in the form of \eqref{eq:PO_loc}. Thus, the argument above will directly allow us to determine the asymptotic limit of the optimal cost of the CS-SVM. In our case, the optimal cost equals $\|\wh\|_2$; thus, this shows the first part of \eqref{eq:2prove_app}. For the other parts, we will employ the following ``deviation argument" of the CGMT framework \cite{thrampoulidis2015regularized}. For arbitrary $\eps>0$, consider the desired set 
\begin{align}\label{eq:Sc}
\Sc:=\left\{ (\vb,c)~\Big|~ \max\Big\{\abs{\|\vb\|_2- q_\delta}\,,\, \Big|\frac{\vb^T\mub_+}{\|\vb\|_2} - \eb_1^T\Vb\Sb\rhob_\delta\Big|,,\,\Big|\frac{\vb^T\mub_-}{\|\vb\|_2} - \eb_2^T\Vb\Sb\rhob_\delta\Big|
\,,\, |c-  b_\delta| \Big\}\leq\eps\, \right\}.
\end{align}
Our goal towards \eqref{eq:2prove_app} is to show that with overwhelming probability $(\w,b)\in\Sc$. For this, consider the following constrained CS-SVM that further constraints the feasible set to the complement $\Sc^c$ of $\Sc$:
\begin{align}\label{eq:CS-SVM_con}
\hspace{-0.1in}\Phi_{\Sc^c}(\A):=\min_{(\w,b)\in\Sc^c}~\|\w\|_2~\text{sub. to} \begin{cases} \w^T\x_i+b \geq\delta &\hspace{-0.05in}, y_i=+1 \\  \w^T\x_i+b\leq -1 &\hspace{-0.05in}, y_i=-1 \end{cases}, i\in[n],
\end{align}
As per Theorem 6.1(iii) in \cite{thrampoulidis2018precise} it will suffice to find costants $\bar\phi, \bar\phi_S$ and $\eta>0$ such that the following three conditions hold:
\begin{align}\label{eq:cgmt_conditions}
\begin{cases}
\text{(i)~~\,\, $\bar\phi_S \geq \bar\phi + 3\eta$}\\
\text{(ii)~~ \,$\phi(\h_n,\h_d) \leq \bar\phi + \eta$~~ with overwhelming probability}\\
\text{(iii)~~ $\phi_{\Sc^c}(\h_n,\h_d) \geq \bar\phi_S - \eta$~~ with overwhelming probability,}
\end{cases}
\end{align}
where $\phi_{\Sc^c}(\h_n,\h_d)$ is the optimal cost of the constrained AO corresponding to the constrained PO in \eqref{eq:CS-SVM_con}.

To prove these conditions for the AO of the CS-SVM, in the next section we follow the  principled machinery of \cite{thrampoulidis2018precise} that allows simplifying the AO from a (random) optimization over vector variables to an easier optimization over only few scalar variables, termed the ``scalarized AO".

\subsection{Proof of Theorem \ref{thm:main_imbalance}}\label{sec:cs_proof_app}

Let $(\wh,\bh)$ be solution pair to the CS-SVM in \eqref{eq:CS-SVM} for some fixed margin-ratio parameter $\delta>0$, which we rewrite here expressing the constraints in matrix form:
\begin{align}\label{eq:CS-SVM_app0}
\min_{\w,b}\|\w\|_2~~\text{sub. to}~\begin{cases} \w^T\x_i+b \geq\delta ,~y_i=+1 \\  -(\w^T\x_i+b)\geq 1,~y_i=-1 \end{cases}\hspace{-10pt},~i\in[n]~~=~~ \min_{\w,b}\|\w\|_2~~\text{sub. to}~\Db_\y(\X\w+b\ones_n)\geq \deltab_\y,
\end{align}
where we have used the notation
\begin{align*}
\X^T&=\begin{bmatrix} \x_1 & \cdots & \x_n \end{bmatrix},~\y=\begin{bmatrix} y_1 & \cdots & y_n \end{bmatrix}^T,
\\\Db_\y&=\rm{diag}(\y)~\text{and}~\deltab_\y=\begin{bmatrix}
\delta\ind{y_1=+1}+\ind{y_1=-1}& \cdots &\delta\ind{y_n=+1}+\ind{y_n=-1}
\end{bmatrix}^T.
\end{align*}
We further need to define the following one-hot-encoding of the labels: 
$$
\y_i = \eb_1 \ind{y_i=1} + \eb_2 \ind{y_i=-1},\quad\text{and}\quad \Y_{n\times 2}^T=\begin{bmatrix}\y_1 & \cdots & \y_n \end{bmatrix}.
$$
where recall that $\eb_1,\eb_2$ are standard basis vectors in $\R^2$. 

With these, 
notice for later use that under our model, $\x_i=\mub_{y_i}+\z_i = \M\y_i+\z_i,~\z_i\sim\Nn(0,1)$. Thus, in matrix form with $\Z$ having entries $\Nn(0,1)$:
\begin{align}\label{eq:model_cs}
\X = \Y\M^T+\Z.
\end{align}

Following the CGMT strategy \cite{thrampoulidis2015regularized}, we express \eqref{eq:CS-SVM_app0} in a min-max form to bring it in the form of the PO as follows:
\begin{align}
&\min_{\w,b}\max_{\ub\leq0}~\frac{1}{2}\|\w\|_2^2+ \ub^T\Db_\y\X\w+b(\ub^T\Db_\y\ones_n)-\ub^T\deltab_\y\nn\\
=&\min_{\w,b}\max_{\ub\leq0}~\frac{1}{2}\|\w\|_2^2+ \ub^T\Db_\y\Z\w + \ub^T\Db_\y\Y\M^T\w +b(\ub^T\Db_\y\ones_n)-\ub^T\deltab_\y.\label{eq:cs_PO}
\end{align}
where in the last line we used \eqref{eq:model_cs} and $\Db_\y\Db_\y=\mathbf{I}_n$.
We immediately recognize that the last optimization is in the form of a PO (cf. \eqref{eq:PO_loc}) and the corresponding AO (cf. \eqref{eq:AO_loc}) is as follows:
\begin{align}\label{eq:cs_AO}
\min_{\w,b}\max_{\ub\leq0}~ \frac{1}{2}\|\w\|_2^2+\|\w\|_2\ub^T\Db_\y\h_n + \|\Db_\y\ub\|_2\h_d^T\w + \ub^T\Db_\y\Y\M^T\w +b(\ub^T\Db_\y\ones_n)-\ub^T\deltab_\y.
\end{align}
where $\h_n\sim\Nn(0,\mathbf{I}_n)$ and $\h_d\sim\Nn(0,\mathbf{I}_d)$. 

In order to apply the CGMT in \cite{thrampoulidis2015regularized}, we need boundedness of the constraint sets. Thus, we restrict the minimization in \eqref{eq:cs_AO} and \eqref{eq:cs_PO} to a bounded set $\|\w\|_2^2+b^2\leq R$ for (say) $R:=2\left(  q_\delta^2+  b_\delta^2\right)$. This will allow us to show that the solutions $\wh_R,\hat b_R$ of this constrained PO satisfy $\wh_R\rP  q_\delta$ and $\hat b_R\rP   b_\delta$. Thus, with overwhelming probability, $\|\wh_R\|_2^2+{\hat b}_R^2 < R$. From this and convexity of the PO, we can argue that the minimizers $\wh,\hat b$ of the original unconstrained problem satisfy the same convergence properties. Please see also Remark 4 in App. A of \cite{deng2019model}.

For the maximization, we follow the recipe in App. A of  \cite{deng2019model} who analyzed the standard SVM. Specifically, combining Remark 3 of \cite{deng2019model} together with (we show this next) the property that the AO is reduced to a convex program, it suffices to consider the unconstrained maximization. 

Thus, in what follows we consider the one-sided constrained AO in \eqref{eq:cs_AO}. Towards simplifying this auxiliary optimization, note that $\Db_\y\h_n\sim\h_n$ by rotational invariance of the Gaussian measure. Also, $\|\Db_\y\ub\|_2=\|\ub\|_2$. Thus, we can express the AO in the following more convenient form:
\begin{align}\label{eq:cs_AO_11}
&\min_{\|\w\|_2^2+b^2\leq R}\max_{\ub\leq0}~\frac{1}{2}\|\w\|_2^2+ \|\w\|_2\ub^T\h_n + \|\ub\|_2\h_d^T\w + \ub^T\Db_\y\Y\M^T\w +b(\ub^T\Db_\y\ones_n)-\ub^T\deltab_\y.
\end{align}
We are now ready to proceed with simplification of the AO. First we optimize over the direction of $\ub$ and rewrite the AO as
\begin{align*}
&\min_{\|\w\|_2^2+b^2\leq R}\max_{\beta\geq 0}~\frac{1}{2}\|\w\|_2^2+ \beta\left( \Big\| \big( \, \|\w\|_2\h_n +  \Db_\y\Y\M^T\w+b\, \Db_\y\ones_n- \deltab_{\y} \, \big)_- \Big\|_2 -\h_d^T\w \right)
\\
=&\min_{\|\w\|_2^2+b^2\leq R}~\frac{1}{2}\|\w\|_2^2\quad\text{sub. to}~\Big\| \big( \, \|\w\|_2\h_n +  \Db_\y\Y\M^T\w+b\, \Db_\y\ones_n- \deltab_{\y} \, \big)_- \Big\|_2 \leq \h_d^T\w.
\end{align*}
Above, $(\cdot)_-$ acts elementwise to the entries of its argument.

Now, we wish to further simplify the above by minimizing over the direction of $\w$ in the space orthogonal to $\M$. To see how this is possible consider the SVD $\M^T=\Vb\Sb\Ub^T$ and project $\w$ on the columns of $\Ub=\begin{bmatrix} 
\ub_1 & \ub_2\end{bmatrix}\in\R^{d\times 2}$ as follows:
$$
\w = \ub_1(\ub_1^T\w) + \ub_2(\ub_2^T\w) + \w^\perp,
$$
where $\w^\perp=\Ub^\perp\w$, $\Ub^\perp$ is the orthogonal complement of $\Ub$. For simplicity we will assume here  that $\M$ is full column rank, i.e. $\Sb\succ\mathbf{0}_{2\times 2}$. The argument for the case where $\M$ is rank 1 is very similar. 

Let us denote $\ub_i^T\w:=\mu_i, i=1,2$ and $\|\w^\perp\|_2:=\alpha$. In this notation, the AO becomes
\begin{align*}
&\min_{\mu_1^2+\mu_2^2+\|\w^\perp\|_2^2+b^2\leq R}~\frac{1}{2}(\mu_1^2+\mu_2^2+\alpha^2)\quad\\
&\text{sub. to}\quad\Big\| \big( \, \sqrt{\mu_1^2+\mu_2^2+\alpha^2} \h_n +  \Db_\y\Y\Vb\Sb\begin{bmatrix}\mu_1 \\ \mu_2 \end{bmatrix} +b\, \Db_\y\ones_n- \deltab_{\y} \, \big)_- \Big\|_2\\ &\quad\quad\quad\leq \mu_1(\h_d^T\ub_1) + \mu_2(\h_d^T\ub_2) + \h_d^T\Ub^\perp\w^\perp.
\end{align*}
At this point, we can optimize over the direction of $\w^\perp$ which leads to 
\begin{align*}
&\min_{ \mu_1^2+\mu_2^2+\alpha^2+b^2\leq R }~\frac{1}{2}(\mu_1^2+\mu_2^2+\alpha^2)\quad
\\
&\text{sub. to}\quad\Big\| \big( \, \sqrt{\mu_1^2+\mu_2^2+\alpha^2} \h_n +  \Db_\y\Y\Vb\Sb\begin{bmatrix}\mu_1 \\ \mu_2 \end{bmatrix} +b\, \Db_\y\ones_n- \deltab_{\y} \, \big)_- \Big\|_2\\ &\quad\quad\quad \leq \mu_1(\h_d^T\ub_1) + \mu_2(\h_d^T\ub_2) + \alpha\|\h_d^T\Ub^\perp\|_2.
\end{align*}
As a last step in the simplification of the AO, it is convenient to introduce an additional variable 
$
q=\sqrt{\mu_1^2+\mu_2^2+\alpha^2}.
$
It then follows that the minimization above is equivalent to the following
\begin{align}\label{eq:cs_AO_scal_cvx}
&\min_{ \substack{ q\geq \sqrt{\mu_1^2+\mu_2^2+\alpha^2} \\ q^2+b^2\leq R} }~\frac{1}{2}q^2\quad
\\
&\text{sub. to}\quad\Big\| \big( \, q \h_n +  \Db_\y\Y\Vb\Sb\begin{bmatrix}\mu_1 \\ \mu_2 \end{bmatrix} +b\, \Db_\y\ones_n- \deltab_{\y} \, \big)_- \Big\|_2 \leq \mu_1(\h_d^T\ub_1) + \mu_2(\h_d^T\ub_2) + \alpha\|\h_d^T\Ub^\perp\|_2.\nn
\end{align}
In this formulation it is not hard to check that the optimization is jointly convex in its variables $(\mu_1,\mu_2,\alpha,b,q)$. To see this note that: (i) the constraint $q\geq \sqrt{\mu_1^2+\mu_2^2+\alpha^2} \iff q\geq \|\begin{bmatrix} \mu_1 & \mu_2 & \alpha \end{bmatrix}\|_2$ is a second-order cone constraint, and, (ii) the function
\begin{align}\label{eq:Ln}
\Lc_n(q,\mu_1,\mu_2,\alpha,b)&:=\frac{1}{\sqrt{n}}\Big\| \big( \, q \h_n +  \Db_\y\Y\Vb\Sb\begin{bmatrix}\mu_1 \\ \mu_2\end{bmatrix} +b\, \Db_\y\ones_n- \deltab_{\y} \, \big)_- \Big\|_2\nn\\  &\quad- \mu_1\frac{\h_d^T\ub_1}{\sqrt{n}} - \mu_2\frac{\h_d^T\ub_2}{\sqrt{n}} - {\alpha}\frac{\|\h_d^T\Ub^\perp\|_2}{\sqrt{n}}
\end{align}
is also convex since $\|(\cdot)_-\|_2:\R^n\rightarrow\R$ is itslef convex and is composed here with an affine function. 

Now, by law of large numbers, notice that for fixed $(q,\mu_1,\mu_2,\alpha,b)$, $\Lc_n$ converges in probability to 
\begin{align}\label{eq:Ln2L_cs}
\Lc_n(q,\mu_1,\mu_2,\alpha,b)\rP L(q,\mu_1,\mu_2,\alpha,b):=\sqrt{\E\big(qG+E_{Y}^T\Vb\Sb\begin{bmatrix}\mu_1 \\ \mu_2\end{bmatrix} + {b\,Y-\Delta_Y}\big)_{-}^2} - \alpha\sqrt{\gamma},
\end{align}
where the random variables $G,E_Y,Y,\Delta_Y$ are as in the statement of the theorem. But convergence of convex functions is uniform over compact sets as per Cor.~II.I in \cite{andersen1982cox}. Therefore,  the convergence in \eqref{eq:Ln2L_cs} is in fact uniform in the compact feasible set of \eqref{eq:cs_AO_scal_cvx}. 

Consider then the deterministic high-probability equivalent of \eqref{eq:cs_AO_scal_cvx} which is the following convex program:
\begin{align}
&\min_{ \substack{ q\geq \sqrt{\mu_1^2+\mu_2^2+\alpha^2} \\ q^2+b^2\leq R \\ L(q,\mu_1,\mu_2,\alpha,b)\leq 0} }~\frac{1}{2}q^2\nn.
\end{align}
Since $q$ is positive and the constraint $q\geq\sqrt{\mu_1^2+\mu_2^2+\alpha^2}$ must be active at the optimum, it is convenient to rewrite this in terms of new variables
$
\rhob = \begin{bmatrix}\rhob_1\\\rhob_2 \end{bmatrix}:=\begin{bmatrix}\mu_1/q \\ \mu_2/q\end{bmatrix}
$
as follows:
\begin{align}\label{eq:cs_AO_scal_det2}
&\min_{  q^2+b^2 \leq R, q>0, \|\rhob\|_2\leq 1 }~\frac{1}{2}q^2
\\
&\text{sub. to}\quad  {\E\Big[\big(G+E_{Y}^T\Vb\Sb\rhob + \frac{b\,Y-\Delta_Y}{q}\big)_{-}^2\Big]} \nn \leq \left({1-\|\rhob\|_2^2}\right){\gamma}.
\end{align}
Now, recall the definition of the function $ \eta_\delta$ in the statement of the theorem and  observe that the constraint above is nothing but
$$
 \eta_\delta(q,\rhob,b)\leq 0.
$$
Thus, \eqref{eq:cs_AO_scal_det2} becomes
\begin{align}\label{eq:cs_AO_scal_det3}
\min\left\{q^2~\Big|~ 0\leq q\leq \sqrt{R} \quad\text{ and }\quad \min_{b^2 \leq R-q^2, \|\rhob\|_2\leq 1}  \eta_\delta(q,\rhob,b)\leq 0 \right\}.
\end{align}

We will prove that 
\begin{align}\label{eq:eta_dec_cs}
\text{the function $f(q):=\min_{b,\|\rhob\|_2\leq1}  \eta_\delta(q,\rhob,b)$ is strictly decreasing}.
\end{align}
Before that, let us see how this completes the proof of the theorem. Let $  q_\delta$ be as in the statement of the theorem, that is such that $f(  q_\delta)=0$. Then, we have the following relations
$$
f(q)\leq 0 ~\Rightarrow~ f(q) \leq f(  q_\delta) ~\Rightarrow~ q \geq   q_\delta.
$$
Thus, the minimizers in \eqref{eq:cs_AO_scal_det3} are $(  q_\delta,  \rhob_\delta,  b_\delta)$, where we also recall that we have set $R>  q_\delta^2 +   b_\delta^2$.

With all these, we have shown that the AO converges in probability to $  q_\delta^2$ (cf. condition (ii) in \eqref{eq:cgmt_conditions}). From the CGMT, the same is true for the PO. Now, we want to use the same machinery to prove that the minimizers $(\wh,\hat b)$ of the PO satisfy \eqref{eq:2prove_app}. 
To do this, as explained in the previous section, we use the standard strategy of the CGMT framework , i.e., to show that the PO with the additional constraint
$
(\w,b)\in\Sc^c
$
for the set $\Sc$ in \eqref{eq:Sc} has a cost that is strictly larger than $q_\delta^2$ (i.e. the cost of the unconstrained PO). As per the CGMT this can be done again by showing that the statement is true for the correspondingly constrained AO (i.e. show condition (iii) in \eqref{eq:cgmt_conditions}). With the exact same simplifications as above, the latter program simplifies to \eqref{eq:cs_AO_scal_cvx} with the additional constraints:
$$
|q-  q_\delta|>\eps\,,\, \big|\mu_i/q -  \rhob_{\delta,i}\big|>\eps,\, i=1,2\,,\, |b-   b_\delta|>\eps.
$$
Also, using the uniform convergence in \eqref{eq:Ln2L_cs}, it suffices to study the deterministic equivalent \eqref{eq:cs_AO_scal_det3} with the additional constraints above. Now, we can show the desired (cf. condition (i) in \eqref{eq:cgmt_conditions}) again by exploiting \eqref{eq:eta_dec_cs}. This part of the argument is similar to Section C.3.5 in \cite{deng2019model} and we omit the details.

\noindent\underline{Proof of \eqref{eq:eta_dec_cs}:}~To complete the proof, it remains to show \eqref{eq:eta_dec_cs}. Specifically, we show that $\frac{\mathrm{d}f}{\mathrm{d}q}<0$ by combining the following three observations.

First, 
\begin{align}\nn
\frac{\mathrm{\partial} \eta_\delta}{\mathrm{\partial}q} &= \frac{2}{q^2}\E\Big[ (G+ E_Y^T\Vb\Sb\rhob + \frac{b\, Y- \Delta_Y}{q}\big)_{-}\cdot  \Delta_Y \Big] - \frac{2b}{q^2}\E\Big[ (G+ E_Y^T\Vb\Sb\rhob + \frac{b\, Y- \Delta_Y}{q}\big)_{-}\cdot Y \Big]
\\
&<
- \frac{2b}{q^2}\E\Big[ (G+ E_Y^T\Vb\Sb\rhob + \frac{b\, Y- \Delta_Y}{q}\big)_{-}\cdot Y \Big]\label{eq:gs_arg1}
\end{align}
where for the inequality we observed that $(\cdot)_-$ is always non-positive, its argument has non-zero probability measure on the negative real axis, and $ \Delta_Y$ are positive random variables. 

Second, letting $\rhob^\star:=\rhob^\star(q)$ and $b^\star:=b^\star(q)$ the minimizers of $  \eta_\delta(q,\rhob,b)$, it follows from first-order optimality conditions that
\begin{align}\label{eq:gs_arg2}
\frac{\mathrm{\partial} \eta_\delta}{\mathrm{\partial}b} = 0 \iff \E\Big[ (G+ E_Y^T\Vb\Sb\rhob^* + \frac{b^*\, Y- \Delta_Y}{q}\big)_{-}\cdot Y \Big] = 0.
\end{align}

Third, by the envelope theorem
\begin{align}\label{eq:gs_arg3}
\frac{\mathrm{d}f}{\mathrm{d}q} = \frac{\mathrm{\partial} \eta_\delta}{\mathrm{\partial}q}\big|_{\rhob^\star,b^\star}.
\end{align}

The desired inequality $\frac{\mathrm{d}f}{\mathrm{d}q}<0$ follows directly by successively applying \eqref{eq:gs_arg3}, \eqref{eq:gs_arg1} and \eqref{eq:gs_arg2}.

\noindent\textbf{Uniqueness of triplet $(q_\delta,\rhob_\delta,b_\delta)$.}~ First, we prove that the minimizers $\rhob_\delta,b_\delta$ are unique. This follows because $\eta_\delta(q,\rhob,b)$ is jointly strictly convex in $(\rhob,b)$ for fixed $q$. To see this note that the function $x\mapsto (x)_-^2$ is strictly convex for $x<0$ and that the random variable $G+E_Y^T\Vb\Sb\rhob+(bY-\Delta_Y)/q$ has strictly positive measure on the real line (thus, also in the negative axis). Next, consider $q_\delta$, which was defined such that $f(q_\delta)=0$ for the function $f(\cdot)$ in \eqref{eq:eta_dec_cs}. From \eqref{eq:eta_dec_cs} we know that $f(\cdot)$ is strictly decreasing. Thus, it suffices to prove that the function has a zero crossing in $(0,\infty)$, which we do by proving $\lim_{q\rightarrow0} f(q)=\infty$ and $\lim_{q\rightarrow\infty} f(q)<0.$ Specifically, we have
\begin{align*}
\lim_{q\rightarrow0} f(q) &\geq \lim_{q\rightarrow0} \,\min_{b\in\R, \|\rhob\|_2\leq1}{\E\Big[\big(G+E_{Y}^T\Vb\Sb\rhob + \frac{b\,Y-\Delta_Y}{q}\big)_{-}^2\Big]}-\gamma\nn\\
&\geq \lim_{q\rightarrow0}\, \min_{b\in\R, \|\rhob\|_2\leq1}{\E\Big[\big(G+E_{Y}^T\Vb\Sb\rhob + \frac{b\,Y-\Delta_Y}{q}\big)_{-}^2\,\ind{G+E_{Y}^T\Vb\Sb\rhob +(b/q)Y\leq 0}\Big]}-\gamma\\
&\geq \lim_{q\rightarrow0}\, \min_{b\in\R, \|\rhob\|_2\leq1}{1/q^2}-\gamma = \infty,
\end{align*}
where in the last inequality we used the facts that $x\mapsto (x)_-^2$ is decreasing and the event $\{G+E_{Y}^T\Vb\Sb\rhob +(b/q)Y\leq 0\leq 0\}$ has non-zero measure for all $\|\rhob\|_2\leq 1, b\in \R$, as well as, $\Delta_Y\geq 1$ (because $\delta>1$). Moreover,
\begin{align*}
\lim_{q\rightarrow\infty} f(q) &= \lim_{1/q\rightarrow 0^+} f(q) = \lim_{1/q\rightarrow 0^+} \,\min_{b\in\R, \|\rhob\|_2\leq1}{\E\Big[\big(G+E_{Y}^T\Vb\Sb\rhob + \frac{b\,Y-\Delta_Y}{q}\big)_{-}^2\Big]}-(1-\|\rhob\|_2^2)\gamma\\
&=\lim_{1/q\rightarrow 0^+} \,\min_{\tilde{b}\in\R, \|\rhob\|_2\leq1}{\E\Big[\big(G+E_{Y}^T\Vb\Sb\rhob + \tilde{b} Y - \frac{\Delta_Y}{q}\big)_{-}^2\Big]}-(1-\|\rhob\|_2^2)\gamma\\
&\leq\min_{\tilde{b}\in\R, \|\rhob\|_2\leq1}{\E\Big[\big(G+E_{Y}^T\Vb\Sb\rhob + \tilde{b} Y\big)_{-}^2\Big]}-(1-\|\rhob\|_2^2)\gamma\\
&\leq\min_{\tilde{b}\in\R, \|\rhob\|_2\leq1}(1-\|\rhob\|_2^2)\cdot\Big( {\E\Big[\Big((G+E_{Y}^T\Vb\Sb\rhob + \tilde{b} Y)/\sqrt{1-\|\rhob\|_2^2}\Big)_{-}^2\Big]}-\gamma\Big)	\\
&\leq\min_{\tilde{b}\in\R, \|\rhob\|_2\leq1}\, {\E\Big[\Big((G+E_{Y}^T\Vb\Sb\rhob + \tilde{b} Y)/\sqrt{1-\|\rhob\|_2^2}\Big)_{-}^2\Big]}-\gamma	\\
&\leq\min_{\breve{b}\in\R, \tb\in\R^r}\, {\E\Big[\big(\sqrt{1+\|\tb\|_2^2}\,G+E_{Y}^T\Vb\Sb\tb + \breve{b} Y\big)_{-}^2\Big]}-\gamma=\gamma_\star-\gamma	< 0,
\end{align*}
where to get the penultimate inequality we used the change of variables $\tb=\rhob/\sqrt{1-\|\rhob\|_2^2}$ and $\breve{b}=\tilde{b}/\sqrt{1-\|\rhob\|_2^2}$. Also, in the last line above, we used the definition of the phase-transition threshold $\gamma_\star$ in Equation \eqref{eq:gamma_star_CS} and the theorem's assumption that $\gamma>\gamma_\star$ (aka separable regime).

We note that similar uniqueness argument was presented in \cite{deng2019model} for the special case of antipodal means, \emph{no} intercept and $\delta=1.$

\subsection{Antipodal means and non-isotropic data}

\noindent\textbf{Antipodal means.}~In the special case of antipodal means of equal energy $\mub_+=-\mub_-=\mub$ with $s:=\|\mub\|_2$, the formulas of Theorem \ref{thm:main_imbalance} simplify as we have $r=1$ with $\Sb=s\sqrt{2}$ and $\Vb=[1/\sqrt{2}\,,\,-1\sqrt{2}]^T$. Now, the function $\eta_\delta$ can be written as 
$
\E\big[(G+\rhot s + \frac{\bt}{\qt}Y-\frac{1}{\qt}\Delta_Y)_{-}^2\big] \nn- \left(1-{\rhot}^2\right)\gamma.
$
The asymptotic performance of SVM for this special geometry of the means has been recently studied in \cite{deng2019model,mignacco2020role}. We extend this to the CS-SVM classifier, to general means for the two classes and to $\Sigmab\neq\mathbf{I}$.



\noindent\textbf{Non-isotropic data.}~We show how Theorem \ref{thm:main_imbalance} for the isotropic case can still be applied in the general case $\Sigmab\neq \mathbf{I}$. Assume $\Sigmab\succ0$. Write $\x_i=y_i\mub_{y_i}+\Sigmab^{1/2}\h_i$ for $\h_i\sim\Nn(0,\mathbf{I}_d).$ Consider whitened features $\z_i:=\Sigmab^{-1/2}\x_i = y_i\Sigmab^{-1/2}\mub_{y_i}+\h_i$ and let
\begin{align}
(\wh,\hat b) &= \arg\min_{\w,b}\frac{1}{n}\sum_{i\in[n]}\ell(y_i(\x_i^T\w+b)), \nn
\\
(\vbh,\hat c) &= \arg\min_{\vb,c}\frac{1}{n}\sum_{i\in[n]}\ell(y_i(\z_i^T\vb+c)).\nn
\end{align}
Clearly, $\wh=\Sigmab^{-1/2}\vbh$ and $\hat b = \hat c$. 
Thus,
\begin{align}
\Rc_+\left((\wh,\hat b) \right) &= \Pro\{ (\x^T\wh+\hat b) <0 \,|\,y=+1\} = \Pro\{ \mub_+^T\wh+\wh^T\Sigmab^{1/2}\h+\hat b <0 \} = Q\left( \frac{\mub_+^T\wh+\hat b}{\|\Sigmab^{1/2}\wh\|_2} \right) \nn\\
&=Q\Big( \frac{\mub_+^T\Sigmab^{-1/2}\vbh+\hat c}{\|\vbh\|_2} \Big)=\Pro\{ (\z^T\vbh+\hat c) <0 \,|\,y=+1\}\nn\\
&=\Rc_+\left((\vbh,\hat c) \right)\nn
\end{align}
Similar derivation holds for $\Rc_-$. 
Thus, we can just apply Theorem \ref{thm:main_imbalance} for $\Sb,\Vb$ given by the eigendecomposition of the new Grammian $\M^T\Sigmab^{-1}\M$.

\subsection{Phase transition of CS-SVM}\label{sec:PT_CS-SVM}
Here, we present a formula for the threshold $\gamma_\star$ such that the CS-SVM of \eqref{eq:CS-SVM} is feasible (resp., infeasible) with overwhelming probability provided that $\gamma>\gamma_\star$ (resp., $\gamma<\gamma_\star$). The first observation is that the   phase-transition threshold $  \gamma_\star$ of feasibility of  the CS-SVM is the same as the threshold of feasibility of the standard SVM for the same model; see Section \ref{sec:CS_svm_easy_claim}. Then, the desired result follows \cite{gkct_multiclass} who very recently established separability phase-transitions for the more general multiclass Gaussian mixture model

\begin{propo}[\cite{gkct_multiclass}]
 Consider the same data model and notation as in Theorem \ref{thm:main_imbalance} and define the event
$$
\Ec_{{\rm sep}, n}:=\left\{\exists (\w,b)\in\R^d\times\R~~\text{s.t.}~~y_i(\w^T\x_i+b)\geq 1,\,~\forall i\in[n]\right\}.
$$
Define  threshold $\gamma_\star:=\gamma_\star(\Vb,\Sb,\pi)$ as follows:
\begin{align}\label{eq:gamma_star_CS}
\gamma_\star:=\min_{\tb\in\R^r, b\in\R}\E\left[\Big(\sqrt{1+\|\tb\|_2^2} \,G + E_Y^T\Vb\Sb\tb - bY\Big)_-^2\right].
\end{align}
Then, the following hold:
\begin{align}
\gamma>\gamma_\star \Rightarrow \lim_{n\rightarrow\infty}\Pro(\Ec_{{\rm sep}, n}) = 1\quad\text{ and }\quad
\gamma<\gamma_\star \Rightarrow \lim_{n\rightarrow\infty}\Pro(\Ec_{{\rm sep}, n}) = 0.\nn
\end{align}
In words, the data are linearly separable (with overwhelming probability) if and only if $\gamma>\gamma_\star$. Furthermore, if this condition holds, then CS-SVM is feasible (with overwhelming probability) for any value of $\delta>0$.
\end{propo}


%% file: GS-SVM_proofs.tex
\section{Asymptotic analysis of GS-SVM}\label{sec:gs-svm_proof}

\begin{figure}[t]
	\begin{center}
		\includegraphics[width=0.4\textwidth,height=0.3\textwidth]{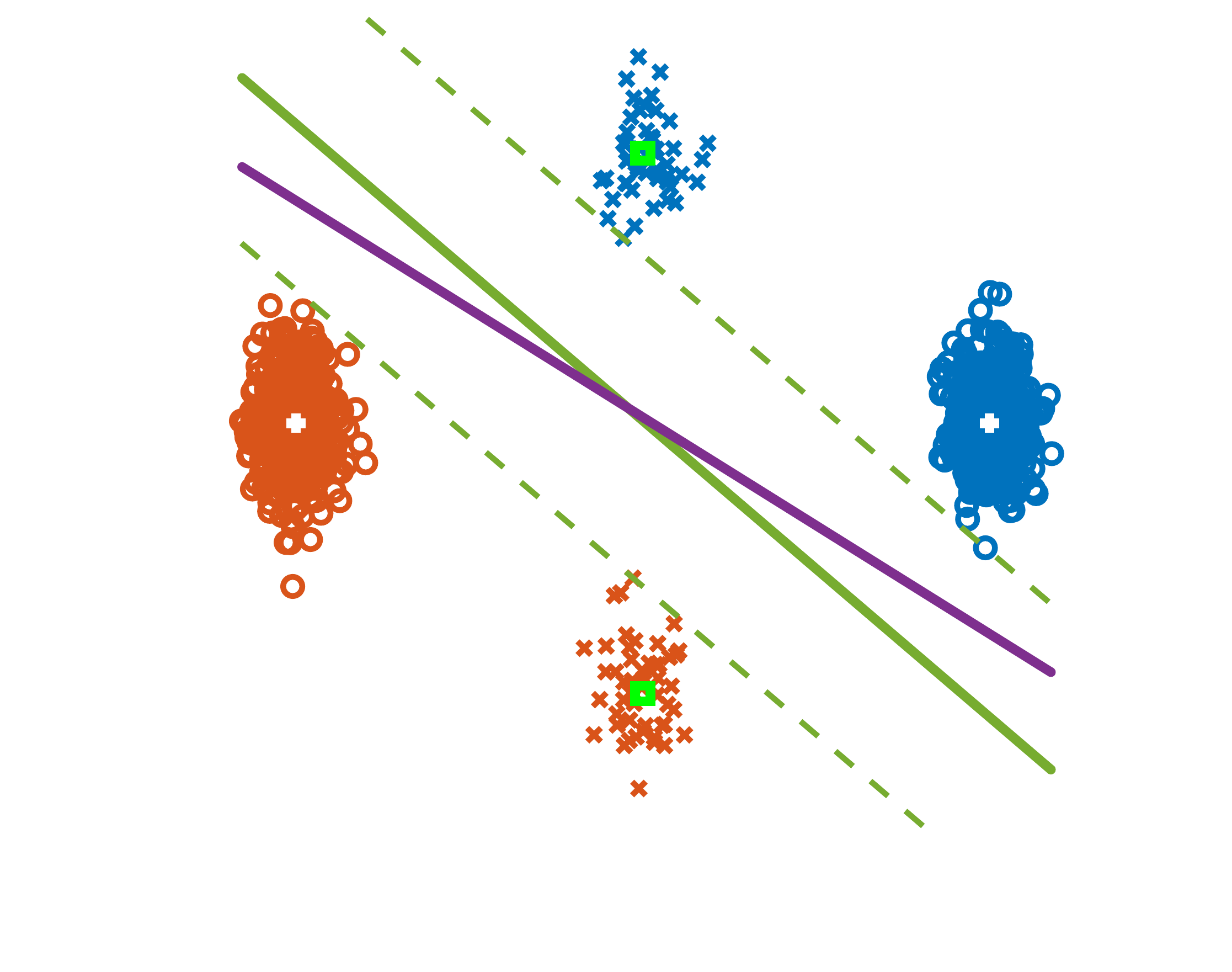}
	\end{center}
	\caption{Visualizing the Gaussian mixture model of Section \ref{sec:generalization} with $K=2$ imbalanced groups in the two-dimensional space ($d=2$). Different colors (resp., markers) correspond to different class (resp., group)  membership. Examples in the minority group correspond to cross markers ($\times$). The means of the majority / minority groups are depicted in white / green markers.  The purple line illustrates the group-sensitive SVM (GS-SVM) classifier that forces larger margin to the minority group examples in relation to standard SVM in green.}
	\label{fig:gssvm_2d}
\end{figure}

In Theorem \ref{thm:main_imbalance} we derived the asymptotic generalization performance of CS-SVM under the Gaussian mixture data model. Here, we state the counterpart result for GS-SVM with an appropriate Gaussian mixture data model with group imbalances, which we repeat here for convenience. 

\noindent\textbf{Data model.}~We study a binary Gaussian-mixture generative model (GMM) for the data distribution $\Dc$. For the label $y\in\{\pm1\}$ let $\pi:=\Pro\{y=+1\}.$ Group membership is decided conditionally on the label such that $\forall j\in[K]:$ $\Pro\{g=j|y=\pm1\}=p_{\pm,j}$, with $\sum_{j\in[K]}p_{+,j}=\sum_{j\in[K]}p_{-,j}=1$. Finally, the feature conditional given label $y$ and group $g$ is a multivariate Gaussian of mean $\mub_{y,g}\in\R^d$ and covariance $\Sigmab_{y,g}$, that is, 
$\x\big|(y,g)\,\widesim{} \Nn(\mub_{y,g},\Sigmab_{y,g}).
$
We focus on two groups $K=2$ with $p_{+,1}=p_{-,1}=p<1-p=p_{+,2}=p_{-,2},\,j=1,2$ and $\x\,|\,(y,g) \sim \Nn(y\mub_{g},\sigma_g\mathbf{I}_d)$, for $\sigma_1^2,\sigma_2^2$ the noise variances of the minority and the majority groups, respectively. As before, let $\M$ denote the matrix of means (that is $\M=\begin{bmatrix} \mub_{+} & \mub_{-}\end{bmatrix}$ and $\M=\begin{bmatrix} \mub_{1} & \mub_{2}\end{bmatrix}$, respectively) and consider the eigen-decomposition of its Gramian:
$
\M^T\M = \Vb\Sb^2\Vb^T,~~ \Sb\succ \mathbf{0}_{r\times r},\Vb\in\R^{2\times r}, r\in\{1,2\},
$
with $\Sb$ an $r\times r$ diagonal positive-definite matrix and $\Vb$ an orthonormal matrix obeying  $\Vb^T\Vb=\mathbf{I}_r$. We study linear classifiers with $h(\x)=\x$. 

\noindent\textbf{Learning regime.}~
Again, as in Theorem \ref{thm:main_imbalance}, we focus on a regime where training data are linearly separable. Specifically, there exists threshold $\wtd\gamma_{\star}:=\wtd\gamma_{\star}(\Vb,\Sb,\pi,p)\leq 1/2$, such
that GMM data with groups are linearly separable with probability approaching one provided that $\gamma>\wtd\gamma_{\star}$ (see Section \ref{sec:PT_GS-SVM}).  
We assume $\gamma>\wtd\gamma_{\star}$, so that GS-SVM is feasible with probability approaching 1.

Although similar in nature, the result below differs to Theorem \ref{thm:main_imbalance} since now each class itself is a Gaussian mixture.

\begin{theorem}[Sharp asymptotics of GS-SVM]\label{thm:GS_app}\label{thm:main_group_fairness}
	Consider the GMM with feature distribution and priors as specified in the `Data model' above. Fix $\delta>0$ (corresponding to group VS-loss with $\Delta_{y,g}=\Delta_{g}, g=1,2$ such that $\delta=\Delta_2/\Delta_1$). Define  $G, Y, S, {\widetilde{\Delta}}_S,\Sigma_S\in\R,$ and $\widetilde{E}_S\in\R^{2\times 1}$ as follows: 
	$G\sim \Nn(0,1)$; $Y$ is a symmetric Bernoulli with $\Pro\{Y=+1\}=\pi$;  $S$ takes values $1$ or $2$ with probabilities $p$ and $1-p$, respectively; 
	$\widetilde{E}_S=\eb_1\ind{S=1}+\eb_2\ind{S=2}$; ${\widetilde{\Delta}}_S=\delta\cdot\ind{S=1} +1\cdot\ind{S=2}$ and $\Sigma_S=\sigma_1\ind{S=1}+\sigma_2\ind{S=2}$.
	 With these define function $\wtd\eta_\delta:\R_{\geq0}\times\Sc^{r}\times\R\rightarrow\R$ as 
$$
		\wtd\eta_\delta(q,\rhob,b):=
		\E\big(G+\Sigma_S^{-1}\widetilde{E}_S^T\Vb\Sb\rhob +\frac{b \Sigma_S^{-1} Y-\Sigma_S^{-1}{\widetilde{\Delta}}_S}{q}\big)_{-}^2 \nn- (1-\|\rhob\|_2^2)\gamma.
$$
	Let $(\wtd q_\delta,\wtd \rhob_\delta, \wtd b_\delta)$ be the unique triplet satisfying \eqref{eq:eta_eq_main} but with $\eta_\delta$ replaced with the function $\wtd \eta_\delta$ above. 
Then, in the limit of $n,d\rightarrow\infty$ with $d/n=\gamma>\wtd\gamma_{\star}$ it holds for $i=1,2$ that
$
\Rc_{\pm,i} \rP Q\big({\eb_i^T\Vb\Sb\wtd \rhob_\delta\pm \wtd b_\delta/{\wtd q_\delta}}\big)$. In particular, 
$
\Rc_{\rm deo} \rP Q\big({\eb_1^T\Vb\Sb\wtd \rhob_\delta+ \wtd b_\delta/{\wtd q_\delta}}\big) - Q\big({\eb_2^T\Vb\Sb\wtd \rhob_\delta+ \wtd b_\delta/{\wtd q_\delta}}\big).$
\end{theorem}

\subsection{Proof of Theorem \ref{thm:GS_app}}
The proof of Theorem \ref{thm:GS_app} also relies on the CGMT framework and is very similar to the proof of Theorem \ref{thm:main_imbalance}. To avoid repetitions, we only present the part that is different. As we will show the PO is slightly different as now we are dealing with a classification between mixtures of mixtures of Gaussians. We will derive the new AO and will simplify it to a point from where the same steps as in Section \ref{sec:cs_proof_app} can be followed mutatis mutandis. 

Let $(\wh,\bh)$ be solution pair to the GS-SVM for some fixed parameter $\delta>0$, which we rewrite here expressing the constraints in matrix form:
\begin{align}\label{eq:GS-SVM_app}
\min_{\w,b}\|\w\|_2~~\text{sub. to}~\begin{cases} y_i(\w^T\x_i+b) \geq\delta ,~g_i=1 \\  y_i(\w^T\x_i+b)\geq 1,~g_i=2 \end{cases}\hspace{-10pt},~i\in[n]~~=~~ \min_{\w,b}\|\w\|_2~~\text{sub. to}~\Db_\y(\X\w+b\ones_n)\geq \deltab_\g,
\end{align}
where we have used the notation
\begin{align*}
\X^T&=\begin{bmatrix} \x_1 & \cdots & \x_n \end{bmatrix},~\y=\begin{bmatrix} y_1 & \cdots & y_n \end{bmatrix}^T,
\\\Db_\y&=\rm{diag}(\y)~\text{and}~\deltab_\g=\begin{bmatrix}
\delta\ind{g_1=1}+\ind{g_1=2}& \cdots &\delta\ind{g_n=1}+\ind{g_n=2}
\end{bmatrix}^T.
\end{align*}
We further need to define the following one-hot-encoding for group membership: 
$$
\g_i = \eb_1 \ind{g_i=1} + \eb_2 \ind{g_i=2},\quad\text{and}\quad \G_{n\times 2}^T=\begin{bmatrix}\g_1 & \cdots & \g_n \end{bmatrix}.
$$
where recall that $\eb_1,\eb_2$ are standard basis vectors in $\R^2$. Finally, let
$$
\Db_\sigma = {\rm diag}\big( \begin{bmatrix} \sigma_{g_1} & \cdots & \sigma_{g_n} \end{bmatrix} \big).
$$
With these, 
notice for later use that under our model, $\x_i=y_i\mub_{g_i}+\sigma_{g_i}\z_i = y_i\M\g_i+\sigma_{g_i}\z_i,~\z_i\sim\Nn(0,1)$. Thus, in matrix form with $\Z$ having entries $\Nn(0,1)$:
\begin{align}\label{eq:model_gs}
\X = \Db_\y\G\M^T+\Db_\sigma\Z.
\end{align}

As usual, we express the GS-SVM program in a min-max form to bring it in the form of the PO as follows:
\begin{align}
&\min_{\w,b}\max_{\ub\leq0}~\frac{1}{2}\|\w\|_2^2+ \ub^T\Db_\y\X\w+b(\ub^T\Db_\y\ones_n)-\ub^T\deltab_\g\nn\\
=&\min_{\w,b}\max_{\ub\leq0}~\frac{1}{2}\|\w\|_2^2+ \ub^T\Db_\y\Db_\sigma\Z\w + \ub^T\G\M^T\w +b(\ub^T\Db_\y\ones_n)-\ub^T\deltab_\g.\label{eq:gs_PO}
\end{align}
where in the last line we used \eqref{eq:model_gs} and $\Db_\y\Db_\y=\mathbf{I}_n$.
We immediately recognize that the last optimization is in the form of a PO and the corresponding AO is as follows:
\begin{align}\label{eq:gs_AO}
\min_{\w,b}\max_{\ub\leq0}~ \frac{1}{2}\|\w\|_2^2+\|\w\|_2\ub^T\Db_\y\Db_\sigma\h_n + \|\Db_\y\Db_\sigma\ub\|_2\h_d^T\w + \ub^T\G\M^T\w +b(\ub^T\Db_\y\ones_n)-\ub^T\deltab_\g.
\end{align}
where $\h_n\sim\Nn(0,\mathbf{I}_n)$ and $\h_d\sim\Nn(0,\mathbf{I}_d)$. 
%
%
%

As in Section \ref{sec:cs_proof_app} we consider the one-sided constrained AO in \eqref{eq:gs_AO}. Towards simplifying this auxiliary optimization, note that $\Db_y\h_n\sim\h_n$ by rotational invariance of the Gaussian measure. Also, $\|\Db_\y\Db_\sigma\ub\|_2=\|\Db_\sigma\ub\|_2$. Thus, we can express the AO in the following more convenient form:
\begin{align*}
&\min_{\|\w\|_2^2+b^2\leq R}\max_{\ub\leq0}~\frac{1}{2}\|\w\|_2^2+ \|\w\|_2\ub^T\Db_\sigma\h_n + \|\Db_\sigma\ub\|_2\h_d^T\w + \ub^T\G\M^T\w +b(\ub^T\Db_\y\ones_n)-\ub^T\deltab_\g\\
=&\min_{\|\w\|_2^2+b^2\leq R}\max_{\vb\leq0}~\frac{1}{2}\|\w\|_2^2+ \|\w\|_2\vb^T\h_n + \|\vb\|_2\h_d^T\w + \vb^T\Db_{\sigma}^{-1}\G\M^T\w +b(\vb^T\Db_\sigma^{-1}\Db_\y\ones_n)-\vb^T\Db_\sigma^{-1}\deltab_\g,
\end{align*}
where in the second line we performed the change of variables $\vb\leftrightarrow\Db_\sigma\ub$ and used positivity of the diagonal entries of $\Db_\sigma$ to find that $\ub\leq0 \iff \vb\leq 0 $.

Notice that the optimization in the last line above is very similar to the AO \eqref{eq:cs_AO_11} in Section \ref{sec:cs_proof_app}. Following analogous steps, omitted here for brevity, we obtain the following scalarized AO:

\begin{align}\label{eq:gs_AO_scal_cvx}
&\min_{ \substack{ q\geq \sqrt{\mu_1^2+\mu_2^2+\alpha^2} \\ q^2+b^2\leq R} }~\frac{1}{2}q^2\quad
\\
&\text{sub. to}\quad\frac{1}{\sqrt{n}}\Big\| \big( \, q \h_n + \Db_\sigma^{-1}\G\Vb\Sb\begin{bmatrix}\mu_1 \\ \mu_2 \end{bmatrix} +b\,\Db_\sigma^{-1}\Db_y\ones_n-\Db_\sigma^{-1}\deltab_{\g} \, \big)_- \Big\|_2 \nonumber\\&-\mu_1\frac{\h_d^T\ub_1}{\sqrt{n}} - \mu_2\frac{\h_d^T\ub_2}{\sqrt{n}} - \alpha\frac{\|\h_d^T\Ub^\perp\|_2}{\sqrt{n}}\leq 0.\nn
\end{align}
where as in Section \ref{sec:cs_proof_app} we have decomposed the matrix of means $\M=\Ub\Sb\Vb^T$ and $\mu_1,\mu_2,\alpha$ above represent $\ub_1^T\w$, $\ub_1^T\w$ and $\|\w^\perp\|_2$.
%
Now, by law of large numbers, notice that for fixed $(q,\mu_1,\mu_2,\alpha,b)$, the functional in the constraint above converges in probability to
\begin{align}\label{eq:Ln2L}
\bar L(q,\mu_1,\mu_2,\alpha,b):=\sqrt{\E\big(qG+\Sigma_S^{-1}\widetilde{E}_S^T\Vb\Sb\begin{bmatrix}\mu_1 \\ \mu_2\end{bmatrix} + {b\,\Sigma_S^{-1}Y-\Sigma_S^{-1}{\widetilde{\Delta}}_S}\big)_{-}^2} - \alpha\sqrt{\gamma},
\end{align}
where the random variables $G,\widetilde{E}_S,Y,{\widetilde{\Delta}}_S$ and $\Sigma_S$ are as in the statement of the theorem. 
%
Thus, the deterministic equivalent (high-dimensional limit) of the AO expressed in variables
$
\rhob = \begin{bmatrix}\rhob_1\\\rhob_2 \end{bmatrix}:=\begin{bmatrix}\mu_1/q \\ \mu_2/q\end{bmatrix}
$
becomes (cf. Eqn. \eqref{eq:cs_AO_scal_det2}):
\begin{align}\label{eq:gs_AO_scal_det2}
&\min_{  q^2+b^2 \leq R, q>0, \|\rhob\|_2\leq 1 }~\frac{1}{2}q^2
\\
&\text{sub. to}\quad  {\E\big(G+\Sigma_S^{-1}\widetilde{E}_S^T\Vb\Sb\rhob + \frac{b\,\Sigma_S^{-1}Y-\Sigma_S^{-1}\widetilde\Delta_S}{q}\big)_{-}^2} \nn \leq \left({1-\|\rhob\|_2^2}\right){\gamma}.
\end{align}
Now, recall the definition of the function $\wtd\eta_\delta$ in the statement of the theorem and  observe that the constraint above is nothing but
$$
\wtd\eta_\delta(q,\rhob,b)\leq 0.
$$
Thus, \eqref{eq:gs_AO_scal_det2} becomes
\begin{align}\label{eq:gs_AO_scal_det3}
\min\left\{q^2~\Big|~ 0\leq q\leq \sqrt{R} \quad\text{ and }\quad \min_{b^2 \leq R-q^2, \|\rhob\|_2\leq 1} \wtd\eta_\delta(q,\rhob,b)\leq 0 \right\}.
\end{align}

The remaining steps of the proof are very similar to those in Section \ref{sec:cs_proof_app} and are omitted.

\subsection{Phase transition of GS-SVM}\label{sec:PT_GS-SVM}
The phase-transition threshold $\wtd \gamma_\star$ of feasibility of  the GS-SVM is the same as the threshold of feasibility of the standard SVM for the same model (see Section \ref{sec:proof_gd}). But, the feasibility threshold of SVM under the group GMM with $K=2$ groups is different from that of Section \ref{sec:PT_CS-SVM} for $K=1$, since now each class is itself a mixture of Gaussians. We derive the desired result from \cite{gkct_multiclass}, who recently studied the separability question for the more general case of a multiclass mixture of mixtures of Gaussians. 

\begin{propo} Consider the same data model and notation as in Theorem \ref{thm:main_group_fairness} and consider the event
$$
\Ec_{{\rm sep}, n}:=\left\{\exists (\w,b)\in\R^d\times\R~~\text{s.t.}~~y_i(\w^T\x_i+b)\geq 1,\,~\forall i\in[n]\right\}.
$$
Define  threshold $\gamma_\star:=\gamma_\star(\Vb,\Sb,\pi)$ as follows:
\begin{align}
\wtd\gamma_\star:=\min_{\tb\in\R^r, b\in\R}\E\left[\Big(\sqrt{1+\|\tb\|_2^2} \,G + \widetilde{E}_S^T\Vb\Sb\tb - bY\Big)_-^2\right].
\end{align}
Then, the following hold:
\begin{align}
\gamma>\wtd\gamma_\star \Rightarrow \lim_{n\rightarrow\infty}\Pro(\Ec_{{\rm sep}, n}) = 1\qquad\text{ and }\qquad \gamma<\wtd\gamma_\star \Rightarrow \lim_{n\rightarrow\infty}\Pro(\Ec_{{\rm sep}, n}) = 0.\nn
\end{align}
In words, the data are linearly separable with overwhelming probability if and only if $\gamma>\wtd\gamma_\star$. Furthermore, if this condition holds, then GS-SVM is feasible with overwhelming probability for any value of $\delta>0$.
\end{propo}

%% file: main_arxiv_v3_nov2021.bbl
\newcommand{\etalchar}[1]{$^{#1}$}
\begin{thebibliography}{DOBD{\etalchar{+}}18}

\bibitem[AG82]{andersen1982cox}
Per~Kragh Andersen and Richard~D Gill.
\newblock Cox's regression model for counting processes: a large sample study.
\newblock {\em The annals of statistics}, pages 1100--1120, 1982.

\bibitem[AH18]{azizan2018stochastic}
Navid Azizan and Babak Hassibi.
\newblock Stochastic gradient/mirror descent: Minimax optimality and implicit
  regularization.
\newblock {\em arXiv preprint arXiv:1806.00952}, 2018.

\bibitem[AKLZ20]{aubin2020generalization}
Benjamin Aubin, Florent Krzakala, Yue~M Lu, and Lenka Zdeborov{\'a}.
\newblock Generalization error in high-dimensional perceptrons: Approaching
  bayes error with convex optimization.
\newblock {\em arXiv preprint arXiv:2006.06560}, 2020.

\bibitem[AKT19]{ali2019continuous}
Alnur Ali, J~Zico Kolter, and Ryan~J Tibshirani.
\newblock A continuous-time view of early stopping for least squares
  regression.
\newblock In {\em The 22nd International Conference on Artificial Intelligence
  and Statistics}, pages 1370--1378. PMLR, 2019.

\bibitem[BBEKY13]{bean2013optimal}
Derek Bean, Peter~J Bickel, Noureddine El~Karoui, and Bin Yu.
\newblock Optimal m-estimation in high-dimensional regression.
\newblock {\em Proceedings of the National Academy of Sciences},
  110(36):14563--14568, 2013.

\bibitem[BL19]{byrd2019effect}
Jonathon Byrd and Zachary Lipton.
\newblock What is the effect of importance weighting in deep learning?
\newblock In {\em International Conference on Machine Learning}, pages
  872--881. PMLR, 2019.

\bibitem[BLLT20]{bartlett2020benign}
Peter~L Bartlett, Philip~M Long, G{\'a}bor Lugosi, and Alexander Tsigler.
\newblock Benign overfitting in linear regression.
\newblock {\em Proceedings of the National Academy of Sciences},
  117(48):30063--30070, 2020.

\bibitem[BM11]{BayMon}
Mohsen Bayati and Andrea Montanari.
\newblock The dynamics of message passing on dense graphs, with applications to
  compressed sensing.
\newblock {\em Information Theory, IEEE Transactions on}, 57(2):764--785, 2011.

\bibitem[BMM18]{belkin2018understand}
Mikhail Belkin, Siyuan Ma, and Soumik Mandal.
\newblock To understand deep learning we need to understand kernel learning.
\newblock In {\em International Conference on Machine Learning}, pages
  541--549, 2018.

\bibitem[BS16]{barocas2016big}
Solon Barocas and Andrew~D Selbst.
\newblock Big data's disparate impact.
\newblock {\em Calif. L. Rev.}, 104:671, 2016.

\bibitem[CLOT20]{chang2020provable}
Xiangyu Chang, Yingcong Li, Samet Oymak, and Christos Thrampoulidis.
\newblock Provable benefits of overparameterization in model compression: From
  double descent to pruning neural networks, 2020.

\bibitem[cod]{code}
Code for paper: Label-imbalanced and group-sensitive classification under
  overparameterization.
\newblock \url{https://github.com/orparask/VS-Loss}.

\bibitem[CS{\etalchar{+}}20]{candes2020phase}
Emmanuel~J Cand{\`e}s, Pragya Sur, et~al.
\newblock The phase transition for the existence of the maximum likelihood
  estimate in high-dimensional logistic regression.
\newblock {\em The Annals of Statistics}, 48(1):27--42, 2020.

\bibitem[CWG{\etalchar{+}}19]{TengyuMa}
Kaidi Cao, Colin Wei, Adrien Gaidon, Nikos Arechiga, and Tengyu Ma.
\newblock Learning imbalanced datasets with label-distribution-aware margin
  loss.
\newblock In {\em Advances in Neural Information Processing Systems}, pages
  1567--1578, 2019.

\bibitem[DKT19]{deng2019model}
Zeyu Deng, Abla Kammoun, and Christos Thrampoulidis.
\newblock A model of double descent for high-dimensional binary linear
  classification.
\newblock {\em arXiv preprint arXiv:1911.05822}, 2019.

\bibitem[DL20]{dhifallah2020precise}
Oussama Dhifallah and Yue~M. Lu.
\newblock A precise performance analysis of learning with random features,
  2020.

\bibitem[DM16]{montanari13}
David Donoho and Andrea Montanari.
\newblock High dimensional robust m-estimation: Asymptotic variance via
  approximate message passing.
\newblock {\em Probability Theory and Related Fields}, 166(3-4):935--969, 2016.

\bibitem[DMM09]{AMPmain}
David~L Donoho, Arian Maleki, and Andrea Montanari.
\newblock Message-passing algorithms for compressed sensing.
\newblock {\em Proceedings of the National Academy of Sciences},
  106(45):18914--18919, 2009.

\bibitem[DOBD{\etalchar{+}}18]{donini2018empirical}
Michele Donini, Luca Oneto, Shai Ben-David, John Shawe-Taylor, and Massimiliano
  Pontil.
\newblock Empirical risk minimization under fairness constraints.
\newblock {\em arXiv preprint arXiv:1802.08626}, 2018.

\bibitem[FSV16]{friedler2016possibility}
Sorelle~A Friedler, Carlos Scheidegger, and Suresh Venkatasubramanian.
\newblock On the (im) possibility of fairness.
\newblock {\em arXiv preprint arXiv:1609.07236}, 2016.

\bibitem[Gor85]{Gor2}
Yehoram Gordon.
\newblock Some inequalities for gaussian processes and applications.
\newblock {\em Israel Journal of Mathematics}, 50(4):265--289, 1985.

\bibitem[GPSW17]{guo2017calibration}
Chuan Guo, Geoff Pleiss, Yu~Sun, and Kilian~Q Weinberger.
\newblock On calibration of modern neural networks.
\newblock In {\em International Conference on Machine Learning}, pages
  1321--1330. PMLR, 2017.

\bibitem[HMRT19]{hastie2019surprises}
Trevor Hastie, Andrea Montanari, Saharon Rosset, and Ryan~J Tibshirani.
\newblock Surprises in high-dimensional ridgeless least squares interpolation.
\newblock {\em arXiv preprint arXiv:1903.08560}, 2019.

\bibitem[HNSS18]{hu_does}
Weihua Hu, Gang Niu, Issei Sato, and Masashi Sugiyama.
\newblock Does distributionally robust supervised learning give robust
  classifiers?
\newblock In Jennifer Dy and Andreas Krause, editors, {\em Proceedings of the
  35th International Conference on Machine Learning}, volume~80 of {\em
  Proceedings of Machine Learning Research}, pages 2029--2037. PMLR, 10--15 Jul
  2018.

\bibitem[HPS16]{hardt2016equality}
Moritz Hardt, Eric Price, and Nathan Srebro.
\newblock Equality of opportunity in supervised learning.
\newblock {\em arXiv preprint arXiv:1610.02413}, 2016.

\bibitem[HZRS16]{he2016deep}
Kaiming He, Xiangyu Zhang, Shaoqing Ren, and Jian Sun.
\newblock Deep residual learning for image recognition.
\newblock In {\em Proceedings of the IEEE conference on computer vision and
  pattern recognition}, pages 770--778, 2016.

\bibitem[JT18]{ji2018risk}
Ziwei Ji and Matus Telgarsky.
\newblock Risk and parameter convergence of logistic regression.
\newblock {\em arXiv preprint arXiv:1803.07300}, 2018.

\bibitem[JT21]{ji2021characterizing}
Ziwei Ji and Matus Telgarsky.
\newblock Characterizing the implicit bias via a primal-dual analysis.
\newblock In {\em Algorithmic Learning Theory}, pages 772--804. PMLR, 2021.

\bibitem[KA20]{kammoun2020precise}
Abla Kammoun and Mohamed-Slim Alouini.
\newblock On the precise error analysis of support vector machines.
\newblock {\em arXiv preprint arXiv:2003.12972}, 2020.

\bibitem[KHB{\etalchar{+}}18]{cosen}
Salman~H. Khan, Munawar Hayat, Mohammed Bennamoun, Ferdous~A. Sohel, and
  Roberto Togneri.
\newblock Cost-sensitive learning of deep feature representations from
  imbalanced data.
\newblock {\em IEEE Transactions on Neural Networks and Learning Systems},
  29(8):3573--3587, 2018.

\bibitem[KK20]{KimKim}
Byungju Kim and Junmo Kim.
\newblock Adjusting decision boundary for class imbalanced learning.
\newblock {\em IEEE Access}, 8:81674--81685, 2020.

\bibitem[KMR16]{kleinberg2016inherent}
Jon Kleinberg, Sendhil Mullainathan, and Manish Raghavan.
\newblock Inherent trade-offs in the fair determination of risk scores.
\newblock {\em arXiv preprint arXiv:1609.05807}, 2016.

\bibitem[KT20]{gkct_dd2020}
G.~R. {Kini} and C.~{Thrampoulidis}.
\newblock Analytic study of double descent in binary classification: The impact
  of loss.
\newblock In {\em 2020 IEEE International Symposium on Information Theory
  (ISIT)}, pages 2527--2532, 2020.

\bibitem[KT21a]{gkct_multiclass}
Ganesh Kini and Christos Thrampoulidis.
\newblock Phase transitions for one-vs-one and one-vs-all linear separability
  in multiclass gaussian mixtures.
\newblock {\em International Conference on Acoustics, Speech, and Signal
  Processing}, 2021.

\bibitem[KT21b]{gkct_icassp}
Ganesh~Ramachandra Kini and Christos Thrampoulidis.
\newblock Phase transitions for one-vs-one and one-vs-all linear separability
  in multiclass gaussian mixtures.
\newblock In {\em ICASSP 2021 - 2021 IEEE International Conference on
  Acoustics, Speech and Signal Processing (ICASSP)}, pages 4020--4024, 2021.

\bibitem[KXR{\etalchar{+}}20]{kang2020decoupling}
Bingyi Kang, Saining Xie, Marcus Rohrbach, Zhicheng Yan, Albert Gordo, Jiashi
  Feng, and Yannis Kalantidis.
\newblock Decoupling representation and classifier for long-tailed recognition,
  2020.

\bibitem[LGG{\etalchar{+}}18]{lin2018focal}
Tsung-Yi Lin, Priya Goyal, Ross Girshick, Kaiming He, and Piotr Dollár.
\newblock Focal loss for dense object detection, 2018.

\bibitem[LJD{\etalchar{+}}17]{li2017hyperband}
Lisha Li, Kevin Jamieson, Giulia DeSalvo, Afshin Rostamizadeh, and Ameet
  Talwalkar.
\newblock Hyperband: A novel bandit-based approach to hyperparameter
  optimization.
\newblock {\em The Journal of Machine Learning Research}, 18(1):6765--6816,
  2017.

\bibitem[LMZ{\etalchar{+}}19]{liu2019largescale}
Ziwei Liu, Zhongqi Miao, Xiaohang Zhan, Jiayun Wang, Boqing Gong, and Stella~X.
  Yu.
\newblock Large-scale long-tailed recognition in an open world, 2019.

\bibitem[LS20a]{liang2020precise}
Tengyuan Liang and Pragya Sur.
\newblock A precise high-dimensional asymptotic theory for boosting and
  min-l1-norm interpolated classifiers.
\newblock {\em arXiv preprint arXiv:2002.01586}, 2020.

\bibitem[LS20b]{lu2020neural}
Jianfeng Lu and Stefan Steinerberger.
\newblock Neural collapse with cross-entropy loss.
\newblock {\em arXiv preprint arXiv:2012.08465}, 2020.

\bibitem[LVD20]{lorraine2020optimizing}
Jonathan Lorraine, Paul Vicol, and David Duvenaud.
\newblock Optimizing millions of hyperparameters by implicit differentiation.
\newblock In {\em International Conference on Artificial Intelligence and
  Statistics}, pages 1540--1552. PMLR, 2020.

\bibitem[MJR{\etalchar{+}}20]{Menon}
Aditya~Krishna Menon, Sadeep Jayasumana, Ankit~Singh Rawat, Himanshu Jain,
  Andreas Veit, and Sanjiv Kumar.
\newblock Long-tail learning via logit adjustment.
\newblock {\em arXiv preprint arXiv:2007.07314}, 2020.

\bibitem[MKL{\etalchar{+}}20]{mignacco2020role}
Francesca Mignacco, Florent Krzakala, Yue Lu, Pierfrancesco Urbani, and Lenka
  Zdeborova.
\newblock The role of regularization in classification of high-dimensional
  noisy gaussian mixture.
\newblock In {\em International Conference on Machine Learning}, pages
  6874--6883. PMLR, 2020.

\bibitem[MM19]{mei2019generalization}
Song Mei and Andrea Montanari.
\newblock The generalization error of random features regression: Precise
  asymptotics and double descent curve.
\newblock {\em arXiv preprint arXiv:1908.05355}, 2019.

\bibitem[MNS{\etalchar{+}}20]{muthukumar2020classification}
Vidya Muthukumar, Adhyyan Narang, Vignesh Subramanian, Mikhail Belkin, Daniel
  Hsu, and Anant Sahai.
\newblock Classification vs regression in overparameterized regimes: Does the
  loss function matter?
\newblock {\em arXiv preprint arXiv:2005.08054}, 2020.

\bibitem[MPP20]{mixon2020neural}
Dustin~G. Mixon, Hans Parshall, and Jianzong Pi.
\newblock Neural collapse with unconstrained features, 2020.

\bibitem[MRSY19]{montanari2019generalization}
Andrea Montanari, Feng Ruan, Youngtak Sohn, and Jun Yan.
\newblock The generalization error of max-margin linear classifiers:
  High-dimensional asymptotics in the overparametrized regime.
\newblock {\em arXiv preprint arXiv:1911.01544}, 2019.

\bibitem[MSV10]{masnadi2010risk}
Hamed Masnadi-Shirazi and Nuno Vasconcelos.
\newblock Risk minimization, probability elicitation, and cost-sensitive svms.
\newblock In {\em ICML}, pages 759--766. Citeseer, 2010.

\bibitem[NKB{\etalchar{+}}19]{nakkiran2019deep}
Preetum Nakkiran, Gal Kaplun, Yamini Bansal, Tristan Yang, Boaz Barak, and Ilya
  Sutskever.
\newblock Deep double descent: Where bigger models and more data hurt.
\newblock {\em arXiv preprint arXiv:1912.02292}, 2019.

\bibitem[NLG{\etalchar{+}}19]{nacson2019convergence}
Mor~Shpigel Nacson, Jason Lee, Suriya Gunasekar, Pedro Henrique~Pamplona
  Savarese, Nathan Srebro, and Daniel Soudry.
\newblock Convergence of gradient descent on separable data.
\newblock In {\em The 22nd International Conference on Artificial Intelligence
  and Statistics}, pages 3420--3428. PMLR, 2019.

\bibitem[NSS19]{nacson2019stochastic}
Mor~Shpigel Nacson, Nathan Srebro, and Daniel Soudry.
\newblock Stochastic gradient descent on separable data: Exact convergence with
  a fixed learning rate.
\newblock In {\em The 22nd International Conference on Artificial Intelligence
  and Statistics}, pages 3051--3059, 2019.

\bibitem[OA18]{olfat2018spectral}
Mahbod Olfat and Anil Aswani.
\newblock Spectral algorithms for computing fair support vector machines.
\newblock In {\em International Conference on Artificial Intelligence and
  Statistics}, pages 1933--1942. PMLR, 2018.

\bibitem[OS19]{oymak2019overparameterized}
Samet Oymak and Mahdi Soltanolkotabi.
\newblock Overparameterized nonlinear learning: Gradient descent takes the
  shortest path?
\newblock In {\em International Conference on Machine Learning}, pages
  4951--4960. PMLR, 2019.

\bibitem[OSHL19]{oren2019distributionally}
Yonatan Oren, Shiori Sagawa, Tatsunori~B. Hashimoto, and Percy Liang.
\newblock Distributionally robust language modeling, 2019.

\bibitem[OWZY16]{ouyang2016factors}
Wanli Ouyang, Xiaogang Wang, Cong Zhang, and Xiaokang Yang.
\newblock Factors in finetuning deep model for object detection, 2016.

\bibitem[PGC{\etalchar{+}}17]{paszke2017automatic}
Adam Paszke, Sam Gross, Soumith Chintala, Gregory Chanan, Edward Yang, Zachary
  DeVito, Zeming Lin, Alban Desmaison, Luca Antiga, and Adam Lerer.
\newblock Automatic differentiation in pytorch.
\newblock 2017.

\bibitem[PHD20]{NC}
Vardan Papyan, XY~Han, and David~L Donoho.
\newblock Prevalence of neural collapse during the terminal phase of deep
  learning training.
\newblock {\em Proceedings of the National Academy of Sciences},
  117(40):24652--24663, 2020.

\bibitem[RWY14]{raskutti2014early}
Garvesh Raskutti, Martin~J Wainwright, and Bin Yu.
\newblock Early stopping and non-parametric regression: an optimal
  data-dependent stopping rule.
\newblock {\em The Journal of Machine Learning Research}, 15(1):335--366, 2014.

\bibitem[RZH03]{rosset2003margin}
Saharon Rosset, Ji~Zhu, and Trevor Hastie.
\newblock Margin maximizing loss functions.
\newblock In {\em NIPS}, pages 1237--1244, 2003.

\bibitem[SAH19]{salehi2019impact}
Fariborz Salehi, Ehsan Abbasi, and Babak Hassibi.
\newblock The impact of regularization on high-dimensional logistic regression.
\newblock {\em arXiv preprint arXiv:1906.03761}, 2019.

\bibitem[SC19]{Sur14516}
Pragya Sur and Emmanuel~J. Cand{\`e}s.
\newblock A modern maximum-likelihood theory for high-dimensional logistic
  regression.
\newblock {\em Proceedings of the National Academy of Sciences},
  116(29):14516--14525, 2019.

\bibitem[SHN{\etalchar{+}}18]{soudry2018implicit}
Daniel Soudry, Elad Hoffer, Mor~Shpigel Nacson, Suriya Gunasekar, and Nathan
  Srebro.
\newblock The implicit bias of gradient descent on separable data.
\newblock {\em The Journal of Machine Learning Research}, 19(1):2822--2878,
  2018.

\bibitem[SKHL19]{sagawa2019distributionally}
Shiori Sagawa, Pang~Wei Koh, Tatsunori~B Hashimoto, and Percy Liang.
\newblock Distributionally robust neural networks for group shifts: On the
  importance of regularization for worst-case generalization.
\newblock {\em arXiv preprint arXiv:1911.08731}, 2019.

\bibitem[SRKL20]{sagawa2020investigation}
Shiori Sagawa, Aditi Raghunathan, Pang~Wei Koh, and Percy Liang.
\newblock An investigation of why overparameterization exacerbates spurious
  correlations.
\newblock In {\em International Conference on Machine Learning}, pages
  8346--8356. PMLR, 2020.

\bibitem[STK99]{shawe1999optimizing}
Grigoris Karakoulas~John Shawe-Taylor and Grigoris Karakoulas.
\newblock Optimizing classifiers for imbalanced training sets.
\newblock {\em Advances in neural information processing systems}, 11(11):253,
  1999.

\bibitem[Sto13]{StojLAS}
Mihailo Stojnic.
\newblock A framework to characterize performance of lasso algorithms.
\newblock {\em arXiv preprint arXiv:1303.7291}, 2013.

\bibitem[TAH18]{thrampoulidis2018precise}
Christos Thrampoulidis, Ehsan Abbasi, and Babak Hassibi.
\newblock Precise error analysis of regularized $ m $-estimators in high
  dimensions.
\newblock {\em IEEE Transactions on Information Theory}, 64(8):5592--5628,
  2018.

\bibitem[TOH15]{thrampoulidis2015regularized}
Christos Thrampoulidis, Samet Oymak, and Babak Hassibi.
\newblock Regularized linear regression: A precise analysis of the estimation
  error.
\newblock In {\em Conference on Learning Theory}, pages 1683--1709, 2015.

\bibitem[TPT20a]{taheri2020fundamental}
Hossein Taheri, Ramtin Pedarsani, and Christos Thrampoulidis.
\newblock Fundamental limits of ridge-regularized empirical risk minimization
  in high dimensions.
\newblock {\em arXiv preprint arXiv:2006.08917}, 2020.

\bibitem[TPT20b]{taheri2020sharp}
Hossein Taheri, Ramtin Pedarsani, and Christos Thrampoulidis.
\newblock Sharp asymptotics and optimal performance for inference in binary
  models.
\newblock In {\em International Conference on Artificial Intelligence and
  Statistics}, pages 3739--3749. PMLR, 2020.

\bibitem[TWL{\etalchar{+}}20]{tan2020equalization}
Jingru Tan, Changbao Wang, Buyu Li, Quanquan Li, Wanli Ouyang, Changqing Yin,
  and Junjie Yan.
\newblock Equalization loss for long-tailed object recognition.
\newblock In {\em Proceedings of the IEEE/CVF Conference on Computer Vision and
  Pattern Recognition}, pages 11662--11671, 2020.

\bibitem[TXH18]{thrampoulidis2018symbol}
Christos Thrampoulidis, Weiyu Xu, and Babak Hassibi.
\newblock Symbol error rate performance of box-relaxation decoders in massive
  mimo.
\newblock {\em IEEE Transactions on Signal Processing}, 66(13):3377--3392,
  2018.

\bibitem[WC03]{wu2003class}
Gang Wu and Edward~Y Chang.
\newblock Class-boundary alignment for imbalanced dataset learning.
\newblock In {\em ICML 2003 workshop on learning from imbalanced data sets II,
  Washington, DC}, pages 49--56, 2003.

\bibitem[WCLL18]{Wang_2018}
Feng Wang, Jian Cheng, Weiyang Liu, and Haijun Liu.
\newblock Additive margin softmax for face verification.
\newblock {\em IEEE Signal Processing Letters}, 25(7):926–930, Jul 2018.

\bibitem[XM89]{xie1989logit}
Yu~Xie and Charles~F Manski.
\newblock The logit model and response-based samples.
\newblock {\em Sociological Methods \& Research}, 17(3):283--302, 1989.

\bibitem[YCZC20]{CDT}
Han-Jia Ye, Hong-You Chen, De-Chuan Zhan, and Wei-Lun Chao.
\newblock Identifying and compensating for feature deviation in imbalanced deep
  learning, 2020.

\bibitem[ZCO20]{zhao2020role}
Yuan Zhao, Jiasi Chen, and Samet Oymak.
\newblock On the role of dataset quality and heterogeneity in model confidence.
\newblock {\em arXiv preprint arXiv:2002.09831}, 2020.

\bibitem[ZCWC20]{zhou2020bbn}
Boyan Zhou, Quan Cui, Xiu-Shen Wei, and Zhao-Min Chen.
\newblock Bbn: Bilateral-branch network with cumulative learning for
  long-tailed visual recognition, 2020.

\end{thebibliography}
